\documentclass[11pt,fleqn,twoside,titlepage]{cslarticle}
\cslreportnumber{\ifelse{}{CSL} Technical Report \ifelse{for CLARISSA Project}{SRI-CSL-2022-02R2}}
\newcommand{\ifelse}[2]{#2}

\ifelse{\acknowledge{SRI Project 100651 under subcontract to Honeywell
in support of AFRL and DARPA ARCOS Program
Contract FA8750-20-C-0512.}}{}

\newcommand{\yy}{}
\newcommand{\cl}{{\sc Clarissa}}
\newcommand{\cla}{\ifelse{{\sc Clarissa}}{Assurance 2.0}}
\newcommand{\acla}{\ifelse{a {\sc Clarissa}}{an Assurance 2.0}}
\newcommand{\clasce}{{\sc Clarissa/asce}}
\ifelse{\newcommand{\CL}{C{\smaller LARISSA}}}{}
\ifelse{\newcommand{\Cl}{C{\smaller\smaller LARISSA}}}{}

\newcommand{\vbar}{\,|\,}
\newcommand{\conf}{P_{\!\mathit{conf}}}
\newcommand{\pfd}{P_\mathit{fd}}
\newcommand{\pff}{P_{\mathit{fif}}}
\newcommand{\psurv}{P_{\!\mathit{srv}}}
\usepackage[bookmarks=true,hyperfigures=true,colorlinks=true,linkcolor=blue,citecolor=blue,backref=page,pagebackref=false,pdfpagemode=fullscreen,plainpages=false,pdfpagelabels]{hyperref}
\usepackage[all]{hypcap}
\usepackage{wrapfig}
\usepackage[normalem]{ulem} 
\usepackage{caption,cite,url,relative,latexsym,alltt}
\usepackage[utf8]{inputenc}
\DeclareUnicodeCharacter{25A1}{\ensuremath\Box}
\DeclareUnicodeCharacter{25C7}{\ensuremath\Diamond}
\DeclareSymbolFont{symbolsC}{U}{txsyc}{m}{n}
\DeclareMathSymbol{\strictif}{\mathrel}{symbolsC}{74}

\def\BigLaTeX{{\rm L\kern-.36em\raise.3ex\hbox{\smaller\smaller A}\kern-.15em
    T\kern-.1667em\lower.7ex\hbox{E}\kern-.125emX}}
\def\BoldLaTeX{{\bf L\kern-.36em\raise.3ex\hbox{\smaller\smaller\bf A}\kern-.15em
    T\kern-.1667em\lower.7ex\hbox{E}\kern-.125emX}}
\def\BibTeX{{\rm B\kern-.05em{\sc i\kern-.025em b}\kern-.08em
    T\kern-.1667em\lower.7ex\hbox{E}\kern-.125emX}}

\newlength{\hsbw}
\def\extrawidth{0.5in}

\newcounter{sessioncount}
\newenvironment{session*}{\begin{flushleft}
 \refstepcounter{sessioncount}
 \setlength{\hsbw}{\linewidth}
 \addtolength{\hsbw}{-\arrayrulewidth}
 \addtolength{\hsbw}{-\tabcolsep}
 \begin{tabular}{@{}|c@{}|@{}}\hline 
 \begin{minipage}[b]{\hsbw}
 \vspace*{-.5pt}
 \begin{flushright}
 \rule{0.01in}{.15in}\rule{0.3in}{0.01in}\hspace{-0.35in}
 \raisebox{0.04in}{\makebox[0.3in][c]{\footnotesize \thesessioncount}}
 \end{flushright}
 \vspace*{-.57in}
 \begingroup\small\vspace*{1.0ex}\begin{alltt}}{\end{alltt}\endgroup\end{minipage}\\ \hline 
 \end{tabular}
 \end{flushleft}}
\def\sessionsize{\small}
\def\smallsessionsize{\small}

\newcommand{\exmemo}[1]{}
\newcommand{\memo}[1]{\mbox{}\par\vspace{0.25in}
\setlength{\hsbw}{\linewidth}
\addtolength{\hsbw}{-2\fboxsep}
\addtolength{\hsbw}{-2\fboxrule}
\noindent\fbox{\parbox{\hsbw}{{\bf Memo: }#1}}\vspace{0.25in}}

\newcommand{\mem}[2]{\mbox{}\par\vspace{0.25in}
\setlength{\hsbw}{\linewidth}
\addtolength{\hsbw}{-2\fboxsep}
\addtolength{\hsbw}{-2\fboxrule}
\noindent\fbox{\parbox{\hsbw}{{\bf #1: }#2}}\vspace{0.25in}}
\newcommand{\comment}[1]{}

\newcommand{\exfootnote}[1]{}
\newlength{\sblen}
\newlength{\overhang}

\def\SetFigFont#1#2#3{\rm}

\newcommand{\excite}[1]{}

\newcommand{\arxiv}[1]{\href{https://arxiv.org/abs/#1}{\tt arXiv:#1}}

\renewcommand{\memo}[1]{}
\renewcommand{\mem}[2]{}

\newcommand{\mand}{{\texttt{ AND }}}

\title{Assessing Confidence \ifelse{in Assurance Cases\\ With \CL}{with
Assurance 2.0}\\[1ex]
}
\author{Robin Bloomfield (Adelard, part of NCC Group, and City, Univ.\ of London)\\ and John Rushby (SRI)
\\[1ex]As Members of the  C{\small LARISSA} Team\\
\emph{\smaller Honeywell, Adelard, UT Dallas, and SRI}
\ifelse{}{\\[1ex]Also issued as a  C{\small LARISSA} Technical Report under the title\\
Assessing Confidence in Assurance Cases with  C{\small LARISSA}}
}
\begin{document}
\maketitle

\newpage
\vspace*{\ifelse{-5ex}{-8ex}}

\centerline{\textbf{Abstract}}
\mbox{}\\\vspace*{-2ex}

An assurance case is intended to provide justifiable confidence in the
truth of its top claim, which typically concerns safety or security.
A natural question is then ``how much'' confidence does the case
provide?  

In this report, we explore issues in assessing confidence for
assurance cases developed \ifelse{in \cl}{using the rigorous approach
we call Assurance 2.0}.  We argue that confidence cannot be reduced to
a single attribute or measurement.  Instead, we suggest it should be
based on attributes that draw on three different perspectives:
positive, negative, and residual doubts.

\emph{Positive Perspectives} consider the extent to which the evidence
and overall argument of the case combine to make a positive statement
justifying belief in its claims.  We set a high bar for justification,
requiring it to be \emph{indefeasible}.  The primary positive measure
for this is \emph{soundness}, which interprets the argument as a
logical proof and delivers a yes/no measurement.  The interior steps
of \acla\ case can be evaluated as logical axioms, but the evidential
steps at the leaves derive logical claims epistemically---from
observations or measurements about the system and its
environment---and must be treated as premises.  Confidence in these
can be expressed probabilistically and we use \emph{confirmation
measures} to ensure (possibly informally) that the probabilistic
``weight'' of evidence crosses some threshold.

In addition, probabilities can be aggregated from evidence through the
steps of the argument using probability logics to yield what we call
\emph{probabilistic valuations} for the claims (in contrast to
soundness, which is a logical valuation).  The aggregated probability
attached to the top claim can be interpreted as a numerical measure of
confidence.  We apply probabilistic valuations only to sound cases,
and this avoids some of the difficulties that attend probabilistic
methods that stand alone.  The primary uses for probabilistic
valuations are with less critical systems, where we trade assurance
effort against confidence, and in assessing residual doubts.

\emph{Negative Perspectives} record doubts and challenges to the case,
typically expressed as \emph{defeaters}, and their exploration and
resolution.  Assurance developers must guard against confirmation bias
and should vigorously explore potential defeaters as they develop the
case, and should record them and their resolution to avoid rework and
optionally to aid reviewers.

\emph{Residual Doubts}: the world is uncertain so not all potential
defeaters can be resolved.  For example, we may design a system to
tolerate two faults and have good reasons and evidence to suppose that
is sufficient to cover the exposure on any expected mission.  But
doubts remain: what if more than two faults do arrive?  Here we can
explore consequences and likelihoods and thereby assess risk (their
product).  Some of these residual risks may be unacceptable and
thereby prompt a review, but others may be considered acceptable or
unavoidable.  It is crucial however that these judgments are
conscious ones and that they are recorded in the assurance case.

This report examines each of these three perspectives in detail and
indicates how \ifelse{\cl}{\cl, our prototype toolset for Assurance
2.0,} assists in their evaluation.

\newpage

\tableofcontents

\newpage

\listoffigures

\cleardoublepage

\section{Introduction}

\ifelse{\cl\ (``Consistent Logical Automated Reasoning for Integrated
System Software Assurance'') is a project led by Honeywell within the
DARPA ARCOS Program.  Its other members are Adelard (part of NCC
Group), SRI International, and UT Dallas; its purpose is to develop
improved methods and tools for assurance of critical systems.}{}

Assurance is the process of developing claims and collecting evidence
about a system and its environment and using these in support of an
argument to justify (or reject) deployment of the system on the
grounds of safety, security, or other designated \emph{critical
properties}.  An \emph{assurance case} is a way of organizing and
presenting this information, together with other relevant facts,
knowledge, models, and theories in a manner that facilitates overall
comprehension and assessment.  Notice that an assurance case is a way
of organizing and presenting information, and it can also help
systematize its collection and generation, but this information
articulates underlying reasons why the system justifiably satisfies
its critical properties: an assurance case cannot create these
reasons, which must have been part of the system's conception and
design.  Ideally, the assurance case is developed alongside the
system, but often it is developed retrospectively: for example, when
systems are repurposed, or when COTS elements are used as components
in a new system.  Retrospective assurance requires discovery and
articulation of reasons why the system or component possesses
properties claimed for it, or locating it within an architecture where
it needs little or no trust.  Note that the first of these may be
infeasible for systems that use Artificial Intelligence or Machine
Learning because they lack the structure and design that could provide
reasons for belief in claimed properties.

An assurance case serves (at least) two different audiences having
different goals.  First, the developers of the system and of its
assurance case (who may or may not be the same people) need to satisfy
themselves that the system is safe (or has some other designated
critical property) and that its assurance case justifies this---and to
do this they need to consider and explore every aspect of the system
and its assurance.  Second, evaluators and certifiers who will decide
or recommend deployment of the system need to assure themselves that
the developers have indeed done this, and that their conclusions are
unimpeachable: it is not the task of the evaluators to duplicate the
work of the developers by reconstructing the assurance case nor (in
general) to examine its every detail.  Instead, they need to
understand the underlying reasons why the system has the claimed
properties, and to comprehend the overall argument and supporting
evidence, with perhaps some ``deep dives'' into areas of particular
interest.

For the first purpose and audience, an assurance case must support the
management, organization, exploration, and analysis of a large body of
information.  Typically, a substantial part of the case will be an
argument justifying claims about the system on the basis of observed
or measured evidence, and the overall case may be summarized in an
``assurance report.''  For the second, a case must support the
socio-technical process of making, justifying, and communicating the
decision to deploy a system or service in a given context.  Just as
development of a system and its assurance concludes with completion of
its assurance case and report, so we recommend that evaluation of an
assurance case should conclude with an explicit ``sentencing
statement'' that describes the evaluators' understanding of the system
and its issues, their assessment of the case, and their decision on
deployment of the system.  (We sometimes refer to this as the ``case
about the case'' or ``metacase.'')

This report describes capabilities and measures that are intended to
assist developers in constructing, exploring, and critically examining
their assurance case in its full detail and in achieving justifiable
confidence in its quality; many of these capabilities will also be
helpful to evaluators and useful in support of their sentencing
statement, and some functions are provided specifically to assist
them.  Notice that we may wish to perform some assessments and
measures differently during construction of a case than at its end and
during evaluation.  During construction, we may wish to evaluate the
part of the case we have constructed and ignore parts that are missing
(or assume they are good), whereas the existence of missing parts
would be catastrophic in a final evaluation.  Similarly we may
tolerate uninvestigated doubts during construction but will want them
resolved for final evaluation.  For the most part, we describe the
topics in this report from the developer's point of view during and at
the conclusion of construction.  Figure \ref{process} illustrates a
possible process for assessment of the aspect we call ``soundness'' at
various stages of development and review.  Other aspects will follow
similar processes, and these may be interleaved with each other.
Evaluation, as opposed to construction of an assurance case is topic
that will be explored in more detail in a forthcoming report \cite{x}.

Assurance cases have evolved and changed over the past 30 or more
years (see \cite[Chapter 2]{Rushby-etal:NASACases15} for a brief
history); we advocate a further step in their development that we call
Assurance 2.0 \cite{Bloomfield&Rushby:Assurance2} whose slogan, which
might seem paradoxical at first, is ``simplicity through rigor.''  In
keeping with this, our fundamental requirement is that a case should
provide \emph{indefeasible justification} for the decision to deploy
the system or service concerned; this means that the justification
must be so well supported, and all reasonable doubts and objections
must have been so thoroughly considered and countered, that we are
confident no credible doubts remain that could change the decision.
Indefeasibility therefore provides the ``stopping criterion'' for
both development and evaluation of an assurance case.  (See
\cite{Rushby:Shonan16} for an introduction to the epistemological
notion of indefeasibility in the context of assurance.)

The world is uncertain and our understanding imperfect, so this
assessment is very demanding.  Thus it is often necessary to examine and
evaluate an assurance case from several diverse perspectives that are
combined to yield an overall assessment of its indefeasibility.  For
example, some perspectives will focus on ``positive'' aspects of the
case, such as the evidence and argument in support of its claims,
while others will consider the ``negative'' aspects (i.e., its
potential defeasibility), such as doubts and objections that have been
considered and refuted, and any that remain as residual risks.

This report is about different perspectives on the assessment of an
Assurance 2.0 case and the different measures that can support and
quantify those perspectives.  \ifelse{We also outline capabilities
required of tools that can manage and assist these activities and
describe those of the prototype \cl\ tool, which is based on Adelard's
{\sc ASCE} \cite{ASCE}.}{We also outline capabilities required of
tools that can manage and assist these activities; in particular we
are part of a project led by Honeywell that is developing prototype
tool support for Assurance 2.0, based on Adelard's {\sc ASCE} tool
\cite{ASCE}, that we call \cl\ (``Consistent Logical Automated
Reasoning for Integrated System Software Assurance'').}  These
capabilities have been explored and developed during construction of
the \clasce\ prototype.

Before we proceed to describe these assessments and measures, we need
to outline some ways in which an Assurance 2.0 case may differ from
traditional interpretations of assurance cases as described, for
example, in the tutorial by Holloway \cite{Holloway:5parts}.
\ifelse{Assurance cases in \cl\ follow the approach presented as
Assurance 2.0 \cite{Bloomfield&Rushby:Assurance2}, which}{Assurance
2.0 cases follow an approach that} builds on the earlier ``Claims,
Arguments, Evidence'' or CAE \cite{ASCAD} method, where the main
component of an assurance case is a \emph{structured argument}
represented as a tree of \emph{claims} linked by \emph{argument
steps}, and grounded on \emph{evidence}.  As we will explain in
Section \ref{structure}, argument steps in \cla\ are restricted to
just five basic forms that we refer to as (building) \emph{blocks} and
these are subject to strong conditions that permit a rigorous, logical
interpretation of the overall case.  We contend that these
restrictions simplify construction of an assurance case by reducing
the ``bewilderment of choice'' and clarifying what must be achieved.

Another way in which we simplify construction and evaluation of
assurance cases is by limiting the content and therefore the size of
their arguments.  Traditionally, assurance cases were largely
identified with (a graphical presentation of) their structured
argument, but we maintain that many parts of the case are best
presented outside the argument.  We concur with the traditional view
that the purpose of an assurance case is to collect and integrate
evidence and knowledge of the diverse topics that contribute to
assurance of a complex system.  But much of the knowledge pre-exists
and the contribution of the case is to select it, not develop it.
Similarly, evidence may be distilled from data or produced by
specialized analyses (e.g., formal verification) and applied to
specific representations of the system.  These should be constructed
as theories, models, and ``evidential assemblies'' that use their own
established methods of validation and presentation, and are referenced
and integrated by the assurance argument, not developed within it.
For example, the argument may have a subcase concerning hazard
analysis and will cite evidence discharging side-claims that this was
performed by a competent team using accepted practice, and will use
the list of hazards found, but it will not itself present the details
of hazard analysis.

Thus, anything that can be considered a single topic, with its own methods,
knowledge, notations and documentation is a candidate for an external
theory, model, or evidential assembly.  These are external to the
argument but part of the larger assurance case: the role of the
argument is to gather and integrate them.  The assurance
case argument will include justification that these external elements
were selected or developed and applied appropriately, but will not
include the internal details or justifications for these elements as
part of the argument itself, although it will provide links to them.
In justifying and reviewing the overall case, it is important that the
assurance argument is consistent with this wider body of knowledge and
evidence, and the ability to conduct deep dives into it is part of the
evaluative probing of the case.  The following sections develop these
ideas in a little more detail.

\subsection{Models and Theories in Support of an Assurance Case}
\label{theories}

We noted above that an assurance case will reference models and
theories in addition to evidence.  It is essential to understand the
relationships of these to each other, and how they are used in an
assurance case.  At the bottom of an assurance case, we have evidence
about the system in its operating context (henceforth we will speak of
simply the system and take this to include its context): evidence may
be concrete observations, measurements, or analyses about the system
in operation or under test, or about its design and construction.  This
evidence is used to justify logical claims about the system.  The
claims are relative to various descriptions of the system: for
example, if we say ``the tests show the system performs correctly'' we
must have a notion of ``correct'' that is described somewhere.  These
descriptions concern the behavior or attributes of the system from
some point of view (e.g., timing, power consumption, functional
behavior) and at some level of abstraction; we refer to all these
descriptions as \emph{models}.

\label{substitution}

The properties that can directly be stated and observed about the
system and its low-level models (e.g., percentage of objects correctly
identified by a vision system under test) will generally be far
removed from the properties about which we seek assurance: those will
generally concern emergent properties stated about highly abstract
models (e.g., safety of an autonomous car).  It follows that a central
task of an assurance case is to connect properties of low level models
of the system to those of high level models, and this is accomplished
by argument steps that iterate through a series of intermediate models
that generally align with steps in the design and development of the
system, as in the classical ``V'' Diagram \cite{Vmodel-wiki}.  A
typical step of this kind will seek to justify that a property $A$ of
some model $P$ ensures property $B$ of a next higher model $Q$\@.  In
Assurance 2.0 \ifelse{and \cl}{} this is precisely the purpose of a
\emph{substitution block}, one of the five kinds of step that may
appear in an assurance case argument.  The justification may be fairly
large, intricate, and possibly mathematical: it is an instance of what
we referred to earlier as a \emph{theory}.  The theory may depend on
the models $P, Q$ or properties $A, B$ satisfying some constraints,
and these will appear in the argument as side-claims on the
substitution and must be justified by their own arguments and
evidence.  Notice that this evidence may concern the model (e.g., is
it ``well formed'') and not the actual system.

An example theory is that underlying Modified Condition/Decision
Coverage (MC/DC) for requirements-based tests
\cite{Chilenski&Miller:mcdc,Chilenski&Miller94:mcdc,Hayhurst-etal01}:
such a theory must explain this method of testing and coverage
evaluation, how is it performed, why is it useful, what issues need to
be considered, and what claims it can support.  The theory could then
be used, for example, to explain how MC/DC coverage of executable code
can justify a claim that the code contains no unintended functions.
When using a theory, the argument must provide justification that it
is suitable and credible, and that it is applied appropriately but it
does not present the theory nor its assurance as part of the
argument: it merely references these.

We do this to control the size of the argument, and to allow
compositional reasoning.  The purpose of the structured argument in an
assurance case is to collect, organize, and present evidence and
information relevant to assurance of the system.  This is already a
formidable task and an assurance case argument is therefore generally
large and difficult to comprehend in its entirety.  This task should
not be further complicated by presenting and justifying technical
means of assurance or analysis within the argument.  Such analyses are
better developed, analyzed, justified, and evaluated separately as
self-contained models and theories managed by experts and presented
and assessed by the scientific and engineering methods traditional to
their fields.  These theories may conclude with suggested
``templates'' for the structured argument of an assurance case
referencing the theory, but those templates emerge as part of the
theory rather than vice versa.  

Theories and models are part of the assurance case and must be
assessed as such, but they are developed and assessed separately and
not within the argument.  The assessments of widely-used theories may
become generally accepted, so that these theories amount to
``pre-certified'' subcases for the main assurance case.  The argument
of an assurance case will integrate subcases that use several
different theories but, because the theories are referenced rather
than developed therein, the argument can be fairly systematic: at each
point, we choose a suitable kind of argument step (see the ``helping
hand'' in Figure \ref{helping-hand}) and the theory or other
``warrant'' that will justify it; instantiation of these will
determine the evidence or subclaims required, and any necessary
side-claims; the argument of the main case must establish that the
chosen theory is appropriate and its instantiation is sound, but
generally need not reconsider the assurance of a pre-certified theory.
Selection and instantiation of appropriate theories based on their
templates is sufficiently systematic that much of it can be automated.
In particular, a library of theories and their templates can provide
the basis for automated synthesis of parts of an assurance case.
\clasce\ can do this, and its synthesis procedures are described in a
companion report \cite{x}.  Furthermore, much of an assurance case can
be summarized by listing the theories that are used and the models to
which they are applied, and this can assist comprehension by
evaluators; the \clasce\ synthesizer can generate a summary ``theory
view'' for this purpose.

An example due to Holloway \cite{Holloway:5parts}, illustrates these
points by taking a contrary course (Holloway was using the example for
illustration, not advocating its method).  The example concerns a
teenager Jon who asks his dad if his friend Tim may drive him to the
game.  The dad would like to see an assurance case to justify the
claim that Tim is a safe driver.  In Holloway's presentation, four
topics are identified (and later elaborated) and provide the main
substructure of the argument for the case:
\begin{enumerate}
\item Tim has satisfied all legal requirements for driving.

\item Tim has not been in an accident.

\item Tim has a reputation for driving safely.

\item Nothing is going on in Tim's life that might cause him to drive
less safely than usual.
\end{enumerate}

These seem reasonable, but doubts remain: are they the most important
topics, and are they sufficient?  Surely we would like to see some
extended discussion of driving safety by young men, together with
historical data, statistics, and risk factors.  Holloway does provide
some of this but it is represented explicitly in the argument of the
case and would overwhelm it if included in full detail.  Furthermore,
we cannot expect the developers and evaluators of an assurance cases
to be experts in every topic that might be relevant to a case: for
example, this case is about driving safety, whereas another might
concern the safety of plans for radiation therapy, and yet another
is about mission risk due to pyrotechnic bolts in a spacecraft.

Instead, we advocate that such material is developed as a separate
theory; here, this would be an extended ``theory of driving safety by
young men'' that is constructed and evaluated by those with specialist
knowledge of the topic.  The assurance case argument references this
theory, and is structured accordingly, but the theory is not developed
within the argument.\footnote{An alternative point of view is that
these topics are the hazards posed by young male drivers and should
emerge as part of hazard analysis.  Hazard analysis is a form of
evidence assembly (see Section \ref{assembly}), so in this approach
the topics would be delivered to the argument through an evidence
incorporation step rather than a substitution step, but the principle
remains the same: evidence assembly (and hazard analysis in
particular) is an important part of the assurance case, but it is
external to the argument, just like a theory.}

Typically, the theory will justify a substitution step in which
property $A$ of model $P$ ensures property $B$ of model $Q$; most
often, either the properties or the models are the same (i.e., either
$A=B$ or $P=Q$).  Here, $P$ and $Q$ are the same ``model'' of young
men's driving, and $A$ is a collection of observable or measurable
properties of young men whose conjunction supports the claimed
property $B$ that such men are safe drivers.  It might well be that
the observable properties identified by the theory are the same four
identified by Holloway, but the structure and justification of the
assurance case will be different.  In Holloway's case, a detailed
justification for the four properties must be embedded in the
argument, whereas the alternative case justifies them by reference to
the theory, supported by side-claims attesting to the credibility of
the theory (authors, reviewers, accepted practice, previous
applications etc.) and its application to the system under
consideration.  The theory is independent of the case and is developed
and evaluated by experts and may evolve over time, whereas the
embedded justification is developed and evaluated as part of the case
by persons who presumably must also consider other topics requiring
specialized expertise such as the safety of Tim's car, and any hazards
of the route to be taken---topics that we would recommend as
additional candidates for external theories or evidence.

\subsection{Evidence and its Assembly}
\label{assembly}

An assurance case is based on evidence, and evidence can take many
forms.  For example, DO-178C \cite{DO178C} identifies 71
``objectives,'' which roughly correspond to evidentially supported
claims in an assurance case that might be retrofitted to DO-178C\@.
Some of the evidence supporting these claims we call \emph{monolithic}
because it is a single judgment or observation: for example,
``software load control is established'' \cite[Section 7.1.h]{DO178C}.
Others, we call \emph{aggregated} because they comprise judgments or
observations iterated over some set of items: for example ``test
coverage of high level requirements is achieved'' \cite[Section
6.4.4.a]{DO178C}, which iterates over the high level requirements to
deliver aggregated evidence that coverage is achieved.  In fact,
monolithic evidence often proves to be aggregated on closer
inspection.  For example, ``software load control is established,''
mentioned above, is defined as follows \cite[Section 7.1.h]{DO178C}.
\begin{quote}
``Software load control ensures that the Executable Object Code and
Parameter Data Item Files, if any, are loaded into the system or
equipment with appropriate safeguards.''
\end{quote}
This suggests that the evidence should reference a theory explaining
what load control means and what safeguards are required and how these
are achieved, and evaluation of its application will be iterated
(i.e., aggregated) over each file that supplies ``Executable Object
Code'' or ``Parameter Data Items.''  We therefore take aggregated
evidence as the standard case.

Within a structured argument, the purpose of evidence is to support a
claim; without association to a claim, we do not have evidence, merely
data.  In \cla, this association is performed by the \emph{evidence
incorporation} block of a structured argument, but it must be
justified by some process or theory that links the evidence to the
claim.  Furthermore, there must be some activity (e.g., a test
campaign) that performs the observations or judgments that provide the
underlying data and, if necessary, aggregates it into evidence.  In
addition, evidence must generally be bound together with other
information about its generation and provenance (e.g., test harness,
oracle, coverage analyzer).  We refer to this entire collection of
tasks as \emph{evidence assembly}; currently this is seldom recognized
as a coherent activity and its tasks are distributed somewhat
arbitrarily between the assurance case argument and its supporting
data collection and this can have wide-ranging consequences.

Consider, for example, what happens when one or a few items of
aggregated evidence are changed (for example, some tests are rerun).
Conceptually, evidence assembly updates its aggregated results and
makes these available to the argument of the assurance case.  We must
then consider where is evidence assembly performed?  The natural
location, based on the preceding description, is that it is external
to the tool that manages the assurance case argument.  But an
alternative design, which seems to be assumed in some discussions of
tools for assurance cases, is that it is part of that tool.  Both
approaches seem viable, and one could also imagine mixed forms, where
some kinds of aggregated evidence are assembled locally, by the
argument management tool, and others are assembled externally.

These choices may influence the structure of the assurance case that
is developed.  If evidence assembly is performed in the argument
management tool, then the tool potentially has access to unaggregated
or ``fine-grained'' evidence (e.g., individual test results) and
developers of the assurance case may choose to structure some part of
the argument at this level of detail.  We do not favor this option, as
some assessments of aggregated evidence (e.g., ``does every
requirement have a test?'') may move from evidence assembly, where we
believe it belongs, into the assurance case argument itself.  As we
have noted before, assurance case arguments can easily become
overwhelmingly large and complex and correspondingly difficult to
evaluate; we therefore suggest that it is important to identify
components of the case that can be delegated to separate analysis and
review: this includes models of the system and its artifacts, theories
concerning both design (e.g., fault tolerance) and review (e.g.,
various kinds of testing), and the aggregation and assessment of
evidence.

Observations and judgments may be the central focus of evidence, but
they do not stand alone.  Several other items that we refer to
generically as ``provenance'' are usually needed as well.  For
example, test results might be the primary evidence provided by some
activity, but we also need information on the test oracle, the test
suite, the method of test generation and measurement of coverage, the
execution or experimental platform employed, and the versions of the
requirements and software employed.  It is possible that some of these
could be supplied as separate items of evidence, but the advantage of
supplying it all in a single package is that it can be reviewed and
assessed as part of evidence assembly, and then bound together (e.g.,
with a cryptographic checksum) so that its integrity and coherence are
established and preserved.  Evidence assembly must manage all these
concerns, which will differ across different kinds of evidence, and
\yy
for this reason we favor an architecture where evidence assembly is
performed by tools that are specific to the kind of evidence concerned
and that are  external to, but closely allied with,
the tool that manages the assurance case argument.

\ifelse{
\subsection{Structure of an Assurance Case in \Cl}
}{
\subsection{Structure of an Assurance 2.0 Case}
}

\label{structure}

Following the discussion of the previous subsections, we see that an
assurance case argument is but part of a ``full'' assurance case that
also includes supporting models and theories, together with evidence
assembly and its underlying data.  \yy A tool that supports assurance
cases does not need to manage or generate the development and
assessment of models, theories, nor the data underlying evidence,
although it may do some of these and it does need to be able to
reference them and to communicate with their supporting tools (e.g.,
\clasce\ narratives can link to spreadsheets and into text documents).
And it is a design choice whether it manages evidence assembly or
delegates this to external tools but, however it is done, evidence
assembly is separate from the assurance case argument, whose
management is the prime function of the assurance case tool.

Assurance case arguments are typically presented graphically, as a
tree-like structure composed of \emph{nodes} and \emph{links} (see
Figure \ref{busy}).  \cla\ follows this approach and the \clasce\ tool
provides a graphical user interface for the construction of graphical
arguments.\exfootnote{We are considering other representations.}
There are three basic node types: \emph{claims}, \emph{arguments}, and
\emph{evidence} and these combine to produce \emph{argument steps} in
which (sub)claims or evidence nodes link to an argument node that
links to a parent claim.  The argument node provides justification
that the subclaims or evidence logically entail the parent claim; the
justification (which is typically a \emph{narrative} in natural
language that may reference external discussion or data) may require
additional (side) claims to ensure attributes such as indefeasibility.
Other kinds of node may appear in \acla\ argument, such as assumption,
comment, defeater, and subcase; some of these serve dialectical rather
than logical purposes, see Section \ref{defeaters}.

An assurance case argument, as we have specified it, has rather
limited scope and can therefore take a rather restricted form: it is
not concerned with constructing intricate chains of logical
inferences, but just a few basic steps for structuring and organizing
references to external models, theories and evidence.  Thus, \cla\
uses only five different kinds of argument steps: these are called
(building) \emph{blocks} \cite{Bloomfield&Netkachova14} and they
comprise \emph{decomposition}, \emph{substitution}, \emph{concretion},
\emph{calculation}, and \emph{evidence incorporation} and are
described in more detail in Sections \ref{interior} and \ref{leaves}
and illustrated in Figure \ref{busy}.

In the typical structure of an Assurance 2.0 case, general claims at
the upper level are refined into more precise claims using concretion
steps, then substitution steps are used to elaborate these claims
about high level models into claims about low level models and their
implementations, and these lowest level claims are discharged by
evidence.  Application of evidence is generally accomplished in two
steps: the lowest step performs evidence incorporation to justify a
claim about ``something measured'' (e.g., ``we did requirements-based
testing and achieved MC/DC coverage'') and this supports a second step
(typically a substitution based on application of an external theory)
that connects this to a claim about ``something useful'' (e.g., ``we
have no unreachable code''); see Sections \ref{leaves} and
\ref{evincsec}.  At any stage, the argument may divide into subcases
using decomposition or calculation steps that enumerate a claim over
some structure (e.g., over components, requirements, hazards, etc.) or
that split the conjuncts of a compound claim.  This structure may
recurse within subcases.

We require that argument steps are \emph{deductive} whenever possible.
This means that the claim justified by a step should be logically
implied by the conjunction of evidence or subclaims supporting it.
This is in contrast to \emph{inductive} argument steps where the
evidence and subclaims merely ``suggest'' their parent claim with
various degrees of force.  \cla\ blocks generally have
\emph{side-claims} whose purpose is to enforce deductiveness: for
example, a step that decomposes over hazards will have a side-claim
that requires justification that all hazards have been identified and
that all the individual hazards \emph{and their combinations} are
considered in the decomposition.

The reasons for advocating deductivism are that a) every inductive
step harbors an anonymous \emph{doubt} (otherwise it would be
deductive) and we prefer to eliminate such doubts if possible, or at
least to identify them explicitly so that they can be analyzed and
recorded, and b) we wish to interpret completed arguments as logical
proofs using \emph{Natural Language Deductivism} (NLD)
\cite{Groarke99}.  In NLD, the evidence incorporation steps that
constitute the leaves of the argument tree are interpreted as
\emph{a xxxx} and the interior steps as \emph{axioms} having the
form of definite clauses: that is, conjunctions of subclaims that
deductively imply their claim.

NLD is an informal counterpart to deductive proof in formal
mathematics and logic but differs in that its premises and axioms are
``reasonable or plausible'' rather than certain, and hence its
conclusions are likewise reasonable or plausible rather than certain
\cite[Section 4.2]{sep-informal-logic}.  NLD differs significantly
from standard interpretations of informal argumentation, where weaker
or different forms of inference may be used \cite{Blair15}; the very
term ``natural language deductivism'' was introduced by Govier
\cite{Govier87} as a pejorative to stress that this style of argument
does not adequately represent ``informal argument'' (of the kind that
may be employed in debate).  However, our focus is not informal
arguments in general, but the structured arguments of assurance cases,
where deductive validity is a natural counterpart to the requirement
for indefeasibility, and so we depart from the association of
assurance cases with informal argument and adopt the label NLD with
pride.

Because our treatment is close to that of formal logic, we adopt its
terminology and say that an argument is \emph{logically valid} if its
reasoning steps are logically so (i.e., true in all interpretations)
and that it is \emph{sound} if, in addition, all its steps are so well
justified that they can be accepted as true (i.e., we set a high bar
for ``reasonable or plausible'').\footnote{We explain later, in
Section \ref{interior}, that we prefer (but do not require) completed
arguments to be sound and ``fully valid'' meaning that, in addition to
being logically valid, they are deductive and have no unaddressed
doubts or defeaters.  During development, we allow arguments to be
incomplete and lacking full validity; we evaluate these as best we can
and tackle their deficiencies in an iterative manner until we can
declare that the case is complete.}

As the logic used in assurance cases is elementary (basically,
propositional calculus), logical validity simply requires that all the
argument steps ``fit together'' correctly.  That is, if a step has a
subclaim $C$, then the subargument that justifies this must deliver
precisely $C$, and not some $C\,'.$ \yy The \ifelse{\cl}{graphical}
tools that construct Assurance 2.0 arguments do so in a way that
largely guarantees logical validity.  For example, at the top of
Figure \ref{busy}, we have a decomposition block delivering a ``more
precise claim'' that is used by the concretion block above it.  It
would be a logical validity fault if the two instances of the claim
(i.e., the one delivered and the one used) were different, but this is
excluded because there is just one instance of the claim and it is
referenced by both blocks.  Nonetheless, it is possible that the
justification for the lower block is inadequate, and it really
delivers only some $C\,'$ and not $C$.  Note, however, that this is
not logical invalidity but unsoundness, which is discussed next.

A logically valid argument is \emph{sound} when all its evidence
incorporation steps cross some threshold for credibility, and all its
interior or reasoning steps have indefeasible justifications.  Whereas
logical validity is concerned only with the syntax of the argument,
soundness is concerned with its semantics and, in particular, with the
\emph{meaning} attached to its claims.  The claims in an assurance
case argument are expressed in natural language so they need to be
expressed carefully and possibly explained in more detail in the
narrative justifications.  We have observed that many assurance
arguments are internally inconsistent in the way they employ natural
language to express the concepts appearing in claims, so we are exploring
use of tools based on Large Language Models (LLMs), such as ChatGPT,
to transform natural language claims into more systematic and
consistent expressions.

As we will explain in Section \ref{confirmation}, we use
\emph{confirmation measures} to assess the ``weight'' of evidence in
support of an evidentially useful claim and thereby judge soundness
for evidence incorporation steps.  For interior reasoning steps, we
rely on human judgment to assess the justification supplied, but we
expect most of these steps to be the application of some well-accepted
theory, so the judgment builds on a reliable foundation.

Observe that the requirements for soundness (indefeasibility,
deductiveness, confirmation measures), although rigorous, are
straightforward and in that sense simple: there is no doubt what is
required in both construction and evaluation.  This is what we mean by
``simplicity through rigor'' and it may be contrasted with the complex
criteria proposed for less constrained forms of assurance case
\cite{Chowdhury-etal:Safecomp20}.

Soundness is the most fundamental property we desire of an assurance
case, but there are other useful properties and in the next subsection
we identify different perspectives and measures that can be used to
develop confidence in a case.

\subsection{Confidence in an Assurance Case}

As we stated in the beginning, the purpose of an assurance case is to
construct and deliver indefeasible justification for its top claim.
After assessing the case, we will have some degree of belief that this
purpose has been achieved and we call this degree of belief the
\emph{confidence} in the case.  Confidence is a human judgment that
should be based on evaluable and measurable attributes of the case but
we do not think it can be reduced to a single attribute or
measurement.  Instead, we think that confidence should be based on
three related criteria.

\begin{description}
\item[Positive perspectives:] Is the assurance case argument logically
valid and sound, and does it provide indefeasible justification for
its claims, and the top claim in particular?

\item[Negative perspectives:] Are there gaps, discrepancies, or
weaknesses in the claims, evidence, argument steps or justifications?

\item[Residual risks:] Are all identified gaps and weaknesses within a
    tolerable threshold of risk?

\end{description}
These perspectives are interrelated: negative perspectives propose and
explore possible \emph{defeaters} while positive perspectives require all
defeaters to be resolved or identified as residual risks.  We now
describe these perspectives in more detail.

\paragraph{Positive Perspectives:} These consider the extent to which
the evidence and overall argument make a positive case to justify
belief in its claims.  We have already encountered one positive
measure: namely, \emph{soundness}, which builds on logical
\emph{validity} and interprets the argument as a logical proof by NLD
and delivers a yes/no measurement.  Given a valid argument (which we
discuss below), soundness can be assessed compositionally: that is, by
considering the interpretation of each step in the argument.  We do
this differently for interior or reasoning steps, and leaf or
evidential steps.

\begin{figure}[p]
\begin{center}
\vspace*{-12ex}
\includegraphics[height=1.10\textheight]{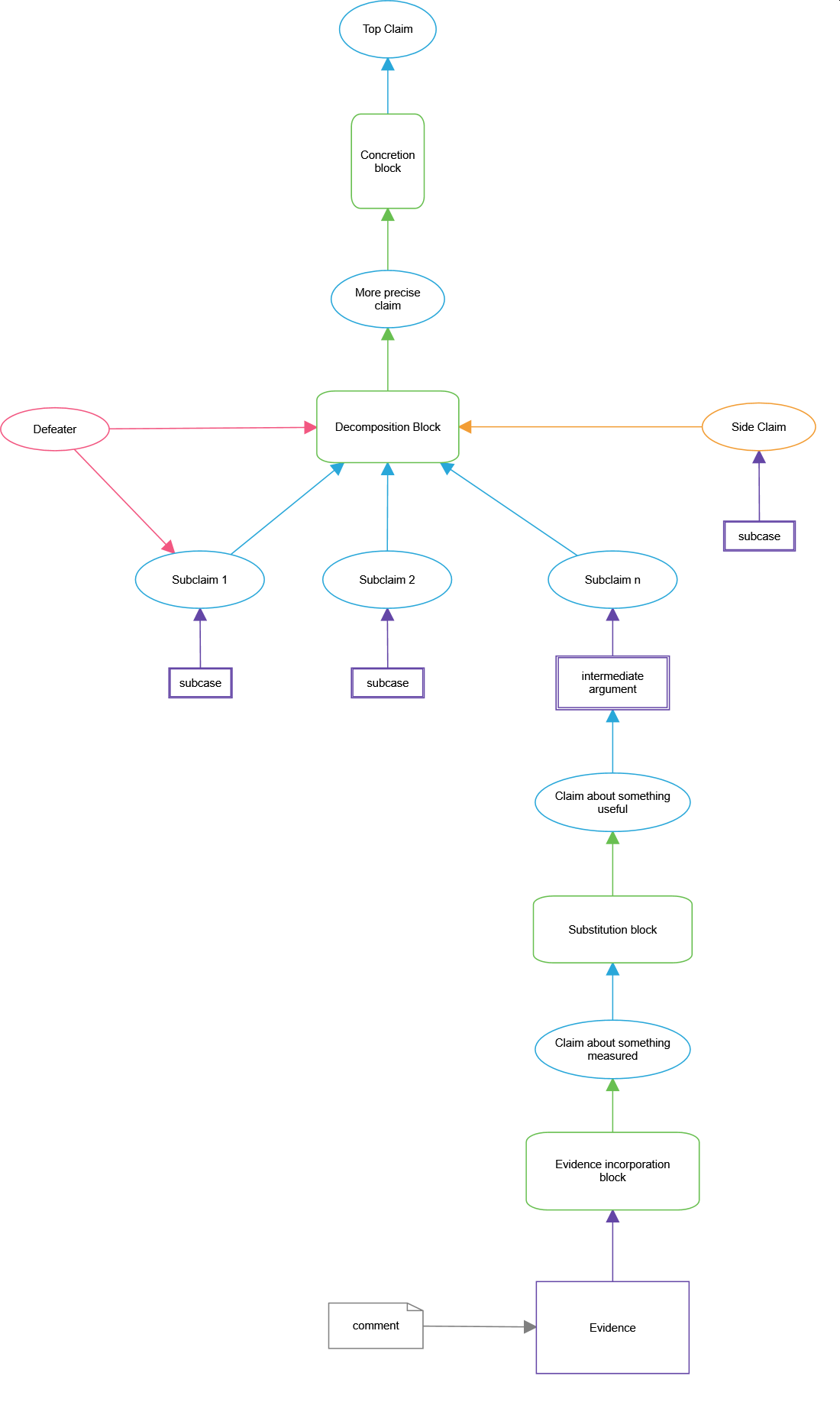}
\end{center}
\vspace*{-5ex}
\caption{\label{busy}Example \cla\ Argument}
\end{figure}

For each of the four kinds of block that may appear as reasoning
steps, soundness requires that the narrative justification ``makes the
case'' that, given any side-claim, the subclaims indefeasibly entail
the parent claim.  Tool support for Assurance 2.0 could provide a
checkbox for each argument node that allows developers or assessors to
indicate their concurrence with the supplied justification, but in
\clasce\ we manage this rather differently.  The default assumption is
concurrence with the supplied justification, and any disagreement is
indicated by attaching a \emph{defeater} node to the suspect argument
(or other) node.  The reason for this choice is that defeaters make a
claim (e.g., ``this justification overlooks the possibility
that\ldots'') that can be confirmed or refuted in its own subcase,
which can be retained to assist later developers and assessors who may
have similar doubts.  The presence of defeaters and the associated
introduction of refutational reasoning complicates assessment of
validity and this is discussed in a companion report
\cite{Bloomfield-etal:defeaters23}.

Assessment of evidence incorporation blocks needs first to assess
whether the evidence supports the claim made by the block; usually,
this claim should state ``something measured'' that is close to what
the evidence \emph{is} (e.g., ``we performed requirements-based
testing and achieved MC/DC coverage'').  Then we need to evaluate
whether that claim justifies a further claim that states ``something
useful'' (i.e., what the evidence \emph{means}) with sufficient
``weight'' to be accepted as a premise in the NLD interpretation.  The
second claim is usually related to the first via a substitution block
and, to assess the weight of evidence in a principled way, we use the
measures of \emph{confirmation theory}.  These are based on (possibly
informal) subjective probabilities indicating, for example, the extent
by which our prior confidence in the claim is increased when provided
with the evidence.  These valuations are discussed in Section
\ref{confirmation}.

For soundness, we merely require the weight of evidence, as assessed
by confirmation measures and possibly other, more informal
evaluations, to cross some threshold, so that the assertion that the
evidence supports its claim can be considered a justified premise.
But we can also use probabilistic measures to indicate \emph{how
confident} we are in this assessment.  These probabilistic assessments
can then be aggregated upwards from evidence through the reasoning
steps of the argument using various kinds of probability logic to
yield what we call \emph{probabilistic valuations} for the claims (in
distinction to soundness, which is a \emph{logical valuation}),
including the top claim.  Advocates of several different methods for
assessment of assurance cases treat their preferred probabilistic
valuations as \emph{the} assessment of a case; Graydon and Holloway
show these treatments are problematic
\cite{Graydon&Holloway:quant16,Graydon&Holloway:quant17}.  We apply
probabilistic valuations only to cases that have already been judged
valid and sound and use them as augmentations to those judgments, not
replacements.  Furthermore, our methods of probabilistic aggregation
are conservative and our assurance cases are limited to steps
comprised of just the five building blocks; together, these
restrictions eliminate many of the sources of difficulty exposed by
Graydon and Holloway.  The main value that we see in probabilistic
valuations is that they assist in apportioning effort across a case
(e.g., we typically want similar confidence throughout a case, with
possible exceptions for areas explicitly considered to pose greater or
lesser risk), and they can help with \emph{graduated assurance}.  This
arises in many types of system where we wish to balance effort and
confidence for entire subsystems according to risk and is exemplified
by the DALs of DO-178C for airplane software, and by the ESILs of ISO
26262 for automobiles.

An alternative to probabilistic valuations constructed by external
examination of the case is to include probabilistic statements in the
claims.  This is particularly appropriate for claims concerning
properties such as reliability and for those based on probabilistic
evidence such as statistically valid random testing
\cite{Currit86,Parnas90}.  The claims will then be justified within
the argument by reference to some accepted theory.

\exmemo{Possible numerical confirmation of counterclaims might
constitute negative numerical confirmations, and one might then be
able to evaluate confirmation measures for interior claims.}

\paragraph{Negative Perspectives:} These record doubts about the case,
and their exploration and resolution.  Reviewers will naturally have
doubts and questions about an assurance case and it is sensible to
anticipate these and respond to them within the case.  In addition,
the developers of the case need to guard against \emph{confirmation
bias} \cite{Leveson11:JSS,Nimrod09}.  To ensure that the positive
perspective does not become an optimistic one, they should actively
propose and explore doubts and challenges as they develop the
case---and should record these, both to avoid rework, and to aid
reviewers.

We refer to any concern about a case as a \emph{doubt}; as we explore
the origin and nature of a doubt we will refine it to a
\emph{defeater}: that is, a specific (counter)claim or challenge that
can be attached to a particular point in the argument.  Investigation
of a defeater may lead to correction or improvement of the assurance
case, in which case the modifications become part of the positive case
and the defeater that motivated them may be mentioned only in the
narrative of some of its justifications.  Alternatively, the defeater
may, on investigation, be considered a ``false alarm''; the
investigation that reveals this can be represented as a (sub)case in
its own right, and may be recorded within the main case. \yy \clasce\
has functions for recording defeaters and for developing and recording
the subcases that lead to their resolution (we refer to this as
defeating the defeater), and these can be revealed or hidden as
desired.

\yy The number of resolved defeaters might be considered a measure of
the effectiveness or diligence with which the negative perspective has
been explored \cite{Weinstock:Baconian13}; this can be subject to
gaming (e.g., using trivially different defeaters to raise the number
of resolutions), but can be effective when undertaken with care as
``eliminative argumentation''
\cite{Goodenough-etal:TR2014,Diemert&Joyce2020}.

A defeater to an assurance case is rather like a hazard to a system:
that is a conjecture why things might go wrong.  Hazard analysis is
not judged by how many hazards are found, but by the historical record
and rational justification of the method used, and the diligence of
its performance.  We think similar evaluations should be applied to
the negative perspectives of an assurance case and so we advocate use
of systematic (and potentially automated) methods for discovery of
defeaters.  These include methods based on Answer Set Programming
\cite{Murugesan-all:GDE23} and application of ``knowledge bases''
concerning historical flaws in systems \cite{Driscoll:dashlink} and
fallacies in assurance cases \cite{Greenwell-etal06} (although the
strict requirements of Assurance 2.0 cases are intended to exclude
most of the latter ``by construction'').  Because Assurance 2.0
arguments are composed of only five types of (building) blocks, it is
feasible to enumerate candidate defeaters for each type and apply
these systematically, and potentially automatically, to assurance
cases as they are developed.  We discuss methods of defeater discovery
in forthcoming report \cite{x}.

\paragraph{Residual Doubts:} The world is uncertain so not all doubts
can be resolved.  For example, we may design a system to tolerate any
two faults and have good reasons and evidence to suppose that is
sufficient to cover the exposure on any expected mission.  But doubts
remain: what if more than two faults do arrive?  Or, for a security
perspective, what if advances elsewhere render our cryptographic key
length insufficient?  In these cases we can acknowledge the existence
of a valid defeater to our case and then explore their consequences
and likelihoods and thereby assess \emph{risk} (their product).  Some
of these \emph{residual risks} may be unacceptable and thereby prompt
a review, but others may be considered acceptable or unavoidable.  It
is crucial however that these judgments are conscious ones and that
they are recorded in the assurance case.

We now develop valuations and measurements for each of the
perspectives introduced above.

\exmemo{The measures and assessments that support the various perspectives
used in evaluating an assurance case are bottom-up aggregations that
start with evidence and work step by step upward through the argument
of the case to its top claim.  At each step, the aggregated measure of
its claim is calculated as some combination of the aggregated measures
of its supporting subclaims or evidence.  The form of this combination
may depend on the type of block used in that step.

So many sections structured into subsections by block.}

\newpage
\section{Positive Perspectives: Logical Valuation (Soundness)}
\label{soundness}

\exmemo{Consider (logical) soundness, (probabilistic) valuation,

all contributing to overall confidence in indefeasible justification.}

Recall that our fundamental requirement is that an assurance case
should provide \emph{indefeasible justification} for the decision to
deploy the system or service concerned; this means that the
justification must be so well supported, and all plausible doubts and
objections must be so thoroughly considered and countered, that we are
confident there are no credible doubts remaining that could change the
decision.

Confidence in the investigation and resolution of doubts and
objections is discussed in Section \ref{defeaters}, and confidence
that any remaining as residual doubts pose insignificant or
manageable risks is discussed in Section \ref{residuals}.  During
development, we will probably attempt to identify and resolve doubts
before proceeding to assessment from a positive perspective.
Nonetheless, we present the positive perspective first, because it
aligns with the most basic goal of an assurance case argument.

So, in this section and the next we are concerned with positive aspects
of the case: namely that its evidence and argument justify confidence
in its claims.  In the next section, we consider these positive
aspects from a probabilistic point of view, while in this section we
consider a logical point of view, which we call \emph{soundness}.

Soundness was introduced in Section \ref{structure} and is discussed
in the paper on Assurance 2.0 \cite{Bloomfield&Rushby:Assurance2} and
its more theoretical precursors \cite{Rushby:Shonan16,Rushby:AAA15}
that together provide the basis for Assurance 2.0.  Briefly, soundness
interprets the argument of an assurance case as an informal proof
based on \emph{Natural Language Deductivism} (NLD).  This means that
the leaf steps of an assurance case argument are interpreted as
\emph{premises} in which evidence establishes a claim, and the
interior steps are interpreted as \emph{axioms} in which a conjunction
of (sub)claims implies a parent claim (see Figure \ref{busy} for an
example).  The full argument is then a tree of claims (possibly with
cross links if a (sub)claim is used in support of multiple claims)
whose top claim is the conclusion of the assurance case.

We need to be sure that evidence really does establish its claim, and
that subclaims really do imply their parent, so each argument step is
supplied with a \emph{justification} why this is so.  The
justification will typically be a narrative in natural language but it
may reference some external theory, calculation, proof, or mechanized
analysis etc.

The benchmark for an interior step is that the conjunction of its
subclaims deductively implies or entails its parent claim, given any
side-claim (so that each interior step is what logicians call a
``definite clause'').  Doubts about the deductivism of a step---that
is, a suspicion there is ``something missing'' among the subclaims, so
that their conjunction merely ``suggests'' rather than entails the
parent claim---causes a step to be considered ``inductive'' rather
than deductive.\footnote{There is a stronger form of doubt, where the
argument step is considered wrong rather than merely incomplete.  This
case is discussed in Section \ref{residuals}.}

We now consider how confidence in deductivism can be established.

\subsection{Confidence in Deductiveness (Interior Steps)}
\label{interior}

Free-form arguments often challenge their developers with the
``bewilderment of choice''; in contrast, \cla\ arguments are built from
a limited repertoire of just five different argument blocks, of which
four apply to interior steps.  Selection of an appropriate block is
assisted by the ``helping hand'' pictorial memory aid shown in Figure
\ref{helping-hand}.

\begin{figure}[ht]
\begin{center}
\includegraphics[width=\textwidth]{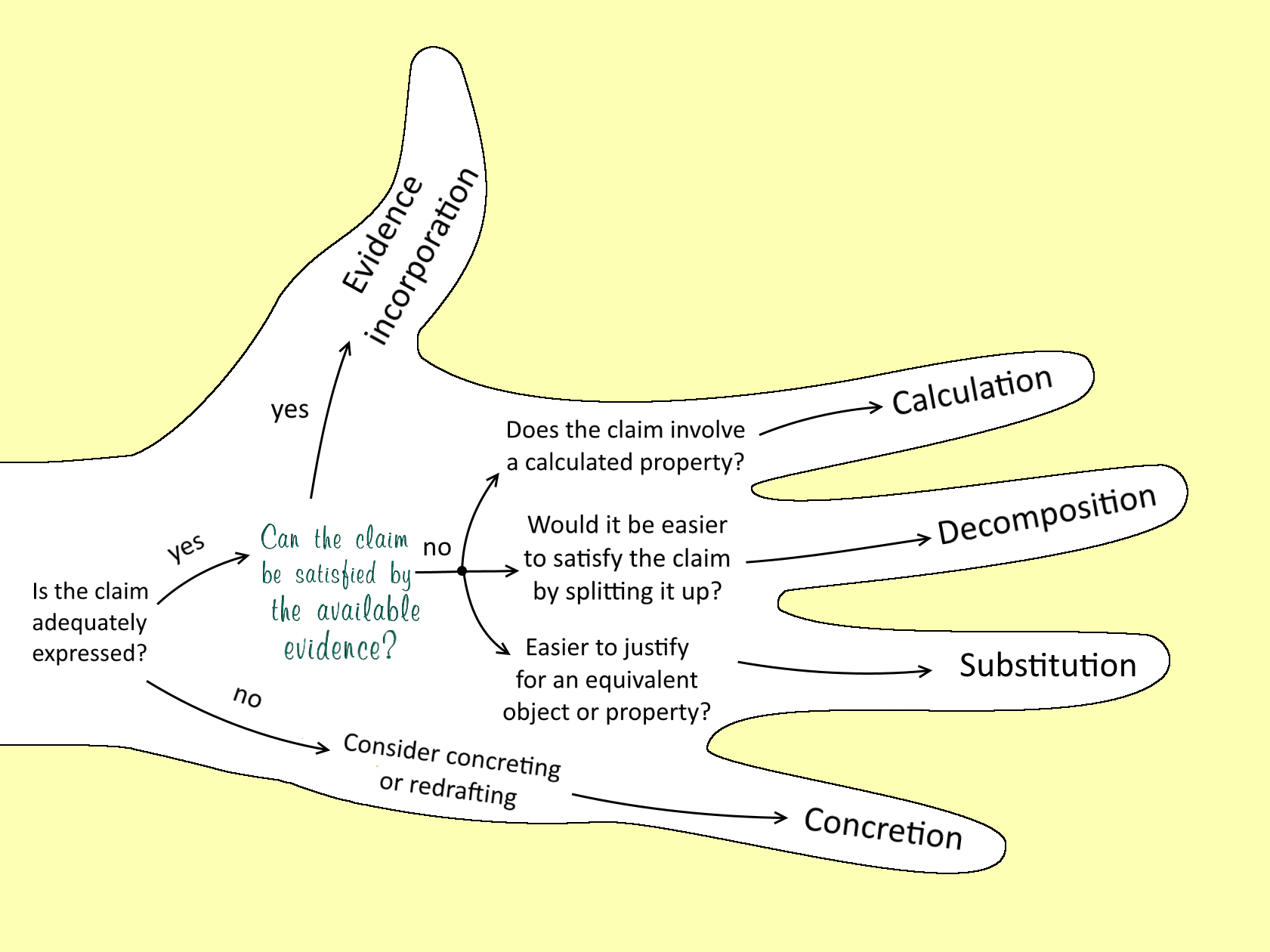}
\end{center}
\caption{\label{helping-hand}The Helping Hand Memory Aid}
\end{figure}

\begin{figure}[ht]
\begin{center}
\includegraphics[width=4in]{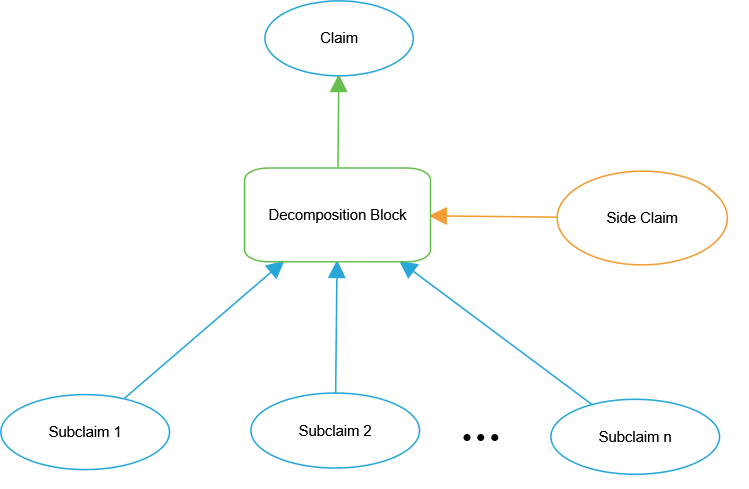}
\end{center}
\vspace*{-3ex}
\caption{\label{decomp-block}Generic Decomposition Building Block}
\end{figure}

As illustrated in Figure \ref{decomp-block},
Assurance 2.0 blocks generally have a \emph{side-claim}, which is a
(conceptually distinct) subclaim whose purpose is to ensure that the
other subclaims are well formed and really do deductively entail the
parent claim.  The relationship is then
$$ \mbox{side-claim} \supset (\mbox{conjunction of other subclaims} \supset
\mbox{parent claim}).\footnote{We use $\supset$ for
material implication, $\wedge$ for conjunction, $\vee$ for
disjunction, and $\neg$ for negation.}$$
This is logically equivalent to
\begin{equation}
\label{decomp-logic}
(\mbox{side-claim} \wedge \mbox{conjunction of other subclaims}) \supset
\mbox{parent claim}
\end{equation}
and so we see that although the side-claim has a conceptually
distinct status, it is logically no different from the other subclaims.

Every branch in a complete assurance case argument must terminate in
evidence (or, exceptionally, a calculation).  \yy Assurance case
arguments or subarguments that have unsupported claims as in Figure
\ref{decomp-block} are incomplete and unfinished.
Some assurance case notations have a symbol that explicitly indicates
unfinished parts of the case; \clasce\ can use subcase nodes for this
purpose, as illustrated in Figure \ref{busy}, but does not require it,
regarding unsupported claims as sufficient indication of an unfinished
branch.\footnote{\yy Unsupported claims can also be marked as
\emph{assumptions}, in which case the branch is considered complete.}

The argument step of Figure \ref{decomp-block} illustrates a decomposition
block, whose purpose is to divide a claim into subclaims over some
explicit enumeration, such as components of the system, or time (e.g.,
past, present, future), or hazards, and so on.  In each case, the
side-claim must establish that the decomposition is complete and
satisfies any other properties that may be needed, such as that the
claim distributes over components, or that some theory justifying the
decomposition is properly applied.  For example, if the decomposition
is over hazards, then the side-claim will require that all hazards
have been identified and that the decomposition considers them all,
both individually and in combination; such a side-claim might be
discharged by evidence that attests to use of a well-accepted method
of hazard analysis, performed diligently.

\begin{figure}[ht]
\begin{center}
\includegraphics[width=4.5in]{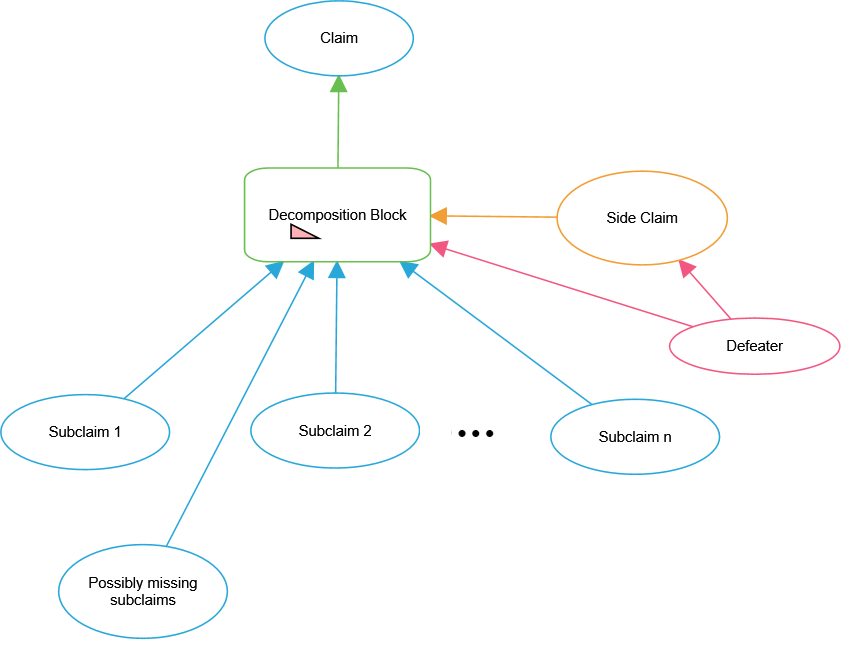}
\end{center}
\vspace*{-3ex}
\caption{\label{decomp-doubts}Three Ways to Indicate Doubt in an
    Argument Step}
\end{figure}

\yy Now it may be that we have doubts about the decomposition (and
therefore also about justification for its side-claim) and wish to
indicate this in the developing assurance case so that we can return
to it later.  One approach would be to annotate the argument step as
inductive.  This can be indicated informally by adding a marker to the
argument step as illustrated by the triangle in Figure
\ref{decomp-doubts}; \clasce\ allows this, but does not provide an
interpretation for it at present.  An alternative is to use an
explicit \emph{defeater} node; the defeater can point to the
argument node or to a specific subclaim to localize the perceived
source of doubt, or to both as illustrated on the right of Figure
\ref{decomp-doubts}.  A third alternative is to add an unsupported
subclaim (equivalently, an uninterpreted assumption) acknowledging
something is ``possibly missing'' as shown at the bottom of Figure
\ref{decomp-doubts}.  The three options illustrated in this figure are
alternatives: we should select one.

Identified defeaters must be investigated and resolved before a case
can be considered complete, but it is not required to perform this
resolution before considering other parts and other aspects of the
case.  If the doubt motivating a defeater is found to be unwarranted,
then it is said to be a ``defeated defeater'' and the justification
and possible subargument for this assessment will be recorded as part
of the case, rather like a comment, and selectively revealed or hidden
as will be described in Section \ref{defeaters-in-cl}.  If, on the other
hand, the defeater identifies a legitimate doubt, then the missing or
incorrect claim(s) must be discovered and the argument step (and the
subarguments that justify it) suitably amended.

\begin{wrapfigure}{r}{.5\textwidth}
\vspace*{-1.5ex}
\begin{center}
\includegraphics[width=0.45\textwidth]{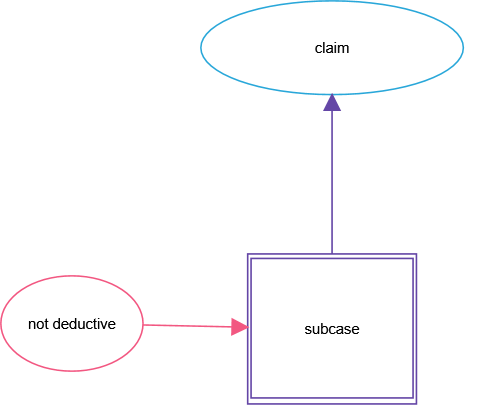}
\end{center}
\vspace*{-3ex}
\caption{\label{repair-start}Defeater to a Generic Subcase}
\vspace*{-2.5ex}
\end{wrapfigure}

Missing and suspected missing subclaims are easily added to
decomposition steps as shown by the ``possibly missing'' subclaim at
the bottom of Figure \ref{decomp-doubts}, which will either be removed
or be replaced by the truly missing subclaim(s) as the doubt is
resolved.  This is not so straightforward when the doubt concerns
other kinds of argument steps because these take only a single
subclaim.  An example is illustrated in Figure \ref{repair-start}
where a ``not deductive'' defeater is aimed at a subcase (reduced to a
single generic node for simplicity).
Suppose we determine that the defeater is legitimate and that an additional
subclaim is needed.  There are two approaches: we can add this
subclaim below the original argument, as shown on the left of Figure
\ref{repair}, or above it, as shown on the right of that figure.

\begin{figure}[ht]
\raisebox{16ex}{\hfill\includegraphics[width=0.46\textwidth]{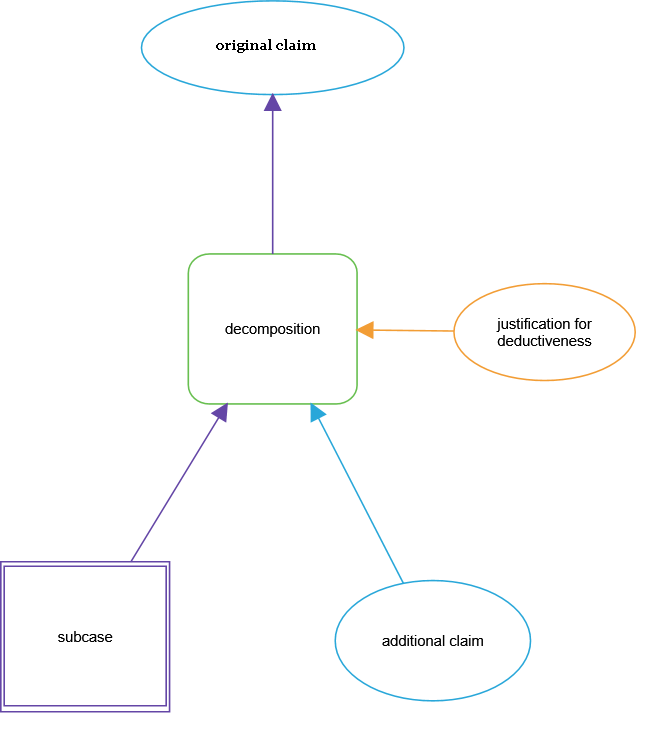}}
\hfill\includegraphics[width=0.46\textwidth]{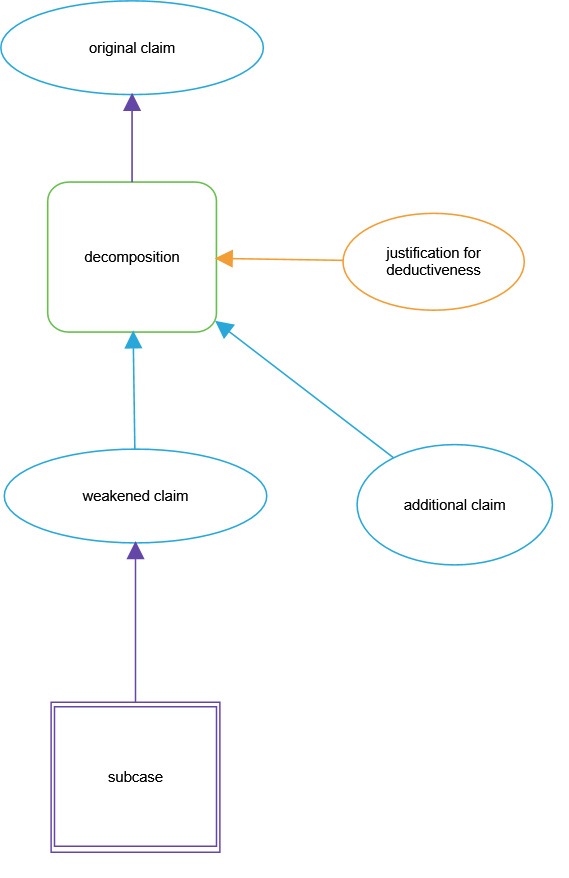}\hfill
\caption{\label{repair}Additional Claim Added Below, or Above
    Original Argument}
\end{figure}

Notice that when the additional claim is added above the existing
argument, we must recognize that only a weakened form of the original
claim is supported by the existing subcase.  Apart from this
adjustment, however, the original case is augmented rather than
changed, and this may be considered an advantage for the ``added
above'' choice.

\subsubsection{Embedded Links}

The argument of an assurance case is required to satisfy certain
structural properties.  For example, claims cannot link directly to
claims (there must be an intermediate argument node), and the pattern
of links must be non circular.  These requirements are easily checked
and are also visibly obvious when all links are explicit, as we have
assumed until now.  However, the nodes of an Assurance 2.0 argument
must each be supplied with narrative descriptions of their
interpretation or justification, and in \clasce\ these may contain
embedded links to other nodes.

\begin{figure}[ht]
\begin{center}
\includegraphics[width=\textwidth]{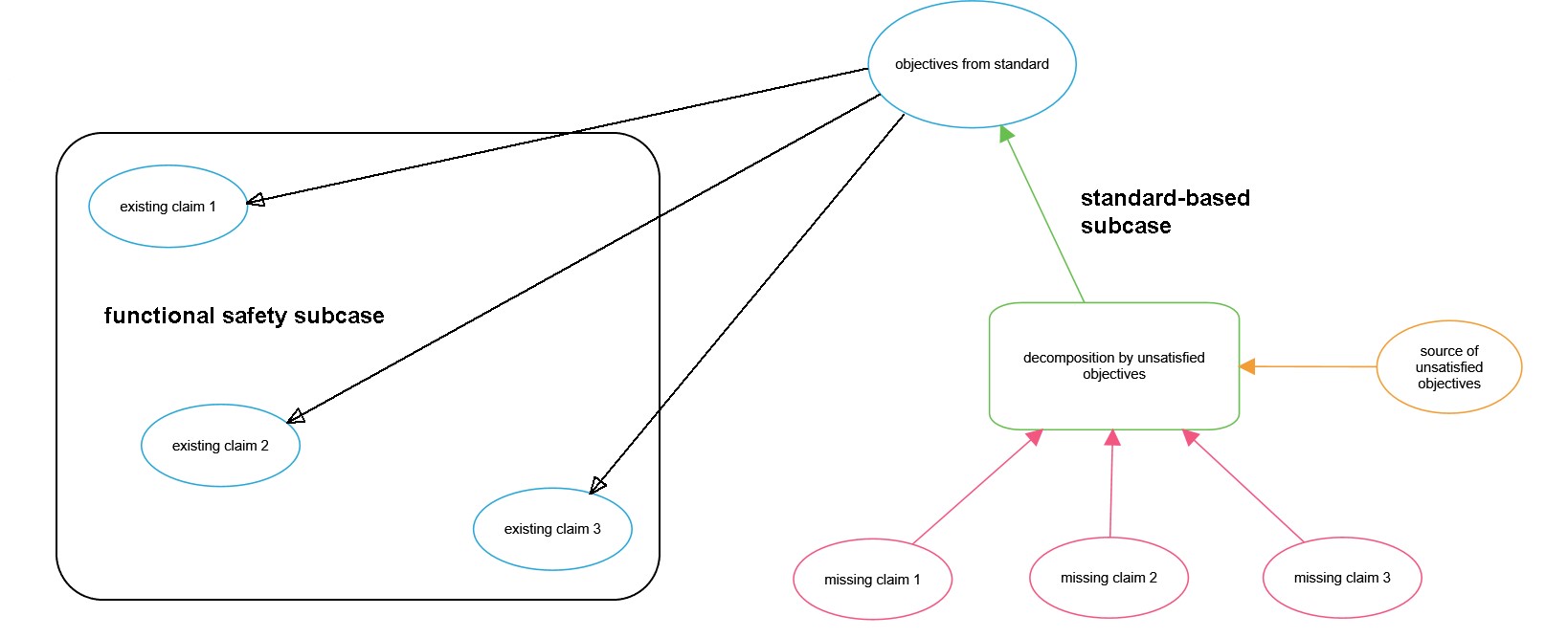}
\end{center}
\vspace*{-3ex}
\caption{\label{embedded-links}Embedded Links}
\end{figure}

These links serve a different, narrative, purpose than conventional
links, whose purpose is logical.  Consequently, they do not
participate in logical (i.e., soundness) or probabilistic valuations
of the argument.  For example, in an assurance case for an autonomous
drone, we may have a primary case that argues it is functionally safe
in the style of ``Overarching Properties'' (OPs)
\cite{Overarching18,Holloway:OP19}, and a secondary case that argues
it complies with a relevant standard called F3269-17 \cite{F3269-17}.
The narrative for the node making the latter claim may contain a table
of all the objectives required by the standard and, for each of them,
an embedded link to an existing node in the case where that objective
has been satisfied as part of the OP subcase.  \yy \clasce\ is able
selectively to display such links or not; they are shown displayed (in
gray) in Figure \ref{embedded-links}.  Note that the arrows are
reversed compared with conventional logical links as an indication of
their different purpose.

In this example, the secondary case can also serve as a source of
potential defeaters.  In particular, those objectives from the
F3269-17 standard that are not satisfied by the OP subcase can be
added to the argument as defeaters as shown on the lower right of
Figure \ref{embedded-links}.  The interpretation of assurance
arguments with defeaters (see Section \ref{defeaters-in-cl}) ensures
that the local top claim cannot be considered true while any of these
defeaters is present.  As each of the missing objectives is satisfied,
its defeater can be replaced by a positive claim.  If (or when) all
the F3269-17 objectives are satisfied, the secondary case can be used
to increase confidence in the primary case.

\exmemo{Notice that here we use defeater nodes for the missing objectives, as
opposed to the ordinary claim node used in Figure \ref{decomp-doubts}.
As we will see in Section \ref{defeaters}, resolution of a legitimate
defeater begins by replacing it with (or interpreting it as) a claim,
so Figure \ref{decomp-doubts} can be considered to have skipped a step
compared to Figure \ref{embedded-links}.  

I no longer understand the above defeater/claim discussion: need
to look at Section \ref{defeaters}.}

\exmemo{To be continued.  Needs an example such as static analysis.}

\exmemo{Defeater could also be attached to the side-claim...logically no
different but may permit a divide and conquer approach...main case
proceeds OK, defeater quarantined in the side-claim}

\exmemo{Basic case establishes positive argument for top claim, but we
are also interested in what negative evidence was considered and how
it was dismissed.  And residual doubts.  So 3 things.

We want indefeasibility.

Not only true, but all doubts addressed.

Logical soundness or just soundness is all premises true and all
    reasoning steps valid

Interpreted as weight of evidence above threshold and reasoning steps
    are definite clauses

It's only the positive aspects

confidence is the full thing, with negatives taken into account

But then we have to address doubts:

doubt is general unease, unspecific

special cases: inductive steps, nugatory evidence

becomes specific as a defeater

need to deal with these if valid, or defeat them if not

or let them remain as acknowledged doubts

Reasonable doubt, Blackstone criterion, articulable

@article{pi2019quantifying,
  title={Quantifying Reasonable Doubt},
  author={Pi, Daniel and Parisi, Francesco and Luppi, Barbara},
  journal={Rutgers UL Rev.},
  volume={72},
  pages={455},
  year={2019},
  publisher={HeinOnline}
}

want to know how likely, worst case impact, hence risk

can evaluate individually, or how they combine:
loads of small risks remain small, or they become significant, or we
    need to model them probabilistically

What about quantification: relationship to DALs and SILs?

Can propagate the ingredients of confirmation measures: $P(E \vbar C)$
and $P(E \vbar \neg\, C)$... explain how these go for each kind of
    block.

Also see if doubts also propagate in some way over blocks.

}

\exmemo{Another use for embedded links is in referencing the
application of a theory and its template to an argument subcase, as
described in the following subsection.

Is this still true?}

\subsubsection{Theories and Templates}
\label{templates}

We introduced the notion of \emph{theories} early in this report, in
Section \ref{theories}; the idea is that a theory provides
justification for an argument step or series of steps, or for a
complete subcase: each of the argument steps in a theory-supported
subcase will cite some aspect of the theory as its justification and
any side-claims will likewise reference aspects of the theory, as will
interpretation of the evidence supplied.

The linkage just described between theory and subcase is informal and
we may have differently structured subcases supported by the same
theory.  On the other hand, we can provide more automated assistance
if the theory explicitly provides a generic subcase that can be
instantiated to yield the specific subcase required in the context
concerned.  Such generic subcases have much in common with what some
other tools for assurance cases call ``templates,'' and we will also
use this term.  However, unlike other methodologies and tools, we see
a template as a representation of the full argument for a subcase,
including its narrative justifications and theoretical underpinnings,
whereas others see it as just syntax (i.e., a subdiagram) and the
instantiated template becomes simply another part of the overall case
and must be justified along with the rest of the case.
Metaphorically, we see templates as ``subroutines'' whereas others see
them as ``macros.''

By instantiate, we mean that the generic template will contain
placeholders that behave like formal parameters in a programming
language: for example, a theory to ``establish correctness of CLEAR
requirements using CLEAR tools'' may have a generic top claim
``requirements for \$x\$ are correct,'' where \$x\$ indicates a
parameter.  When this is applied to a claim ``requirements for
ArduCopter AFS are correct,'' the parameter will be instantiated as
``ArduCopter AFS,'' with similar substitutions elsewhere in the
template.\footnote{CLEAR is a tool-supported requirements notation
from Honeywell \cite{Bhatt:CLEAR18} and AFS stands for ``Advanced
Failsafe Monitor,'' which is a safety monitor for an
autonomous quadcopter (``ArduCopter'') being developed as part of the
ARCOS program by the DesCert team \cite{DesCert:phase1-arxiv,Bhatt:FM22}.}

When theories provide templates, we can contemplate tool support that
uses templates a) to check manually constructed theory-supported
subcases, b) to construct those subcases automatically by
instantiating templates under human direction, and c) to synthesize
automatically parts of an assurance case by searching for applicable
theories and templates and instantiating them appropriately.  \clasce\
provides prototype support for the third of these and, by constraining
its application, it can also perform the second.  These synthesis
capabilities are described in a draft report
\cite{Bloomfield:CLARISSA-SA}.  Here, we merely sketch some ways in which
manual or automated synthesis impacts assessment of an assurance case.

Most significantly, the overall assessment of soundness, and
confidence therein, can be ``lifted'' from consideration of the
individual steps of the assurance argument to the selection and
arrangement of the theories employed and the ``glue logic'' that
weaves them together.  Thus, whereas assessment of a bespoke assurance
case argument must consider every node in the argument, those parts
instantiated from the template of a ``pre-certified'' theory need not
be reexamined.  However, the assurance case under development does
need to check the provenance of any theory that it uses and must
justify suitability of the theory to the context of its use, and also
the relevance and weight of any evidence that it provides to the
theory.  The developers of the theory can anticipate some of this by
attaching side-claims to their template.  Then, when a theory/template
is applied, we will eventually need to discharge the instantiations of
its side-claims.  If we are unable to do so, this may indicate that
the selected theory/template was an inappropriate choice.  We would
like to learn this early, before effort has been wasted, so some
side-claims in the template may be considered \emph{preconditions};
that must be discharged before the theory/template may be applied.
For example, it would make no sense to apply the CLEAR theory for
correctness assurance to requirements that are not written in CLEAR,
so ``requirements for \$x\$ are written in CLEAR'' would be a
precondition for this example.

\exmemo{Older text: needs to be updated to mention synthesis, but is it
correct about manually-guided instantiations?

At present, \clasce\ provides the first of these and will soon provide
the second; the third is a topic of ongoing research.  To indicate the
place where a theory/template should be applied, we use an embedded
link (recall the previous subsection) to connect the theory/template
concerned to the claim that it will justify.  The terms appearing in
that claim are used to instantiate the generic template, which is then
used to check or construct the subcase supporting the claim concerned.

}

\subsection{Confidence in Evidential Support for Claims (Leaf Steps)}
\label{leaves}\label{confirmation}

\memo{Explain later that confirmation can be used for nondeductive
steps.  Not sure I understand this.}

The argument of an assurance case is logically valid when all its
claims are ultimately supported by evidence or assumptions and the
claims supporting and delivered by interior reasoning steps ``match
up'' to provide a coherent and connected graph structure.  \yy We say
the argument is \emph{fully valid} when, in addition, all its steps
are (judged to be) deductive, and there are no unresolved defeaters
other than those marked as residual doubts.  It is not important in
what order logical validity, deductiveness, and resolution of
defeaters are tackled and assessed for full validity.  A fully valid
argument is \emph{sound} when human evaluators attest that the
residual doubts are negligible (see Section \ref{residuals}), that the
justification for each interior step is satisfactory, and that the
weight of each evidential step is sufficient to justify its claim, as
we now describe.

\exmemo{The notion of confirmation relates to single hypotheses.  From a
Bayesian perspective, it has to do with the ways in which, and the
degree to which, belief in a hypothesis is reasonable.

Whether data constitute evidence, on the other hand, has to do with
the ways in which they serve to distinguish and compare competing
hypotheses.  It is a three-part relation involving data and two
hypotheses.  Data that cannot tell for or against such hypotheses do
not constitute evidence for one or the other.  A natural way to express
this very basic intuition is through the use of likelihood
ratios.  Thus, data D constitute (positive) evidence for hypothesis H
just in case the ratio of likelihoods, $Pr(D|H)/Pr(D|H')$ is greater
than 1, where H and H' are not necessarily mutually exclusive.  If D is
equally likely on H and its competitors, then D does not constitute
evidence for any of them.  The point needs to be emphasized: data by
themselves do not constitute evidence, but only in a context provided
by competing hypotheses \cite{Bandyopadhyay-etal16}.

}

In the logical valuation of an assurance case by NLD, we interpret
evidence incorporation steps as premises.  For soundness we therefore
require strong justification that the evidence supplied to the step
really does support its claim.  Unlike other assurance steps, which
inhabit the world of logic, evidence incorporation steps are bridges
to that world from the world of observation and measurement.
Therefore we cannot assess evidence incorporation steps by the methods
of logic, we need the methods of epistemology.  Epistemology is about
justified belief (as an approximation to truth) and it is natural to
express the strength of our confidence in a belief as a number.  We
will expect those numbers to obey certain rules (the Kolmogorov
Axioms) and so they function as (subjective) probabilities
\cite{Jeffrey04}.

A natural measure of confidence in a claim $C$ given the evidence $E$
is the subjective posterior probability $P(C \vbar E),$ which may be
assessed numerically or qualitatively (e.g., ``low,'' ``medium,'' or
``high'').  However, confidence in the claim is not the same as
confidence that it is justified by the evidence.  It is possible that
the reason for a high valuation of $P(C \vbar E)$ is that our prior
estimate $P(C)$ was already high, and the evidence $E$ did not
contribute much.  So it seems that to measure justification we ought
to consider the difference from the prior $P(C)$ to the posterior $P(C
\vbar E)$ as an indication of the ``weight'' of the evidence $E$\@.
Difference can be measured as a ratio, or as arithmetic difference,
thereby producing the following two \emph{confirmation measures}, due
to Keynes in 1921 and Eells in 1982, respectively.\footnote{The
logarithm (which may use any positive base) in Keynes' and other ratio
measures serves a normalizing purpose so that, as with arithmetic
difference, positive and negative confirmations correspond to
numerically positive and negative measures, respectively, and
irrelevance corresponds to a numerical measure of zero.}
\begin{eqnarray*}
\mbox{Keynes}(C, E) & = & \log\frac{P(C \vbar E)}{P(C)},\\[1ex]
\mbox{Eells}(C, E) & = & P(C \vbar E) - P(C).
\end{eqnarray*}

There are many other confirmation measures proposed in the literature
\cite{Tentori-etal07}.  For example, some prefer to use the likelihood
$P(E \vbar C)$ rather than the posterior $P(C \vbar E),$ because it is
generally easier to estimate the probability of concrete observations
(i.e., evidence), given a claim about the world, than vice-versa,
thereby giving us the likelihood variants of Keynes' and Eells'
measures:\footnote{Notice that $\mbox{L-Keynes}(C, E) =
\mbox{Keynes}(C, E)$
and $\mbox{L-Eells}(C, E) = \mbox{Eels}(C, E)\times\frac{P(E)}{P(C)},$
since Bayes' Theorem gives $P(C \vbar E) \times P(E) = P(C
\wedge E) = P(E \wedge C) = P(E \vbar C) \times P(C)$.}
\begin{eqnarray*}
\mbox{L-Keynes}(C, E) & = & \log\frac{P(E \vbar C)}{P(E)},\\[1ex]
\mbox{L-Eells}(C, E) & = & P(E \vbar C) - P(E).
\end{eqnarray*}
We can see that these will tend toward their maxima when $P(E)$ is
small, meaning that $E$ should be unlikely in general.  This suggests
that we should favor evidence whose occurrence (in the absence of $C$)
would be a \emph{surprise}.

Similarly, if we have accepted evidence $E_1$ and seek additional
evidence, we should look for $E_2$ that is (or remains) surprising in
the presence of $E_1.$  Thus, for example, if $E_1$ is evidence of
successful tests, it will not be surprising if additional tests are
successful; instead we should seek evidence $E_2$ that is ``diverse''
from $E_!,$ such as static analysis.  More formally, we have, by the
chain rule
\begin{eqnarray*}
P(C \wedge (E_2 \wedge E_1)) & = & P(C \vbar E_2 \wedge E_1) \times
    P(E_2 \vbar E_1) \times P(E_1), \mbox{\ and}\\
P(E_2 \wedge (C \wedge E_1)) & = & P(E_2 \vbar C \wedge E_1) \times P(C \vbar E_1) \times P(E_1).
\end{eqnarray*}
The left (and hence right) hand sides are equal, and so
\begin{equation}
\frac{P(C \vbar E_2 \wedge E_1)}{P(C \vbar E_1)} =  \frac{P(E_2 \vbar
    C \wedge E_1)}{P(E_2 \vbar E_1)}.\label{diverse}
\end{equation}
Thus, $E_2$ delivers the largest ``boost'' to Keynes' measure for the
justification provided by $E_1$ (i.e., the left hand side of
(\ref{diverse})) when $E_2$ would be surprising given only $E_1,$ but
not when given $C$ as well, which confirms that $E_2$ should be
\emph{diverse} from $E_1.$
These observations about ``surprising'' and ``diverse'' evidence are
intuitively natural, but it is satisfying to see them put on a
rigorous footing.

An additional consideration when evaluating evidence is that
it is not enough for the evidence to
support a given claim $C,$ it should also discriminate between that
claim and others, and the negation, or ``counterclaim'' $\neg\, C$ in
particular.  Again, discrimination or distance can be measured as a
ratio or as arithmetic difference, leading to the following two
measures; the first is due to Good (and Turing) from codebreaking
activities during World War 2:
\[\mbox{Good}(C, E) = \log\frac{P(E \vbar C)}{P(E \vbar \neg\, C)},\]
and the second is due to Kemeny and Oppenheim in 1952:
\[\mbox{KO}(C, E) = \frac{P(E \vbar C) - P(E \vbar \neg\, C)}{P(E
\vbar C) + P(E \vbar \neg\, C)}.\]

We will refer to these as ``Type 2'' confirmation measures, and the
previous examples as ``Type 1.''  Likelihoods are related to
posteriors by Bayes' Rule, and appearances of $P(\neg\, x)$ in Type 2
measures can be replaced by $1 - P(x)$, so
\begin{eqnarray*}
\mbox{Good}(C, E) & = & \log\frac{P(C \vbar E) \times P(E) / P(C)}
{P(\neg\, C \vbar E) \times P(E) / P(\neg\, C)}\\
& = & \log\frac{P(C \vbar E) \times P(\neg\, C)}
{P(\neg\, C \vbar E) \times P(C)}\\
& = & \log\frac{P(C \vbar E) \times (1 - P(C))}
{(1 - P(C \vbar E)) \times P(C)}\\
& = & \log\frac{O(C \vbar E)}{O(C)}
\end{eqnarray*}
where $O$ denotes \emph{odds} (i.e., $O(x) = P(x)/(1-P(x))$) and
Good's measure is therefore sometimes referred to as the ``log odds''
or ``log odds-ratio'' measure for weight of evidence
\cite{Good:weight83}.

Similar manipulations can be performed on other Type 2 measures, so
that appearances of $\neg\, C$ revert to just $C$ and the distinction
between Type 2 and Type 1 measures disappears.  Furthermore, notice
that appearances of $P(E)$ cancel out in the second line of the
derivation above; manipulations of other measures generally exhibit
the same behavior and we find that they then satisfy the following
conditions:

\begin{enumerate}
\item They can be expressed as functions of $P(C \vbar E)$ and $P(C)$ only,

\item They are increasing functions of $P(C \vbar E),$ and

\item They are decreasing functions of $P(C).$

\end{enumerate}

Not all confirmation measures satisfy 1 above.  For example, the
following measure due to Carnap in 1962 \[\mbox{Carnap}(C, E) = P(C
\wedge E) - P(C)\times P(E)\] depends nontrivially on
$P(E).$\footnote{Notice that $\mbox{Carnap}(C, E) = \mbox{L-Eells}(C,
E) \times P(C),$ so the L-Eells measure does not satisfy condition 1
either.}  However, such measures can be manipulated by irrelevant
evidence \cite[section 2]{Atkinson12}, so we prefer measures that do
satisfy condition 1.

All confirmation measures indicate the extent to which evidence
justifies a claim, but they are not ordinally equivalent.  That is to
say, a given confirmation measure may rank one scenario (i.e.,
combination of $P(C \vbar E)$ and $P(C)$) higher than another, but a
different measure may do the reverse.  This is acceptable because,
although all confirmation measures evaluate the degree to which
evidence $E$ justifies claim $C,$ they do so in different ways and we
may prefer one measure to the other (or prefer different measures for
different purposes) \cite{Hajek-Joyce:confirmation08}.

Nonetheless, it is possible to add a fourth condition to the list above
and all measures satisfying this enhanced set of conditions are
ordinally equivalent.  This condition is the following
\cite{Atkinson12,Shogenji12}.
\begin{enumerate}
\setcounter{enumi}{3}
\item Let $C_1$ and $C_2$ be claims, unconditionally independent and
also conditionally independent on evidence $E.$  If both $C_1$ and
$C_2$ have measures of confirmation greater (resp.\ less) than $t$
then their conjunction must also have measure greater (resp.\ less)
than $t.$
\end{enumerate}
Measures that satisfy all four conditions are called
\emph{justification measures} \cite{Shogenji12}.  An example is
Shogenji's measure \[\mbox{Shogenji}(C,E) = 1 - \frac{\log P(C \vbar E)}{\log
P(C)}.\]

Shogenji argues that only justification measures serve to increase
true beliefs (e.g., claims) while reducing false ones
\cite{Shogenji12}, asserting that simple confidence measures (e.g.,
$P(C \vbar E)$) fail to do this and that traditional confirmation
measures may rank beliefs inconsistently.  In assurance, however, we
are usually seeking \emph{strong} indications that evidence justifies
a claim and the measures are likely to concur on this, so we are
generally content to use confirmation rather than justification
measures.  

\yy \clasce\ can attach confirmation measures to evidence
incorporation steps and it allows these steps to be marked as
``accepted'' when the weight of evidence (e.g., as indicated by an
attached confirmation measure) is judged to exceed some threshold.
Since the discussion above concludes that all confirmation measures
will deliver similar conclusions, it may seem that we could have
selected one and bypassed the discussion.  However, although the
conclusions may be similar, they are based on elicitation of different
judgments and we believe there can be value in asking those who assess
evidence to consider the different points of view underlying these
judgments.  For example, $\mbox{Keynes}(C, E)$ elicits judgments $P(C
\vbar E)$ and $P(C)$, while $\mbox{L-Keynes}(C, E)$ elicits $P(E \vbar
C)$ and $P(E)$.  Furthermore, the two measures should yield the same
value; we can therefore provide feedback to assessors if their
judgments are inconsistent.  Similarly, the original formulation of
$\mbox{Good}(C, E)$ elicits judgment of $P(E \vbar \neg\,C)$, which
requires consideration of a contrary point of view.

Some will be skeptical that human developers and evaluators are able
to assess and manipulate probabilistic measures correctly, even
qualitatively, and will also contend that confirmation measures are
beyond everyday experience.  They may point to alleged flaws in human
evaluation of probabilities.  Here is a standard illustration
\cite{Tversky&Kahneman:conjunction83}.
\begin{description}
\item[Evidence $E$\,:]
Linda is 31 years old, single, outspoken and very
bright. She majored in philosophy.  As a student, she was deeply
concerned with issues of discrimination and social justice, and also
participated in anti-nuclear demonstrations.
\end{description}
\noindent
The challenge is to assess which of the following two claims is most
probable, which we interpret as best
supported by the evidence.
\begin{description}
\item[Claim $C_1$:] Linda is a bank teller,

\item[Claim $C_2$:] Linda is a bank teller and active in the feminist movement.
\end{description}

When human subjects are exposed to this and similar examples, they
overwhelmingly favor $C_2.$ Psychologists label this the ``conjunction
fallacy'' because $C_2$ is the conjunction of $C_1$ with another
clause and a conjunction must always be \emph{less} probable than
either of its components; they then cite it as evidence for the
assertion that intuitive human reasoning is poor at probabilities
\cite{Tversky&Kahneman:conjunction83,Kahneman11}.  However, a more
recent interpretation is that humans evolved to weigh evidence and
actually base their judgments on mental measures more akin to
confirmation than simple probabilities (even when asked about probabilities)
\cite{Crupi-etal08,Shogenji12,Jonsson&Shogenji19}.\footnote{There are
larger claims, widely accepted in areas of psychology and cognitive
science, that much human perception and unconscious decision making
are based on processes akin to Bayesian analysis over models
\cite{Clark13,Friston:free-energy10,Friston12:history,Knill&Pouget04,Rao&Ballard99,Gregory:eye-brain}.}
To see this, we observe that the evidence $E$ seems to add nothing to
our prior belief in $C_1$ but it does seem to support the second
clause of claim $C_2$ (i.e., ``is active in the feminist movement'')
and so by item 2 of the list of properties for confirmation measures,
we conclude that the evidence indeed confirms $C_2$ more than $C_1,$
thereby refuting the ``fallacy'' charge.

Another area where human reasoning about probabilities is claimed to
be notoriously unreliable concerns the ``base-rate fallacy''
\cite{Bar-Hillel80}.  The standard examples involve diagnostic tests
for disease.  Suppose we have a test that is perfectly accurate at
diagnosing the disease when it is present (i.e., no false negatives),
but also has 10\% false positives.  A person tests positive in a
population where 1\% has the disease.  Human subjects are asked which
of the following probabilities is closest to the true probability the
person has the disease: a) 90\%, b) 10\%, c) 50\%, d) 89\%.

The correct answer is b) but human subjects overwhelmingly choose one
of the other answers, with an average of 85\% \cite[page
44]{Bandyopadhyay-etal16}.  The psychologists' explanation is that
humans overlook the very low base rate of the disease, which means
that false positives (10\%) overwhelm true positives (1\%).  (Others
would say it is because they do not know or do not apply Bayes' Rule.)
An alternative explanation again involves confirmation: the evidence
(positive test) increases the probability of the claim (having the
disease), so by item 2 of the list of properties for confirmation
measures, we obtain positive confirmation; human subjects opt for a
large number as a way of expressing this, despite being asked about
probabilities, not confirmation.\footnote{Few subjects will have
technical knowledge of probabilities or confirmation; the point is
that their intuitive reasoning is sound for many purposes, and uses
confirmation rather than probabilities.}  Although intuitive human
reasoning is again exonerated by supposing it employs (informally)
confirmation rather than probability, we would hope that developers
and evaluators of assurance cases are explicit in any choice between
probability and confirmation, and also apply Bayes' Rule in
circumstances like these (on representative numbers if only
qualitative estimates are being used).

A different example where population probabilities may confound
elementary reasoning is the ``Paradox of the Ravens'' \cite{Hempel45}.
Here, we seek evidence for the claim ``all ravens are black''; the
equally valid contrapositive of this claim is ``all non-black objects
are non-ravens'' for which a white shoe is produced in evidence,
allowing the triumphant declaration ``that proves it---all ravens
\emph{are} black!''

A rational escape from this ``paradox'' is \emph{Nicod's criterion}
\cite{Nicod30} that only observations of ravens should affect our
judgment whether all ravens are black.  More generally, claims about
some class of objects can be confirmed or refuted only by evidence
about those objects.  Under this criterion, we expect that
observations of black ravens would tend to confirm our claim, while a
non-black raven definitely refutes it.  Good, in a cleverly titled
one-page paper \cite{Good:red-herring67}, rebuts this expectation with
an example where observation of a black raven disconfirms the claim
``all ravens are black.''
\begin{quote}
Suppose that we know we are in one or other of two worlds, and the
claim under consideration is that all the ravens in our world are
black.  We know in advance that in one world there are a hundred black
ravens, no non-black ravens, and a million other birds; in the other
world there are a thousand black ravens, one white raven, and a
million other birds.  A bird is selected equiprobably at random from
all the birds in our world and turns out to be a black raven.  This is
strong evidence that we are in the second world, wherein not all
ravens are black.
\end{quote}

Examples such as this are challenging to philosophers seeking to
explain and justify the methods of science, but for assurance the
salient points are that we need to be skeptical about evidence (hence
consideration of alternative claims and counterclaims) and may need to
collect additional evidence to rule out alternative explanations.  (In
Good's example, observations of additional birds would allow us to
determine if we are in the world with a hundred ravens, or the one
with a thousand.)
Confirmation measures provide an attractive framework in which to
probe these issues and, far from being difficult for human evaluators,
they correspond to inbuilt human faculties for the weighing of evidence.

In the Linda example, claim $C_2$ entails the further claim ``Linda
works outside the home'' (since she is a bank teller), but the
evidence provides no direct support for this and it could easily be
false.  Thus, we have evidence that soundly supports a claim that
logically entails a further claim, yet that second claim could be
false.  For a more extreme example, the evidence that a card drawn
from a deck is an Ace supports the claim that the card is the Ace of
Hearts, and this entails the further claim that the card is red.  But
the card used in evidence could have been the Ace of Clubs, which
refutes the derived claim that it is red.  A pragmatic resolution to
this apparent paradox is that the standard for assurance should be
more demanding than basic confirmation: if an evidentially supported
claim is a conjunction (e.g., card is both Ace and Hearts), then we
need indefeasible support for all elements of the conjunction, and so,
in the context of assurance, we should not accept that the evidence
about Linda is sufficient to establish claim $C_2$ (because it does
not establish its $C_1$ conjunct).

\exmemo{Need to improve the above: consider counterclaim.  Both parts of
a conjunction is then a short cut.}

Another approach would be to propagate probabilities and likelihoods
from evidence through the directly supported claims to these
``second-level'' claims and to evaluate confirmation there.  In the
Linda and Ace of Hearts examples, we see that the evidence provides no
support for the second-level claims (i.e., ``works outside the home''
and ``card is red'').  These ``two level'' steps for exploiting
evidence are common in assurance cases: the bottom step, using an
evidence incorporation block, connects the evidence to a claim about
``something measured'' (e.g., ``we performed requirements-based
testing and achieved MC/DC coverage'') while the second step
(typically a substitution block based on application of an external
theory) connects it to a claim about ``something useful'' (e.g., ``we
have no unreachable code'').  In Assurance 2.0 and its \clasce\ tool
support, we recommend that evidence incorporation steps deliver
``evidentially measured'' claims (i.e., assessments of what the
evidence \emph{is}) that then supports a substitution step delivering
an \emph{evidentially useful} claim (i.e., an assessment of what the
evidence \emph{means} in this context) that is weighed using the ideas
of confirmation measurement.  We say ``ideas'' because we do not
require (although we do prefer) numerical or qualitative evaluation of
specific confirmation measures (and \clasce\ does provide an
experimental tool for experimenting with these) but only that the
narrative justification should consider the extent to which the
evidence increases positive assessment of the claim or distinguishes
it from other possible claims.

Sometimes a mismatch between the claims about ``evidentially
measured'' and ``evidentially useful'' leads to the realization that
one or the other is misstated.  For example, during World War
2, the US Army Air Force came to its Statistical Research Group in New
York seeking advice on where best to add armor to improve the survival
of their airplanes.  Many damaged planes returning from engagements
had been examined and this produced the following evidence.

\begin{center}
\begin{tabular}{|l|l|}
\hline
Section of plane & Bullet holes per sq. ft. \\
\hline
Engine & 1.11 \\
Fuselage & 1.73 \\
Fuel system & 1.55 \\
Rest of plane & 1.8 \\
\hline
\end{tabular}
\end{center}

The fuselage seems the most heavily damaged of the identified
components, so the evidence seems to support the claim ``the place
where armor will best improve survival of the plane is the fuselage.''
This is actually a second-level (``evidentially useful'') claim; we
should begin by using the evidence to justify a first-level
(``evidentially measured'') claim.  A plausible candidate for this is
``the fuselage is the part of the plane with heaviest damage.''  An
important difference in these two claims is that the first level
speaks of ``damage'' while the second level speaks of ``survival.''
Thus we need either some inference from damage to survival, or the
first level claim should also speak of survival.  This leads to a key
insight: the evidence comes exclusively from planes that survived.
Hence the first level claim should be changed to ``the fuselage is the
part of the plane that can survive heaviest damage.''  From there it
is a short step to deduce that planes with heavy damage to the engines
did not survive and hence the celebrated recommendation by Abraham
Wald that the best place to apply armor is where there are no bullet
holes \cite{Ellenberg15}.

 \exmemo{As in this example, evidence may comprise multiple items and, unlike
 the example, we may have separate assessments for each of these (e.g.,
 test results, mutation analysis, and the quality of the oracle).
 These items might not be independent and so we will need to use
 methods such as BBNs to combine their probabilities.\exfootnote{What
 about the argument, used later, that everything must be true in an
 assurance case and hence we don't need BBNs.}  Confirmation measures
 may be calculated for the combination or for each item
 separately.\footnote{Note that confirmation measures are not
 probabilities and cannot be combined in BBNs, however the component
 posteriors or likelihoods $P(E_1 \vbar C), P(E_2 \vbar C), \ldots,
 P(E_n \vbar C)$ can be combined.}

It can be difficult to assess suitable values for $P(E_i \vbar C)$ so
we expect these calculations to be performed qualitatively (e.g., low,
medium, high), with representative numbers (e.g., 0.1, 0.5, 0.9) used
for ``what if'' calculations with BBNs to gain experience of the
interactions involved.
}

\exmemo{
Unfortunately, probabilistic and deductive reasoning do not combine
well.  We might hope that if evidence $E$ supports a claim $C$ and $C$
deductively entails $C\,',$ then surely $E$ should also support $C\,'.$
This expectation is dashed by the following counterexample.  Let $C$
be the hypothesis that a certain playing card (drawn at random from a
shuffled deck) is the Ace of Hearts, let $C\,'$ be the claim that the
card is red, and let $E$ be the evidence that the card is an Ace.
Certainly $E$ (Ace) supports $C$ (Ace of Hearts), which entails $C\,'$
(red), but $E$ (Ace) cannot be considered to support $C\.'$
(red).\footnote{This counterexample comes from a talk by Brandon
Fitelson, then of CU Berkeley; his website \url{http://fitelson.org/}
contains much material on these topics.}  It is not clear how to find
a general resolution for the difficulty illustrated by this example
but one observation is that the evidence ``Ace'' provides incomplete
support for the claim ``Ace of Hearts''---this is really a conjunction
``Ace $\mand$ Heart'' and we have evidence for Ace but not for Heart.
In an assurance case, we should surely demand evidence that covers all
aspects of our claim and should be actively seeking defeaters that
satisfy our evidence but violate our claim.  Hence, we will discuss
methods for propagating probabilistic assessments of evidential
strength through the reasoning steps of an argument, even though the
theoretical basis for doing so may not be completely firm.  Our main
motivation for doing this is to seek insight on the consequences of
evidential weakening performed in the service of graduated assurance.

$P(\mbox{ace} \vbar \mbox{ace of hearts}) = 1$

$P(\mbox{ace} \vbar \neg \mbox{ace of hearts}) = 3/52$

so K-O measure is .95/1.05 = 0.90

}

\subsection{Overall Assessment of Soundness}
\label{overall-soundness}

As noted in Section \ref{interior}, the argument of an assurance case
is \emph{fully valid} when, in addition to being logically valid, all
its claims are supported by evidence, all its steps are deductive, and
there are no unresolved defeaters.  A fully valid argument is
\emph{sound} when human evaluators attest that the justification for
each interior step is satisfactory, and that the weight of each
evidential step is sufficient to justify its ``evidentially useful''
claim, as explained in Section \ref{leaves}.  \yy \clasce\ can help
check these requirements and highlight violations.  In particular, its
\emph{validity plugin} can check when a case is valid and can indicate
those parts of a case that are incomplete.  This is straightforward
with the positive arguments that have been discussed so far, but
becomes somewhat more difficult in the presence of defeaters, so we
refer to that document for a detailed description
\cite{Bloomfield-etal:defeaters23}.

Full validity and soundness could be indicated by allowing individual
argument nodes to be marked when they are considered deductive and
again when their justification is considered sound, and a modified
validity checker could propagate these assessments and display the
results.  However, \clasce\ does not do this; instead, it assumes that
argument nodes are fully valid and sound and users indicate otherwise
by adding defeaters.  We prefer this approach because defeaters can
have claims and subcases and thereby record detailed information on
the reasons for perceived deficiencies in the case.

During construction, an assurance argument will have many
imperfections and incompletenesses and the validity plugin helps
identify parts of the case that need attention, within the process
indicated by Figure \ref{process}.  But at completion, we expect the
case to be fully valid and sound, and the validity plugin should
concur with this judgement.  However, the requirements for full
validity are strong and, in practice, it may not be possible (or
credible) to make some steps deductive.  Therefore, exceptions can be
tolerated in the assessment of full validity, but the goal should be
``as deductive as possible and inductive only as strictly necessary.''
The same may be true in the assessment of soundness.  Both of these
exceptions can be indicated and nullified by assumptions (which will require their
own indefeasible justifications) or accepted as residual risks, which
are indicated by unresolved defeaters.  The residual risks due to
unresolved defeaters must be adjudged small, as will be described in
Section \ref{residuals}.  Apart from this circumstance, no unresolved
defeaters should be present when a case is submitted for assessment:
the source of concern should have been eliminated or mitigated and
thereby reduced to a residual risk, or localized to an assumption.

\exmemo{Does Figure \ref{process} need revisiting?}

\begin{figure}[t]
\begin{center}
\vspace*{-12ex}
\includegraphics[width=\textwidth]{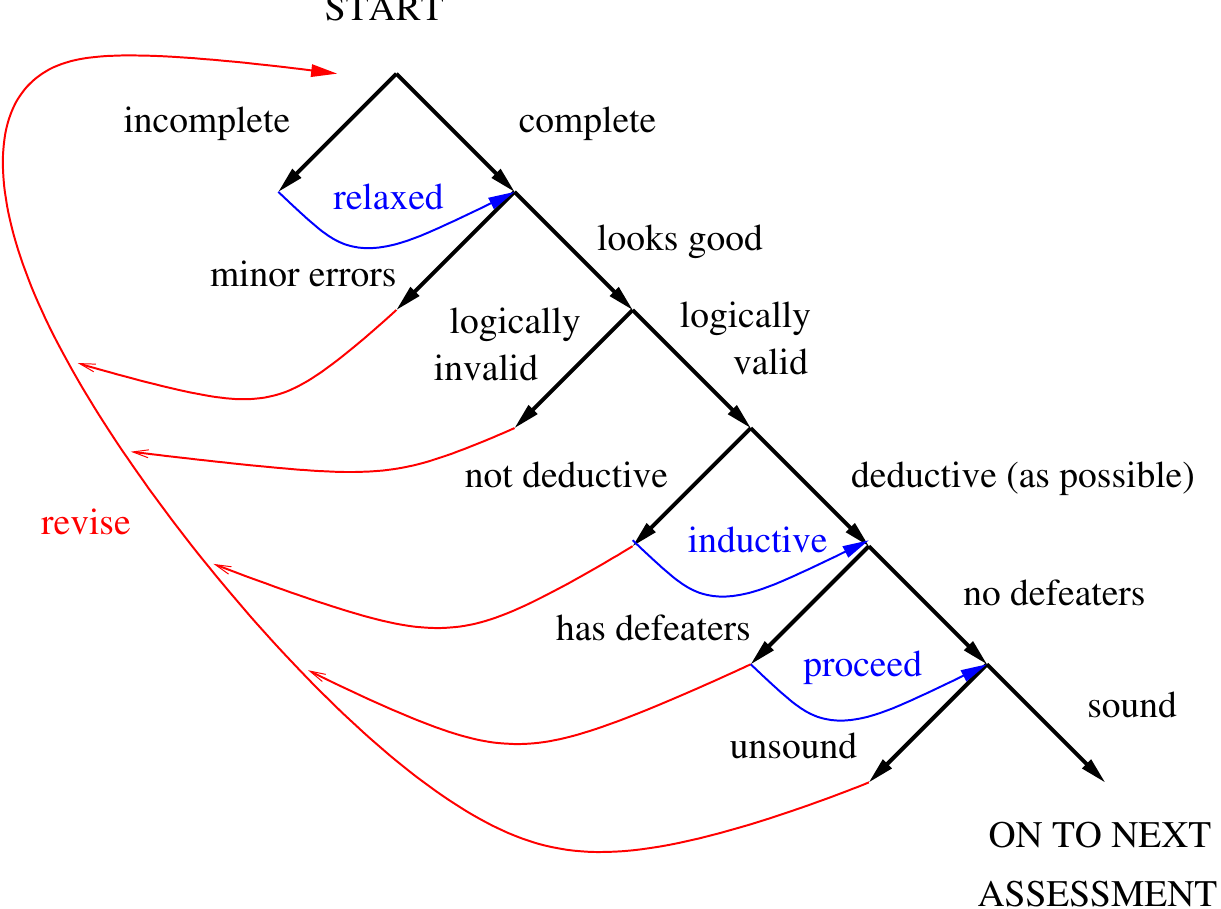}
\end{center}
\vspace*{-2ex}
\caption{\label{process}Logical Development and Assessment Process}
\vspace*{-2ex}
\end{figure}

These steps in the process of logical assessment for an assurance case
argument are portrayed in Figure \ref{process}.
Black arrows heading rightward indicate satisfactory progress, while
those heading leftward indicate problems that require correction,
as indicated by red arrows, or a decision to proceed anyway, as
indicated by blue arrows.

This, if the case is incomplete, the assessment proceeds in a relaxed
mode; if minor errors (e.g., misspelled claims) are present, these
should be corrected but we do not flag them as invalid.  Arguments
flagged as invalid should be corrected.  Defeaters can then be used to
flag steps that are not fully valid, such as those considered
nondeductive or unsound, or that raise some other objection.  The
order in which validity, deductiveness, and other defeaters are
assessed is not important.

A completed case should have a fully valid argument and is considered
sound when all its steps are assessed to indefeasibly justify their
claims, based on the subclaims or evidence provided.  Confirmation
measures can assist this judgment for evidential steps but we do not
recommend developers or evaluators of assurance cases make their
numerical assessments of specific measures a rigid criterion for
judging the evidential justification for a claim.  Rather, we expect
them to apply the ideas presented in Section \ref{leaves}
``qualitatively'' or informally and to consider prior beliefs and
counterclaims when assessing the extent to which evidence justifies a
claim.  However, these qualitative judgments can be developed and
honed by ``what if'' experiments using numerical representations for
the probabilities concerned.  When evidence is composed of multiple
items that are assessed separately but are not conditionally
independent, then it may be necessary to use advanced methods and
tools such as Bayesian Belief Nets (BBNs)
\cite{Littlewood&Wright:tse07,Fenton&Neil12,hugin} to calculate the
probabilities that will be used in the confirmation measure.  Again,
we recommend numerical experiments using BBN tools to develop
understanding of the interactions involved and development of
appropriately simplified rules for qualitative or informal judgments.

Summarizing the discussion from the previous section, we suggest that
selection and assessment of evidence can proceed as follows.  First,
it is sensible to seek evidence that would be surprising (i.e., not
expected) if the claim is untrue and, if multiple forms of evidence
are available, to seek those that are diverse: that is, not associated
with each other unless the claim is true.  The claim should be at the
``something useful'' level and there should be a suitable theory
available that connects such claims to the measurements and
observations delivered as evidence.  We should follow Nicod's
criterion and select evidence that supports the claim directly rather
than by delicate inference and, if the claim is a conjunction, we
should ensure that each conjunct is supported by some part of the
evidence.  Next, a confirmation measure can be evaluated and assessed
as an indication of the weight of that evidence.  Keynes' measure is
attractive as its original and likelihood forms are equal and this can
be used to probe consistency of the judgments required.
Numerical experiments can be performed to establish a suitable
threshold for the measure.  Finally, counterclaims should be
considered and the discriminating power of the evidence assessed with
the aid of measures such as Good's.  Again, numerical experiments can
be used to establish thresholds.

We continue our examination of assurance cases from a positive
perspective by proceeding from their logical to their probabilistic
evaluation before turning to negative perspectives and the important
role of defeaters.

\section{Positive Perspectives: Probabilistic Valuation}
\label{propagation}

Soundness is one aspect in the positive assessment of confidence for
an assurance case, but it lacks graduation.  Suppose, for example, we
have a sound case, then reduce its threshold for weight of evidence
and reduce the quantity or quality of evidence accordingly; the case
remains sound, but we are surely less confident in the verity of its
top claim.  A different ``weakening'' is seen in DO-178C
\cite{DO178C}, where Design Assurance Levels (DALs) A to C require
both High and Low Level Requirements (HLR and LLR), whereas (the
lower) Level D requires only HLR\@.  Intuitively, the idea is that we
are less confident of the large ``leap'' in reasoning from
implementation directly to HLR than of the combination of steps from
implementation to LLR and then to HLR.

The motivation for these ``weakened'' cases is that they should be
cheaper to produce, yet might still be adequate for less critical
systems, or for less critical claims.  Dually, we would like some
basis for believing that the additional cost of the original
``strong'' cases does deliver greater confidence in their claims.
What we seek, therefore, is a way to augment soundness with a
graduated measure that indicates the strength of our confidence in the
case.

The strength of confidence in an assurance case is naturally expressed
as a probability.  We could assess this as a subjective evaluation of
the entire case, but a more principled method is to calculate it as
the composition of assessments for all the basic elements of the case,
such as evidence, and individual argument steps.  This will involve
some combination of logic and probability, which is a notoriously
difficult topic because the semantics of the two fields have different
foundations \cite{Nilsson86,Gaines78,Adams98}.

Nonetheless, there are numerous proposals for calculating
probabilistic confidence in assurance cases by methods of this kind
(e.g., \cite{Denney-etal11:ESEM,Ayoub-etal:SSS13}), but a study by
Graydon and Holloway \cite{Graydon&Holloway:quant17} cast doubt on
many of them.  Graydon and Holloway examined 12 proposals that use
probabilistic methods to quantify confidence in assurance case
arguments: five based on Bayesian Belief Networks
\cite{Fenton&Neil12}, five based on Dempster-Shafer \cite{Shafer76} or
similar forms of evidential reasoning such as J{\o}sang's opinion
triangle and subjective logic \cite{Josang16}, and two using other methods.
By perturbing the original authors' own examples, they showed that all
the proposed methods can deliver implausible results.

We suspect that the reason for this disappointing behavior is that the
methods concerned are attempting a double duty: they aim to evaluate
confidence in the case, but also (implicitly) its soundness.
Probabilistic methods are poorly suited to the latter task, which is
more naturally cast in terms of logic.  In \cla\ we separate these
evaluations and assess soundness as a logical property, as described
in the previous section, and only for cases assessed to be sound do we
proceed to assess probabilistic confidence.  Nonetheless, we intend to
explore Graydon and Holloway's examples when our tools are fully
developed.  Our methods for probabilistic valuation are compositional
over the five basic building blocks of Assurance 2.0 cases, as
described in the following subsections.

\subsection{Evidence Incorporation Blocks}
\label{evincsec}

\begin{wrapfigure}{r}{.5\textwidth}
\vspace*{-3.5ex}
\begin{center}
\includegraphics[width=0.5\textwidth]{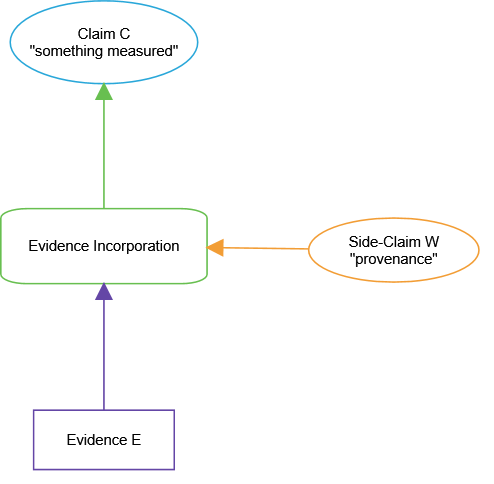}
\end{center}
\vspace*{-2ex}
\caption{\label{evinc}Evidence Incorporation Block}
\end{wrapfigure}

A generic evidence incorporation block is shown in Figure \ref{evinc}.
The basic idea is that observations (e.g., measurements, tests, or
analyses etc.) on the system under consideration yield evidence $E$
that is asserted to support claim $C\,$; the assertion of support is
justified by a narrative attached to the evidence incorporation
argument block, which may be further supported by a side-claim or
warrant W.  As indicated in the figure, the claim $C$ will generally
be about ``something measured'' and the side-claim $W$ will generally
concern provenance of the evidence and will be supported by a subcase
establishing this.  For example, if $E$ is evidence from testing, then
$C$ will be a statement about the test outcomes, such as ``no failures
in $n$ tests'' or ``MC/DC coverage achieved''; it will not be an
interpretation or ``something useful'' statement about the tests, such
as ``achieved reliability 0.99'' or ``no unreachable code present''
since these are ``something useful'' inferences derived from the test
measurements via a suitable theory of testing that will be applied in
a higher-level substitution block.  The provenance side-claim $W$ and
its supporting subcase must establish that the tested artifact is the
real thing, that the test oracle and any measurement harness are
trustworthy, and that the test procedure is sound and was performed
correctly, and so on.

As we explained in Section \ref{leaves}, the subjective posterior
probability $P(C \vbar E)$ is a natural expression of confidence in
the claim $C,$ given the evidence $E.$ However, when assessing
soundness we use a confirmation measure rather than the posterior
probability because we wish to evaluate the discriminating power, or
``weight,'' of the evidence, and confirmation measures do this.  But
once we have assessed soundness, it is reasonable to use the posterior
(or a qualitative approximation to this) as our measure of
probabilistic confidence in the claim $C$ and it is this that will be
propagated through the probabilistic valuation for the rest of the
case.  Note that in Section \ref{confirmation} we condone informal
(i.e., non-numerical) interpretation of confirmation measures, but for
confidence we do require an actual numerical estimate for the
subjective probability $P(C \vbar E)$\@.  (Note further that if the
evidence incorporation step uses multiple items of evidence then their
individual contributions $P(C \vbar E_1), P(C \vbar E_2), \ldots, P(C
\vbar E_n)$ will be combined into the overall $P(C \vbar E)$ using
methods such as BBNs.)

Now $P(C \vbar E)$ is an epistemic judgment and it can be performed
in several ways.  One way would consider the totality of the
information about $E,$ including its provenance and other items
represented in the side-claim $W.$ In this case, probabilistic
confidence in $C,$ which we write as $\conf(C),$ will be simply the
``holistic'' $P(C \vbar E).$ Another way might assess $P(C \vbar E)$
using only the direct contribution of $E$ to $C,$ with the side-claim
$W$ assessed separately.  In this case, probabilistic confidence in
$C$ will be some combination of $P(C \vbar E)$ and probabilistic
confidence in the side-claim, $\conf(W),$ which will be accumulated
over the subcase supporting $W.$ There are several plausible forms for
the combination including arithmetic product (corresponding to logical
conjunction) \[\conf^P(C) = P(C \vbar E) \times \conf(W)\] and ``sum
of (probabilistic) doubts'' \[\conf^D(C) = P(C \vbar E) + \conf(W) -
1.\] We describe these in more detail in the following section.

\subsection{Substitution and Concretion Blocks}
\label{sub-conc}

A generic substitution or concretion block is shown in Figure
\ref{substconc}.  As described earlier, in Section
\ref{substitution}, a substitution block has a subclaim $S$
expressing some property $A$ of a model $P$ that is used to justify
property $B$ of model $Q$ as the claim $C$.  Special cases arise when
the properties or models are the same.  If the properties $A$ and $B$
are the same (so that, for example, we are justifying correctness of
the HLR on the basis of correctness of the LLR), then the method of
justification is generally to show that the models $P$ and $Q$ are
``equivalent'' which may be achieved informally by traceability
analysis, or formally by verification of a homomorphism.  If the
models $P$ and $Q$ are the same (so that, for example, we are
justifying absence of unreachable code on the basis of MC/DC coverage
by requirements-based tests) then the method of justification is
generally to appeal to some theory that addresses the topic.

Concretion blocks are somewhat similar: they typically justify a claim
about an abstract property and model (e.g., ``the system shall be
correct'') on the basis of a subclaim about a more concrete property
and model (e.g., ``the system shall satisfy its HLR'').

The significant feature of substitution and concretion blocks from the
perspective of probabilistic confidence propagation is that the claim
$C$ is derived from just a single subclaim $S,$ subject to a
side-claim or warrant $W.$
For soundness, we require that the parent claim is deductively
entailed by the subclaim, subject to the side-claim.  That is, $ W
\supset (S \supset C),$ which is equivalent to $ W \wedge S \supset
C.$ A narrative in the argument block must justify this relationship
indefeasibly.

\begin{wrapfigure}{R}{.5\textwidth}
\vspace*{-2ex}
\begin{center}
\includegraphics[width=0.5\textwidth]{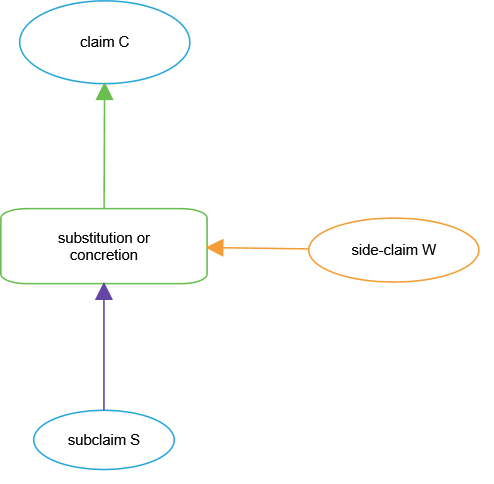}
\end{center}
\vspace*{-3ex}
\caption{\label{substconc}Substitution/Concretion Block}
\vspace*{-4ex}
\end{wrapfigure}

When we consider probabilistic confidence, we apply a probabilistic
interpretation to the implication above, so that \[\conf(C) \approx
\conf(W \wedge S).\] The probabilistic confidence $\conf(W \wedge S)$
is given by $\conf(W) \times \conf(S \vbar W)$ but we expect the lower
steps of the argument (i.e., the subcases supporting $W$ and $S$) to
supply $\conf(W)$ and $\conf(S)$\@.  Since $\conf(S)$ is not the same
as $\conf(S \vbar W)$ (unless $S$ and $W$ are independent), this is
not quite what is required.  However, the structure of a sound
assurance case is such that all the claims and subclaims appearing in
its argument must be true---so when we evaluated the subclaims and
evidence contributing to $S,$ we implicitly did so in a context where
$W$ is true.  Hence, our assessment of probabilistic confidence in the
subclaim $S$ is really confidence \emph{given} the rest of the case,
and so the confidence we labeled $\conf(S)$ is ``really'' $\conf(S
\vbar W)$ and probabilistic confidence in $\conf(W \wedge S)$ can
indeed be taken as the product of probabilistic confidence in its two
subclaims.  Thus \[\conf^P(C) = \conf(W) \times \conf(S)\] where we
use the superscript $P$ in $\conf^P(C)$ to indicate this is the
``product'' calculation.

\exmemo{We might decompose over the truth of $W$: e.g., we consider
separately the case $S \wedge W$ and $S \wedge \neg W.$  Yes, but
would we then have evidence for both cases?  It seems this is more
like an internal case split, rather than something that rises to the
level of the case.}

Some may feel that this calculation is too aggressive and would prefer
a more conservative approach.  One such is the ``sum of doubts''
approach: our doubt in the parent claim is no worse than the sum of
doubts for its subclaims and side-claims \cite{Adams98}.

The ``doubts'' referred to here are \emph{probabilistic doubts} as
opposed to the other use of the term to mean general disquiet or
concern.  Probabilistic doubt in a claim $C$ is probabilistic
confidence in its negation $\conf(\neg C),$ so
\begin{eqnarray*}
\conf(\neg C) & \approx & \conf(\neg(W \wedge S))\\
 & = & \conf(\neg W \vee \neg S)\\
 & = & \conf(\neg W) + \conf(\neg S) - \conf(\neg W \wedge \neg S)\\
 & \leq & \conf(\neg W) + \conf(\neg S).
\end{eqnarray*}
Then, since $\conf(\neg C) = 1 - \conf(C),$ and similarly for $\neg W$
and $\neg S,$ we have
\[\conf^D(C) \geq \conf(W) + \conf(S) - 1\]
where the superscript $D$ indicates this is the ``sum of
(probabilistic) doubts'' calculation.

These derivations of $\conf^P(C)$ and $\conf^D(C)$ also justify the
similar formulas presented in the previous subsection for evidence
incorporation steps.

Choosing a method for propagation of probabilistic confidence is a
matter for judgment by the developers and evaluators of an assurance
case.  We have proposed two candidates for $\conf(C)$: one,
$\conf^P(C)$ is aggressive but makes strong assumptions; the other,
$\conf^D(C)$ is conservative but requires only weak assumptions.
There is no reason to think that either candidate is ``correct'' and
developers may use either of these or an alternative method (with a
satisfactory explanation) of their own devising.  We do, however,
consider it reasonable that any good estimate will lie between those
given by $\conf^P(C)$ and $\conf^D(C).$
\clasce\ has a ``confidence plugin'' that mechanizes the sum of doubts
calculation.

The indefeasibility criterion of Assurance 2.0 requires that the
conjunction of subclaim and side-claim should deductively entail the
parent claim; hence we identify confidence in the parent claim with
confidence in this conjunction.  However, it is possible that,
although we are persuaded of the logical entailment, our probabilistic
confidence in the parent claim differs from that suggested by the
calculation above.  For example, we noted earlier that DO-178C
requires both High and Low Level Requirements at DALs A--C, but only
HLR at DAL D.  Intuitively, the idea is that the more costly
combination of substitution steps from implementation to LLR and then to HLR
engenders more confidence than the single large step from
implementation directly to HLR, even though all the steps are
assessed as deductive.  This can be accommodated by defining a factor $f$ such
that confidence in the parent claim $C$ is given by
\begin{equation}\label{factor-f}
\conf(C) =
f \times \conf(W \wedge S)\footnote{Alternatively, $f$ could be a function
$f(\conf(W \wedge S))$.}
\end{equation}
\footnotetext{Alternatively, $f$ could be a function: $f(\conf(S \wedge W))$.}
(or a qualitative approximation thereto), where
$\conf(S \wedge W)$ represents the chosen method for combining $\conf(S)$
and $ \conf(W)$.

Exceptionally, we allow the \emph{inductive} justification of an
argument step, where the conjunction of the subclaims and side-claims
``strongly suggest'' but do not imply the parent claim.  This means
there must be some missing element\footnote{It is possible that $W$ or
$S$ are \emph{wrong} rather than weak, and therefore cannot be
corrected by conjoining an $M.$ See the discussion of residual
interior doubts in Section \ref{resinterior}.} $M$ that would make the
relationship deductive: \[ W \wedge M \wedge S \supset C.\] Presumably
$M$ is unknown (otherwise we would have included it), but the fact
that we have labeled the argument step inductive means that we
recognize its (possible) existence.  The reason we deprecate inductive
steps is that the absent $M$ represents a \emph{defeater}: that is, a
condition that can invalidate the argument, and we prefer these to be
identified more specifically.

Rather than label the step inductive, we could instead represent $M$
explicitly as an assumption or as an unsupported claim labeled
``something missing here''  (recall Figure \ref{decomp-doubts}).
These may be conjoined either to $W$
(i.e., the side-claim is too weak) or to $S$ (i.e., the subclaim is
too weak), or split between them.  Alternatively, $M$ could be
represented by an explicit defeater node.  We prefer these
alternatives to inductive steps because they are more explicit about
the existence and location of doubt.

When evaluating probabilistic confidence in such a case, a strict
approach would assess zero confidence for inductively justified
claims, unsupported ``something missing'' claims, and the targets of
defeater nodes.  Such assessments will propagate upwards and deliver
zero confidence in the top claim.  However, we have already indicated
doubt in the case and are aware of its defeasible character, so we
would like the confidence assessment to say something useful beyond
this.  \clasce\ accommodates this by allowing manual adjustment to
calculated values for probabilistic confidence: confidence in a claim
targeted by a defeater can be left unchanged, or set to zero, or
reduced to some intermediate value.  Nodes are color coded according
to user-selected thresholds on their probabilistic confidence, as
illustrated in Figure \ref{traffic}, and adjustment to these settings
can provide developers with visualizations that help focus their
attention on weak areas of the case and to comprehend the scope of
their impact.

\subsection{Decomposition Blocks}

Decomposition blocks are used when a claim can be decomposed into
subclaims distributed over some set or structure, such as components,
properties, configurations, hazards, or time, and so on.  A side-claim
ensures that the decomposition is valid: that is, all elements of the
decomposition are considered, and the subcases are disjoint, etc.  A
generic example was shown earlier, in Figure \ref{decomp-block}.

As with substitution and concretion blocks, there are several plausible
ways to estimate probabilistic confidence in the parent claim $C$ from
that in its subclaims $S_1,\ldots,S_n$ and side-claim $W.$  These
include the product calculation
\[\conf^P(C) = \conf^P(W) \times \prod_{i=1}^n \conf^P(S_i)\]
and the sum of probabilistic doubts
\[\conf^D(C) = \conf^D(W) + \sum_{i=1}^n \conf^D(S_i) - n,\]
which can each be derived by generalizing the corresponding description
in the previous subsection.

\subsection{Calculation Blocks}

Calculation blocks are much like decomposition blocks except the claim
and subclaims concern the values of some (usually numerical)
quantities and the quantity in the parent claim is calculated from
those in the subclaims using a formula justified (presumably by citing
some theory) in the body of the calculation block, subject to
constraints cited in the side-claim.  The analysis of probabilistic
confidence then follows that for decomposition blocks.

\subsection{Overall Assessment of Probabilistic Confidence}
\label{probs}

If the goal is to make a strongly argued case for some probabilistic
assessment of the system under consideration, such as its reliability,
then it seems best to make this quantity a part of the top claim and
to arrange the case to justify it explicitly, probably by reference to
suitable theories for reliability estimation (see \cite[Section
3.2]{Bloomfield&Rushby:Assurance2} for an example).  We call this an
\emph{internal} probabilistic assessment because it is constructed
inside (i.e., as part of) the argument.  A variant is an
\emph{indirect} probabilistic assessment where we use a substitution
block to justify a probabilistic claim by adherence to a standard.
For example, DO-178C \cite{DO178C} and governing regulations
\cite{10-9,CS-25} indicate DO-178C Level A is suitable for a
probability of failure on demand (\emph{pfd}) of $10^{-9}$
while Level D is
sufficient for \emph{pfd} of $10^{-3}$.  (The
notion of \emph{demand} is flexible; for example, it can be
interpreted as a single iteration of a cyclic control system, or an
hour of operation---the usual case in avionics---or a complete
mission.)

In contrast to these approaches, an \emph{external} probabilistic
assessment is constructed outside the (otherwise independent) argument
in the manner illustrated by the previous subsections, and will
generally be much more approximate as it depends on generic analyses.
Also, it should be noted that an external assessment will deliver
\emph{probabilistic confidence in a (logical) claim} (e.g., 99.9\%
confident in the absence for critical faults), not a
\emph{probabilistic claim} (e.g., reliability wrt.\ critical failures
better than 99.9\%).  Additional work is required to connect these
concepts, as described in Section \ref{cbi}.

Unlike other authors who develop theories for external assessment of
probabilistic confidence in assurance cases, we do not advocate any
particular approach (although the chosen approach should be used
consistently) nor consider any of them ``correct.''  We have proposed
two plausible approaches that we dub the ``product'' and ``sum of
doubts'' calculations.  The product calculation assumes that subclaims
are independent, which may not be so.  For example, we might establish
partitioning among tasks by decomposing this into time partitioning
and space partitioning.  These are logically disjoint, but both might
be enforced by the operating system kernel, so they are hardly
independent.  Conversely, the sum of probabilistic doubts calculation
is very sensitive to subclaims with low confidence, but analysts may
consider those subclaims to apply only to unimportant (i.e., low risk)
circumstances.  Thus, analysts may adjust either of these estimates by
systematically applying some factor or function $f$ to the calculation
of $\conf(C)$, or they may manually override the calculation for
individual blocks, both to increase their values (e.g., due to
justification by a strong theory) or reduce them (e.g., due to
residual doubts).  Alternatively, they may substitute a calculation of
their own devising, perhaps employing BBNs or Dempster-Shafer's theory
of evidence \cite{Shafer76}, but we do expect that reasonable
estimates will lie between $\conf^D(C)$ and $\conf^P(C).$

\clasce\ has a plugin that can calculate probabilistic confidence
based on user-specified confidence in evidence and using either the
product or sum of doubts methods and can color nodes (e.g., red,
amber, green) according to user-selected thresholds to indicate the
confidence calculated for them, as illustrated in Figure
\ref{traffic}.  In addition to valuations of probabilistic confidence
calculated by propagation from evidence, \clasce\ allows manual
adjustment to these values (as an informal version of the factors
$f$),

\begin{figure}[t]
\begin{center}
\begin{tabular}{|c|c|c||c|c|}
\hline
 n  & $\conf(S_i)$ & $\conf(W)$ & \rule[-1.5ex]{0ex}{3.9ex}$\conf^P(C)$ & $\conf^D(C)$\\
 \hline \hline
 1 & 0.99 & 0.99 & 0.98 & 0.98 \\
 1 & 0.95 & 0.95 & 0.90 & 0.90 \\
 1 & 0.90 & 0.90 & 0.81 & 0.80 \\
 1 & 0.95 & 0.80 & 0.76 & 0.75 \\
 \hline 
 2 & 0.99 & 0.99 & 0.97 & 0.97 \\
 2 & 0.95 & 0.95 & 0.86 & 0.85 \\
 2 & 0.90 & 0.90 & 0.73 & 0.70 \\
 2 & 0.95 & 0.80 & 0.72 & 0.70 \\
 \hline
 3 & 0.99 & 0.99 & 0.96 & 0.96 \\
 3 & 0.95 & 0.95 & 0.81 & 0.80 \\
 3 & 0.90 & 0.90 & 0.66 & 0.60 \\
 3 & 0.95 & 0.80 & 0.69 & 0.65 \\
 \hline
 5 & 0.99 & 0.99 & 0.94 & 0.94 \\
 5 & 0.95 & 0.95 & 0.74 & 0.70 \\
 5 & 0.90 & 0.90 & 0.53 & 0.40 \\
 5 & 0.95 & 0.80 & 0.62 & 0.55 \\
 \hline
\end{tabular}
\end{center} 
\caption{\label{conftab}Confidence Calculations for
Representative Decomposition Blocks}
\end{figure}

\begin{figure}[t]
\vspace*{-4ex}
\begin{center}
\includegraphics[width=1.0\textwidth]{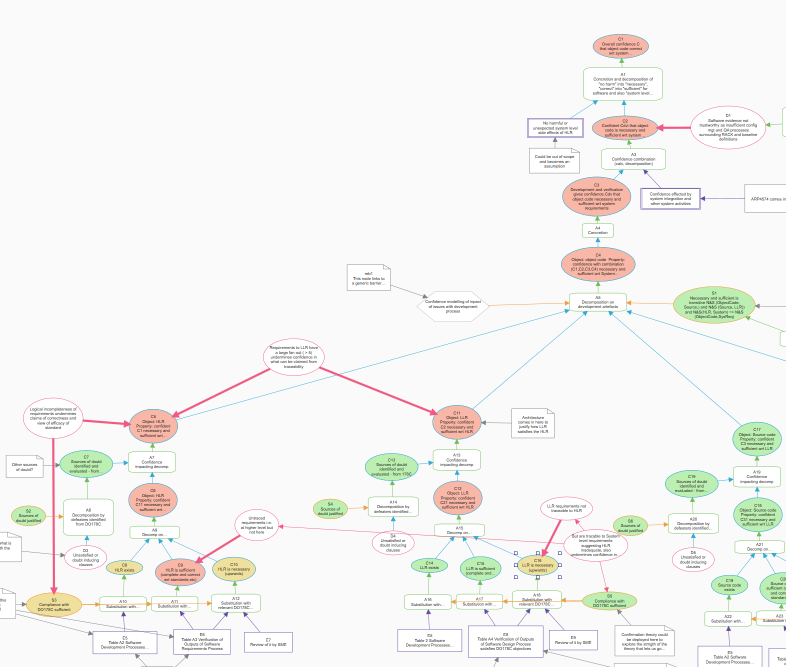}
\end{center}
\vspace*{-2ex}
\caption{\label{traffic}Portion of \clasce\ Canvas Showing Probabilistic Valuations}
\vspace*{-2ex}
\end{figure}

We have often advocated numerical ``what if'' experiments to develop
intuition on the behavior of probabilistic measures and in Figure
\ref{conftab} we present the results of one such experiment.  We
consider a decomposition block with claim $C$ having $n$ subclaims
$S_i$ and a side-claim $W.$  The first three columns of the table list
values for $n,$ $\conf(S_i)$ (assumed to be the same for all
subclaims), and $\conf(W),$\footnote{Of course, the results would be
unchanged if the same probabilities were distributed differently among
the subclaims and side-claim.}  while the two rightmost columns list
corresponding values for $\conf^P(C)$ and $\conf^D(C),$ respectively.
The first four rows consider the case $n=1,$ which covers substitution
and concretion blocks (and, to a certain extent, evidence
incorporation blocks as well). 

We see that for most combinations, $\conf^P(C)$ and $\conf^D(C)$ are
very close\footnote{Note that the product calculation with $n$ total
side- and subclaims yields $(1-d)^n$, where each of these claims is
assumed to have the same probabilistic doubt $d$.  By the binomial
expansion, this is $1-n\times d+ \mathrm{smaller\ terms}$.  Similarly,
the sum of doubts is $n \times d$, which yields confidence $1- n\times
d$.  Thus the product and sum of doubts calculations deliver very
similar values for confidence when $d$ is small.}; they diverge as
confidence in the subclaims is reduced (corresponding to the third row
in each block) and their number is increased.  We also see that one
low-confidence input (corresponding to the fourth row in each block)
has a substantial impact and that this is most marked for the sum of
doubts measure and for larger $n.$

The main conclusions seem to be that decomposition blocks with many
subclaims require strong confidence in those subclaims, and that all
subclaims (and side-claims) need similar levels of confidence.

Because these analyses are generic,
they are conservative to the point where their actual numerical
estimates may be of little direct value.  For example, if we take the
$n=3$ case in Figure \ref{conftab}, we see that when confidence in the
subclaims or evidence supporting the block, and in the side-claim, is
0.99, then confidence in the parent claim is 0.96 (i.e., 0.99 to the
fourth power, using the product calculation); if we iterate this
(i.e., we use four such blocks to support the three subclaims and the
side-claim for another such block), confidence in the next level will
be 0.85 at best (i.e., 0.96 to the fourth power), then 0.52 at the
third level, and a mere 0.07 at the fourth level.  Using sum of
doubts, the corresponding doubts are 0.04 (i.e., 0.01 times 4) at the
first level, 0.16 at the second, then 0.64 at the third, and 1 (since
doubts cannot exceed 1) at the fourth level, corresponding to a
confidence of zero.  These calculations seem to suggest that larger
arguments, with larger quantities of evidence, tend to reduce total
confidence.  This might be true, all other things being equal, but
they do not remain equal as we will now see.

First, we recognize that the basic product and sum of doubts
calculations do indeed depend only on confidence in the evidence
supplied and in any assumptions\footnote{We treat unsupported claims
as assumptions, and expect some probabilistic assessment of confidence
to be assigned to them.}, but not on the shape or size of the argument
tree above them.  That is, for the product calculation, probabilistic
confidence in the top claim is simply the product of probabilistic
confidence in all evidence and assumptions (as can be proved by
structural induction on the height of the argument tree).  Similarly, for the sum
of doubts calculation, the doubt in the top claim is simply the sum of
probabilistic doubts over all evidence and assumptions.

However, although the basic calculation of confidence depends only on
the evidence and assumptions and not on the argument, this does not
imply that confidence is invariant under transformation to the
argument---for different arguments may require different evidence, and
even the same evidence may deliver different confidence for different
claims.  Furthermore, the transformed argument must remain sound and
this may lead to different side-claims that require different evidence
or assumptions.  More significantly, a transformed argument may have
better justification.  Recall that in Section \ref{sub-conc} formula
(\ref{factor-f}), we introduced a factor or function $f$ that is used
to adjust probabilistic confidence according to the strength of the
justification provided, and this may change as the argument changes.

DO-178C illustrates these topics.  Design Assurance Levels (DALs) D
and E of DO-178C reason directly from code to HLR, whereas the higher
DALs A, B, and C insert intermediate steps involving LLR.\@ These
extra steps involve additional evidence and it might seem this will
reduce overall confidence.  It certainly would do so if confidence in
the additional steps were comparable to that in the originals but, as
described earlier, confidence in the smaller steps from code to LLR
and then to HLR can be much greater (due to stronger theories and
methods for verification) than the single large leap from code to
HLR\@.  This will be reflected in stronger confidence in the evidence
and in the justifications (and hence in the factors $f$), resulting in
greater confidence overall.

As we noted earlier, the absolute numbers delivered by our external
methods of probabilistic assessment must be used with caution.
Because it is based on human assessment of confidence in evidence and
assumptions, and because this may vary greatly with different
assessors or circumstances \cite{Kahneman21:noise}, the very
foundation of the calculation may be considered unstable.
Furthermore, because the methods of propagation are generic and
conservative, they likely underestimate the overall confidence that a
``more exact'' probabilistic calculation would assess, if such were
available.

We are comfortable with these limitations: we consider soundness to be
the critical property, with probabilistic confidence as a useful but
inexact augmentation that assessors can use to keep track, in a
rational manner, of the location and extent of weak and strong parts
of an argument and that may be particularly valuable when exploring
residual doubts (see Section \ref{residuals}) and graduated assurance.
That is to say, although probabilistic confidence calculated for the
top claim may be small, the reasons for this apply uniformly and so it
can be used to compare different assessments of residual risk and
different methods for graduating an argument by trading cost
(e.g., for generating and gathering evidence) against confidence for less
critical components or properties.

We might wonder how it can be that the record shows modern aircraft,
say, to be safe and reliable when the calculations discussed here
deliver rather small estimates for probabilistic confidence in these
properties?  One response is that the conservative and generic
calculations considered here severely underestimate the true values
because they take no account of the particular kinds of system
concerned.  In general, these are highly redundant control systems.
As control systems, they sample sensors and inputs and drive actuators
and signals many times a second, and substantial state is stored in
the physical plant itself.  Thus, a ``small'' fault in a calculation
(i.e., one encountered very infrequently) will affect only one or a
few iterations of a controller (otherwise it would be a ``big'' fault
and we are confident these will be found and eliminated) and the
controller will correct the effects of these in the next few cycles,
just as it does for environmental disturbances.\footnote{Of course,
Murphy's Law ensures that some ``small'' defects can have major
consequences: for example, the in-flight upset to an Airbus A330
\cite{VHQPA}.}  In addition, the system will generally be heavily
redundant with a lot of voting and averaging: these are present to
tolerate major failures, but will often mask small defects, too.
Truly accurate assessment of probabilistic confidence would need to
consider these contextual and architectural factors.  

A second response is that the confidence assessments we apply to
individual argument steps \emph{a priori} severely underestimate those
that can be applied retrospectively based on \emph{a posteriori}
experience.

In the next section we present a method for deriving estimates for
long run safety (i.e., reliability with respect to critical failures)
on the basis of \emph{a priori} probabilistic confidence in the
absence of faults as delivered by an assurance case, plus limited
\emph{a posteriori} operational or test experience.  This method
employs Conservative Bayesian Inference (CBI) to derive posterior estimates
for safety from prior confidence in the absence of faults (or, in a
more detailed analysis, absence of ``large'' faults).  To generate
useful conclusions, these methods require the prior probabilistic
confidence to exceed 0.9.  We have seen that generic methods of
confidence propagation may struggle to achieve this and so we expect
the final determination that the assurance case is ``good enough'' to
justify adequate confidence will be a holistic human assessment, based
on a number of factors (such as are being developed throughout
Sections \ref{soundness}--\ref{residuals}), rather than a conservative
calculation of probabilistic confidence using the methods described in
this section.

\newpage
\section{From Confidence to Safety}
\label{cbi}

The top claim of an assurance case often asserts that the system has
some logical property (e.g., ``is safe''); implicitly, this implies
that the system has no faults that could jeopardize the property.  We
have examined positive perspectives on assessment of such claims and
explored methods for gaining and assessing confidence in them, but how
does that relate to the properties we really care about, which are
typically absence (or rarity) of critical failures?  In other words,
how do we get from confidence in absence of faults to rarity of
failures?

In this section we outline a theory called \emph{Conservative Bayesian
Inference} (CBI) that can accomplish this step.  We present it as a
separate theory, external to any assurance case argument, but we note
that it could be incorporated by extending an argument above its
existing top claim using a substitution block to take that (logical)
top claim about absence of faults to a new higher-level claim
about the probability of critical failures over the life of the
system.

Confidence in an assurance case argument can be expressed as
probabilistic confidence in its top claim $\conf(C^T).$ The top claim
$C^T$ typically concerns absence of faults or defects that could lead
to a critical failure.  Let us abbreviate ``suffers a critical
failure'' by simply ``fails'' and ``has a fault that could lead to a
critical failure'' as ``faulty'' (with ``nonfaulty'' as its negation);
these definitions are consistent with standard usage
\cite{Laprie:dependability}.  Then, by the formula for total
probability
\begin{eqnarray}
\lefteqn{P(\mbox{system fails [on a randomly selected
demand]})}\label{one}\\ & = & P(\mbox{system fails} \vbar \mbox{system
nonfaulty}) \times P(\mbox{system nonfaulty})\nonumber\\ & & \mbox{} +
P(\mbox{system fails} \vbar \mbox{system faulty}) \times
P(\mbox{system faulty}).\nonumber 
\end{eqnarray}

The first term in this sum is zero, because the system does not fail
if it is nonfaulty (as we have defined those terms).  We
let $\pff$ be the probability that the system ``fails if faulty,'' we
have $\conf(C^T)$ as the probability the system is nonfaulty (so that
$P(\mbox{system faulty}) = 1 -\conf(C^T)$), and therefore its
probability of failure on demand, $\pfd$ is given by
\begin{equation}\label{two}
 \pfd = \pff \times (1-\conf(C^T)).
\end{equation}

Different industries make different assessments about the parameters
to (\ref{two}).  Early nuclear protection, for example, seemed to
presume the system \emph{is} faulty, so in effect it set $\conf(C^T)$
to 0 and performed extensive random testing to substantiate
(typically) $\pff < 10^{-3}.$  If those regulators had accepted that
modest amounts of assurance could deliver $\conf(C^T) \geq 0.9,$ then by
(\ref{two}) the same probability of failure could be
achieved\footnote{We are cutting a lot of corners here: the full
treatment must distinguish \emph{aleatoric} from \emph{epistemic}
assessment and must justify that beliefs about the two parameters can
be separated;
\cite{Littlewood&Rushby:TSE12,Littlewood&Povyakalo13,Littlewood&Povyakalo13b,Littlewood-etal:RESS17}
give details.} with the much less costly testing required to validate
merely $\pff < 10^{-2}.$

Dually, FAA AC 25.1309 \cite{10-9} and the corresponding European
regulations \cite{CS-25}  for aircraft certification indicate $\pfd
\leq 10^{-9}$ for catastrophic failure conditions and seem to presume
the system \emph{will} fail if it is faulty, so in effect they set $\pff =
1.$  The whole burden for assurance then rests on the value assessed
for $\conf(C^T)$---so that we need $\conf(C^T) \geq 1 -10^{-9},$ which
is completely implausible.  In fact, there is no credible assignment
of values to the parameters of (\ref{two}) that delivers $\pfd \leq
10^{-9}$ per hour \cite{Butler&Finelli91}; an alternative model is
needed.

Rather than the figure of $10^{-9}$ per hour, which is intended only
as ``an aid to engineering judgment'' \cite{10-9}, let us look at the
fundamental requirement of these regulations: that a catastrophic
failure condition is ``not anticipated to occur during the entire
operational life of all airplanes of one type.''  Extending
(\ref{two}), the probability of surviving $n$ independent demands
without failure, denoted $\psurv(n),$ is given by
\begin{equation}\label{longrun}
\psurv(n) = \conf(C^T) + (1-\conf(C^T)) \times (1-\pff)^n.
\end{equation}

Demands can be interpreted as hours of operation, or flights, or some
other measure of exposure and, whichever is chosen, a suitably large
$n$ can represent ``the entire operational life of all airplanes of
one type.''  The notable feature of (\ref{longrun}) is that the first
term establishes a lower bound for $\psurv(n)$ that is independent of
$n.$  Thus, if assurance gives us the confidence to assess, say,
$\conf(C^T) \geq 0.9$ (or whatever threshold is used to interpret
``not anticipated to occur'') then it seems we have sufficient
confidence to certify the aircraft as safe.

However, we can imagine using this procedure to provide assurance for
multiple airplane types; if $\conf(C^T) = 0.9$ and we assure 10 types,
then we can expect that one of them will have faults.  In this case,
we need confidence that the system will not suffer a critical failure
despite the presence of faults, and this means we need to be sure that
the second term in (\ref{longrun}) will be well above zero even though
it decays exponentially.  This confidence could come from prior
failure-free operation (e.g., flight tests).  Calculating the overall
$\psurv(n)$ can then be posed as a problem in Bayesian inference: we
have assessed a value for $\conf(C^T),$ have observed some number $r$
of failure-free demands, and want to predict the probability of seeing
$n-r$ future failure-free demands.  To do this, we need a prior
distribution for $\pff,$ which may be difficult to obtain and
difficult to justify for certification.  However, Strigini and
Povyakalo \cite{Strigini&Povyakalo13} show there is a distribution
(specifically, one in which $\pff$ is concentrated in a probability
mass at some $q_n \in (0,1]$) that delivers \emph{provably worst-case}
predictions; hence, we can make predictions that are \emph{guaranteed} to be
conservative, given only $\conf(C^T),$ $r,$ and $n.$  Using this
approach, which is known as Conservative Bayesian Inference (CBI),
Strigini and Povyakalo show that if $\conf(C^T)$ is above $0.9,$ then
$\psurv(n)$ is well above this floor, provided $r > \frac{n}{10}.$

If we regard a complete flight as a demand, then ``the entire
operational life of all airplanes of one type'' might require $n$ to
be in the range $10^{8}$ to $10^{9}$ (e.g., as of 2019, the Airbus
A320 series had performed over 150 million flights
\cite{Boeing:accidents}).  Flight tests prior to certification might
provide only $r = 10^3,$ so it appears this is insufficient for
certification by the criterion above.  However, it can be argued that
when an airplane type is certified we do not require (and in fact
cannot feasibly obtain) sufficient evidence to predict failure-free
operation over the entire lifetime of the type; instead, we initially
require sufficient confidence only for, say, the first six months of
operation and the small number of aircraft that will be manufactured
and deployed in that period.  This will be a much smaller value of
$n,$ and our $\conf(C^T)$ (from assurance) and our $r$ (from flight
tests) will be sufficient for confidence in its failure-free
operation.  Then we will need confidence in the next six months of
operation, with a larger fleet, (i.e., a larger $n$) but now have the
experience of the prior six months failure-free operation (i.e., a
larger $r$) and in this way we can ``bootstrap'' our way forward
\cite{Strigini-etal21:bootstrapping}.

It remains to consider what happens if experience in operation does
reveal a fault (by manifesting a failure, hopefully not
catastrophic---indeed, there is an FAA requirement that no single
fault may cause a catastrophic failure condition).  Commercial
airplanes operate in a legal and ethical framework where all incidents
and accidents are promptly reported and dispassionately investigated.
The FAA issues Airworthiness Directives mandating workarounds or
corrections to detected faults; in extreme cases it may temporarily
ground the fleet (as it did for Boeing 787 battery problems in January
2013 and 737 MCAS faults in 2019\footnote{The fatal crashes caused by
design faults in the MCAS system of 737 Max aircraft may seem to
repudiate the safety and certification arguments made here.  However,
it is clear that Boeing was not following either the spirit or the
letter of established safety requirements and guidelines, and FAA
oversight was weak and possibly captured.  Thus, although this example
does not repudiate the methods described here, it does illustrate that
they cannot operate effectively outside a genuine safety culture.}).  Bishop
\cite{Bishop13} constructs a statistical model for this scenario and
shows that, under plausible assumptions, detection and repair of
faults significantly increases long run safety, even if the fleet
continues to operate after a fault has been discovered, and even if
repairs may be imperfect.

In its totality, the analysis above (which is based on research at
Adelard and City University in London
\cite{Bishop13,Bertolino&Strigini98,Bloomfield96:t-of-n,Strigini&Povyakalo13,Strigini-etal21:bootstrapping})
provides---for the first time, we believe---a plausible statistical
model that retrospectively explains the success of aircraft
certification, and other certification regimes based on similar
practices.  At the base of this analysis is an assessed confidence
(e.g., $\conf(C^T) > 0.9$) that the system is nonfaulty or
``fault-free'' with respect to critical requirements.

Traditionally, software assurance cases have delivered a top claim
that the software is nonfaulty with respect to critical requirements.
Confidence in this claim was generally assessed separately (or left
implicit) but we suggest that the case should now be expanded to
include this analysis.  The connection from confidence in
nonfaultiness to reliability in operation was also assessed separately
(or left implicit as prior to CBI and other conservative approaches
\cite{Bishop&Bloomfield02:ISSRE} there was no good theory to account
for it) and we suggest that this, too, should now be made explicit and
included in an expanded assurance case whose top claim would now
become reliability with respect to critical failures.

\memo{Need to explain the prior chapter doesn't do it, as admitted in
its final paragraph.}

There are other top claims, architectures, and methods of analysis
that function similarly to that described above.  A modification to
the analysis above replaces strict fault-freeness by \emph{quasi}
fault-freeness, meaning the system is either nonfaulty or is faulty
but with only a minuscule probability of failure
\cite{Littlewood-etal:RESS17}.  This is a more robust model and yields
attractive results \cite{Zhao-etal19:AVtests}, but the details are
more complicated.  Alternative properties to nonfaulty and
failure-free include mission risk, and the claim that a new system is
no worse than the old one.  And an alternative to external calculation
of probabilistic confidence in the case is internal justification of
probabilistic quantities such as these within the case itself (i.e.,
the claims in the case make probabilistic statements; recall section
\ref{probs}).  And as an alternative to single-threaded architectures,
Littlewood and Rushby \cite{Littlewood&Rushby:TSE12} provide a
rigorous analysis of ``monitored architectures'' in which a highly
trustworthy monitor checks the behavior of a less trusted primary
system as advocated, for example, by the F3269-17 standard for
unmanned aircraft \cite{F3269-17}.  Bishop and Bloomfield
\cite{Bloomfield96:t-of-n,Bishop&Bloomfield02:ISSRE} develop an
alternative worst-case analysis that predicts long term reliability
from an estimate of the number of faults $N$ at time of release (as
opposed to confidence in their absence) and the operating time $T$.
The predicted number of faults can be based on models of the software
development process, such as the empirically validated ``barrier''
model \cite{Bloomfield&Guerra-dsn02}.

This concludes our discussion on positive views of assurance cases and
we now turn from these to negative perspectives and the important role
of doubts and defeaters.

\newpage
\section{Negative Perspectives: Doubts and Defeaters}
\label{defeaters}

In \cla, the criterion for a completed assurance case is that it should
be \emph{indefeasible}, meaning that all identified doubts have been
addressed, and we are confident no credible doubts remain that
could change the decision supported by the assurance case 
\cite{Rushby:Shonan16,Bloomfield&Rushby:Assurance2}.  When we say that
all doubts have been addressed, we do not require that they are
eliminated; in appropriate circumstances, some doubts may be accepted as
residual risks, as will be discussed in Section \ref{residuals}.  What
makes the case indefeasible is that we know about these doubts, have
examined them, and made a conscious decision about them so that no
credible new information would make us change our decision.
Indefeasibility is lost when there may be doubts that we do not know
about, or doubts we do know about but have not consciously addressed.

Doubts are suspicions that some part of an assurance case may be inadequate or
wrong.  On investigation, the location and nature of the doubt should
become sharpened so that it can be expressed as a \emph{defeater}:
that is, a node in the assurance case argument that challenges or
refutes the specific claims, arguments, or evidence represented by
other nodes.  
In the graphical notation of \clasce, a defeater is represented by a
node having the same oval shape as a claim node, but colored red, and
the nature of its challenge is expressed by a claim within the node
(recall the discussion around Figure \ref{decomp-doubts}).
Initially, the defeater's claim may be rather vague (e.g., ``I think
there is something wrong here'') and later it may be refined into a
more specific ``counterclaim'' (e.g., ``the justification for this
argument step is inadequate'') whose investigation is recorded in a subcase
attached to the defeater.  In \clasce, we use the term ``doubt'' to refer to a
defeater that has not yet been investigated (i.e., has no subcase).

There are several ways to resolve doubts and their associated
defeaters.  One way is to argue that the defeater is unjustified or
incorrect, so there will be a subcase that \emph{defeats the
defeater}.  If a doubt is legitimate, however, we must adjust either
the argument, or the system (and then possibly the argument as well)
to eliminate or mitigate its defeater(s).  Alternatively, we may
consciously choose to accept the issue identified in the defeater as a
residual risk.  Acknowledged defeaters that require adjustment to the
system (as opposed to just its assurance case) are more properly
considered (system) \emph{hazards} than (argument) defeaters, but the
issues discussed here will apply to them as well.

The back and forth investigation of an assurance case argument against
doubts and defeaters is an application of the \emph{Socratic} or
\emph{dialectical} methods for exposing error and refining
beliefs.\footnote{In philosophy, the Socratic method is considered an
instance of dialectic \cite{dialectic-wiki}; the precise distinctions
do not concern us here.}  These date back to ancient Greece but retain
their potency.  In particular, defeaters play a r{\^o}le in argument
that is similar to falsification in science and mathematics
\cite{Lakatos}; as mentioned above, they can also be seen as the
analog, for arguments, of hazards to a system.  Thus, identification
of potential defeaters should not be seen as criticism but as a
contribution to the development and clear formulation of an assurance
case and part of a process to establish its indefeasibility.  In
addition, developers should consciously generate doubts, and
vigorously investigate their associated defeaters as a guard against
confirmation bias, and evaluators may raise potential defeaters as a
way to elicit additional explanation or to clarify their understanding
of some part of an assurance case.

Investigation and resolution of doubts and defeaters actually serves two
purposes: firstly, it is the primary means by which we avoid
confirmation bias and drive the case toward soundness and
indefeasibility; secondly, it helps later (re)developers and reviewers
comprehend the case as they find their own doubts have been
anticipated and answered.  For the first of these, it is enough just
to deal with the doubts and move on, but for the second it is
desirable to record the changes made so that reviewers can see how the
case responded to previously considered doubts.  This raises issues in
the way in which an assurance case should be represented and recorded.

For example, one step in a case may decompose a claim into subclaims
over some enumeration and we may doubt that this is done correctly, so
we establish a defeater that attacks the decomposition step (e.g., by
making the claim that the decomposition is incomplete).  We may then
develop a subcase that supports this claim.  But that subcase may
itself be challenged by another defeater, and so on.  One issue is how
to represent and evaluate assurance cases in the presence of defeaters
and their partially developed subcases.  We have the basic case that
is attempting to substantiate some positive claim, then defeaters
(negative claims) with subcases to substantiate them, and then
possibly further counter-defeaters and their subcases.  Do we show all
of these layers of defeat and counter-defeat?  And do we have some way
of showing which subcases are contributing to the basic case and which
to a defeater or counter-defeater or counter-counter-defeater?  We
will refer to this as the problem of representing and evaluating
\emph{defeasible arguments} and will briefly examine it in Sections
\ref{defeasible} through \ref{elimarg}.

When we satisfy ourselves that a defeater is itself defeated,
the investigation has served its purpose and we could eliminate
everything associated with it because the original case was valid
after all.  But reviewers may have the same doubts, and so for them it
is desirable to retain the defeaters and their subcases in a way that
permits interactive replay and exploration.  We will refer to this as
the problem of representing \emph{dialectical arguments} and will
examine it in Section \ref{dialectics}.

The problem of representing dialectical arguments becomes more complex
when the original case, or system, or both, need to be adjusted in
response to a defeater.  Whereas the subcase for an unsuccessful
defeater and the subcases for its counter-defeaters and so on can be
seen as comments or decorations on the basic, sound, case, a
successful defeater requires a change to the original case and so, for
review, we might need to retain the original case and its defeaters
and then branch to the adjusted case.

An interesting special case is that of missing assumptions.  Slavoj
{\v{Z}}i{\v{z}}ek \cite{Vzivzek06} identifies the ``unknown knowns,''
the ``silent presuppositions we are not aware of,'' as a significant
source of error in all human deliberation.  In an assurance case, a
defeater that identifies a missing assumption may be countered by
supplying the required assumption and, possibly, a subcase with
evidence, to justify it.  This amendment to the case is an addition
rather than a change and its representation and management can be
simpler than those for more general amendments in response to
defeaters.

We describe the treatment and analysis of defeaters in Assurance 2.0
and \clasce\ in detail in a separate document
\cite{Bloomfield-etal:defeaters24} and summarize it in Section
\ref{defeaters-in-cl}.  In the next section we outline alternative
choices and the motivation for the chosen approach.

\subsection[Defeaters in Reasoning, Argumentation, and Dialectics]{Defeaters in Reasoning, Argumentation, and Dialectics}
\label{theory}

Classical formal logic cannot tolerate contradictions among premises:
these render the argument invalid.  But in AI and in the study of
human and commonsense reasoning it is reasonable to draw conclusions
on the basis of incomplete and inconsistent information, and to revise
these as new information becomes available.  Consider, for example, an
emergency room physician, updating her diagnosis and adjusting her
treatment plan as new observations and test results become available.

In AI, the topic of drawing reasonable conclusions from incomplete,
inconsistent, and changing information is referred to as ``defeasible
reasoning''; much the same topic is considered in formal logic as
``nonmonotonic logic'' (``nonmonotonic'' because conclusions are not
stable and may need to be withdrawn as more knowledge is acquired).  A
standard example has premises ``birds can fly'' and ``Tweety is a
bird'' but then we learn that, contrary to the obvious conclusion,
Tweety cannot fly.  It turns out that Tweety is a penguin and the
first premise needs to be modified to ``most birds can fly.''  A
different, though related, application arises in the study of
``argumentation'': here, different parties have different views and
may advance premises that contradict each other.

We briefly examine defeasible argumentation seeking ideas that might
prove useful for interpreting assurance arguments in the presence of
defeaters.  We look first at methods for defeasible reasoning,
followed by argumentation theory, eliminative argumentation, and
dialectics.

\subsubsection{Defeasible Reasoning}
\label{defeasible}

The crucial notion of \emph{defeater} is from Pollock in 1987
\cite{Pollock87} (although he published on these topics as early as
1967).  Pollock distinguished between \emph{undercutting} and
\emph{rebutting} defeaters.  Weinstock and colleagues
\cite{Goodenough-etal:TR2014} later added \emph{undermining} defeaters
in an approach to assurance that they call \emph{eliminative
argumentation} that we examine in Section \ref{elimarg}.  Generally
speaking, an undercutting defeater challenges an argument step (i.e.,
its justification), a rebutting defeater challenges a claim, and an
undermining defeater challenges evidence.

Pollock proposed what is known as an epistemological approach to
evaluation of defeasible arguments.  This is a set of rules that
defines how a cognitively ideal agent would arrive at warranted
conclusions given a set of premises and their defeaters.  Pollock's
system has been criticized for lacking a ``normative standard,'' being
based on \emph{ad hoc} intuitions about how a reasonable agent would respond
to this or that cognitive situation, but Koons
\cite{sep-defeasible-resoning} observes that the same criticism can be
lodged against several other theories of defeasible reasoning.

Theories for nonmonotonic logic do have more justified foundations,
starting with McCarthy's \emph{circumscription} \cite{McCarthy80}
which, roughly speaking, prefers the most specific applicable
premises.  However, this approach gives intuitively incorrect answers
in some cases, epitomized by the ``Yale Shooting Problem''
\cite{Hanks&McDermott87} and so, as with defeasible reasoning,
there is a large literature of attempts to find more satisfactory
treatments.

The goal of defeasible reasoning and of nonmonotonic logic is to work
out what can be concluded when there are contradictory premises or
when premises can change (e.g., in the Yale Shooting Problem, the gun
is initially unloaded but later becomes loaded).  However, these are
not really relevant for assurance cases.  In Assurance 2.0 we use
Natural Language Deductivism (NLD) and expect completed assurance
cases to be deductively sound, although we may tolerate some
``inductive'' argument steps\exfootnote{In an inductive argument step,
the subclaims and evidence ``strongly suggest'' the parent claim but
do not imply or entail it as a deductive step would.  We discourage
inductive steps because there is always a defeater hidden in the
``gap'' that prevents the step being deductive.  In our opinion, it is
preferable to acknowledge this ``gap'' by supplying an explicit
``something missing here'' subclaim (recall Figure
\ref{decomp-doubts}) that makes the argument step deductive.  This
subclaim is never discharged, so it highlights and documents a
residual risk.}  if necessary.  That is, we may accept some ``gaps''
in our knowledge, but not contradictions.  Furthermore,
indefeasibility requires that no feasible changes or additions to the
premises can change our conclusions.

While developing a case, and in supporting its review, we use
defeaters (i.e., contradictions) to probe and challenge a case, but we
do not expect to conclude anything from a case with unresolved
defeaters: such a case is acknowledged as imperfect and incomplete.
What we would like to know is: how much of the overall argument does a
defeater cast into doubt, and how much is repaired by a
counter-defeater?  Unfortunately, these purposes are not served by
defeasible reasoning and so we next turn to argumentation theory.

\subsubsection{Argumentation Theory}
\label{argtheory}

Defeasible reasoning and nonmonotonic logic attempt to understand
what conclusions can be drawn from a single argument when some of its
premises are inconsistent.  The field of argumentation on the other
hand, considers multiple competing arguments and tries to deduce which
ones emerge from the competition with their plausibility intact.
Defeasible reasoning employs a defeater relation on premises, whereas
argumentation has the relation of \emph{attack} between arguments.
The two ideas are related however.  If we have an argument and then
introduce a defeater, we can think of this as two arguments: the
original, and a new one consisting of that plus the defeater, and in
which the second attacks the first.

This is \emph{abstract argumentation} theory, introduced by Dung
\cite{Dung95:AIJ} in a paper that has over 4,500 citations.  Despite
the term ``argumentation'' appearing in the name, the level of
abstraction reduces it to an exercise on graphs: arguments are nodes
in a graph that are connected by attack relations and we ask for rules
to determine which sets of nodes are ``accepted.''  Dung defined this
in terms of ``extensions'' but ``labeling'' provides a more intuitive
treatment \cite{Caminada06}: each argument in the graph is labeled
\emph{in} (accepted), \emph{out} (rejected) or \emph{undecided}.  The
\emph{reinstatement} rule on labeling stipulates: an argument is
\emph{in} iff all arguments that attack it are \emph{out}; an argument
is \emph{out} iff it is attacked by at least one argument that is
\emph{in}; arguments that are neither \emph{in} nor \emph{out} are
\emph{undecided}.

As with defeasible reasoning, the purposes served by abstract
argumentation theory are sufficiently different from those in
assurance cases that we do not find any direct application for its
methods.

\ifelse{However, argumentation theory has a connection to logic
programming that supports a computational interpretation for
defeasible reasoning \cite{Chesnevar-etal00} and this may suggest
methods for applying Answer Set Programming (ASP)
\cite{Gelfond&Kahl14} that are being explored by UT Dallas
\cite{Gupta:sCASP22} as part of \cl \cite{Murugesan-all:GDE23}.}{}

\subsubsection{Eliminative Induction and Argumentation}

\label{elimarg}

``Eliminative Induction'' is a method of reasoning that dates back to
Francis Bacon who, in 1620 \cite{Bacon-organon}, proposed it as a way to
establish a scientific theory by refuting all the reasons why it might
be false (i.e., its defeaters).  Modern treatments see falsifiability
as the key characteristic of science \cite{Popper:2014logic} but the
two can be related via Bayesian Epistemology \cite{Bovens&Hartmann03},
where Bayesian methods are seen as the best way to select among so-far
unfalsified theories \cite{Hawthorne93,Vineberg96}.  These Bayesian
methods relate to Confirmation Theory, discussed in Section
\ref{confirmation}.

Weinstock, Goodenough, and Klein \cite{Goodenough-etal:ICSE2013}
develop the idea of Eliminative Induction into a means of assurance they
call \emph{Eliminative Argumentation} where, instead of confirming a
positive claim such as ``the system is safe,'' we instead attempt to
refute the negative claim ``the system is \emph{un}safe.'' A
successful refutation will establish the negation of that claim,
namely ``the system is \emph{not} unsafe.''  In classical (as opposed
to intuitionistic) logic this establishes the positive claim by virtue
of the rule for elimination of double negation, and thereby provides
the desired assurance.  

Weinstock, Goodenough, and Klein represent an eliminative argument as a
\emph{confidence map}, which is rather like an assurance case with
defeaters included and rules for accepting a claim only if all its
defeaters have been eliminated.  Colors are used to highlight the
``positive'' and ``negative' parts of a case.  Diemert and Joyce
\cite{Diemert&Joyce2020} and others \cite{Millet:CERN-LHC23} report
successful application of eliminative argumentation in assurance of
real systems.  We apply some of these ideas in \cla, as described in
Section \ref{elim}.

\subsubsection{Dialectics and Agreement Technologies}
\label{dialectics}

Dialectics refers to the back-and-forth nature of arguments employed
in active debate.  One of the goals of dialectical debate is to reach
an agreed conclusion, so its methods are sometimes referred to as
agreement technologies \cite{Ossowski13}.  

There are many approaches to dialectics and agreement, but an
influential one that has been applied in several domains and is
supported by tools is framed in terms of \emph{argument schemes}
\cite{Walton96,Walton-etal08}.  These are outlines or templates for
many canonical kinds of argument: for example, argument from analogy,
argument from expert opinion, and so on (there are about 30 in
Walton's book \cite{Walton96}).  These are supported by \emph{critical
questions}, which can be thought of as defeaters customized to each
specific scheme.  For example, argument from expert opinion has six
critical questions (e.g., ``how credible is E as an expert source?'').
Raising and responding to critical questions gives rise to the
dialectical element, just as it does in an assurance case challenged
by defeaters.

Despite this apparent similarity, the compendium on agreement
technologies \cite{Ossowski13} contains no reference to assurance or
safety cases.  Nor do more than 1,000 articles available online that
were published in the journal \emph{Argumentation} (Springer).
Assurance and safety cases likewise make no reference to agreement
technologies.  An exception is the work of Yuan and Kelly, who have
applied argument schemes and critical questions to assurance
\cite{Yuan&Kelly11,Yuan&Kelly15}.

\emph{Carneades} \cite{Gordon-etal:Carneades07} is a system that
supports dialectical reasoning in a different way, allowing a
subargument to be \emph{pro} or \emph{con} its conclusion and allowing
weights to be attached to premises.  A \emph{proof standard} is
calculated by ``adding up'' the \emph{pro}s and \emph{con}s supporting
the conclusion and their attendant weights (rather like the labelings
of argumentation theory).  For example, a claim is ``in'' if it is not
the target of a \emph{con} that is itself ``in'' (unless it is also
the target of \emph{pro} that is ``in'' \ldots); a conclusion is supported
to the \emph{preponderance of evidence} proof standard if it has at
least one \emph{pro} argument that is ``in'' and weighs more than any
``in'' \emph{con} argument.  The system, which is available at
\url{http://carneades.github.io/}, provides several kinds of argument
graphs for visualizing arguments.  Recent work by Takai and Kido
\cite{Takai&Kido14} builds on these ideas and is implemented in the
commercial assurance tool Astah GSN \cite{Astah-gsn}.

\subsection[Approach Adopted in \cl]{Approach Adopted in C{\small LARISSA}}
\label{defeaters-in-cl}

We have briefly reviewed how defeaters are used and represented in
defeasible reasoning, argumentation theory, eliminative argumentation,
and in dialectics and agreement technologies.  The notion of
``defeater'' comes from defeasible reasoning, but we did not find much
else in that field that is relevant to our concern since we require
inconsistencies (i.e., defeaters) to be resolved rather than to reason
in their presence.  Argumentation theory has the same defect, but
provides the useful notion of an argument being \emph{in} or
\emph{out} and argumentation schemes and dialectics apply this to
subarguments and provide tool support in the Carneades framework;
this idea is applied to assurance cases in Astah GSN\@.
Eliminative argumentation uses several of these ideas and strongly
advocates that the search for and elimination of defeaters should be a
key element in assurance.  We build on these latter elements in the
\clasce\ platform.

The desired conclusion to the development of an assurance case is that
all defeaters are resolved, so that only a positive case remains,
although this may contain residual doubts (see Section
\ref{residuals}).  During development, active defeaters may be present
and it can be useful to see how much of the case is thereby called
into question.  Unlike argumentation theory, which supports this using
\emph{in} and \emph{out} judgments, we do it using a three-valued
logic that adds \texttt{unsupported} to the traditional \texttt{true},
\texttt{false} valuations of logic.
The mere act of pointing a defeater at a node means that the claim
affected by the defeater (this is defined below) is considered
\texttt{unsupported} and the overall argument can no longer be
assessed as logically valid or sound.

As stated above, the intended end state in development of an assurance
case is a fully valid and sound argument where defeaters have either
themselves been defeated, or the issues they identify have been
resolved.  The defeaters, therefore, serve only a transient purpose
and we could decide that there is no need to define an interpretation
for arguments where defeaters have their own subcases as these will
not be present in the finished case.  In this view defeaters and their
subcases function rather like specialized comments and the reasons for
accepting or rejecting them can be recorded in the narrative
associated with the defeater.

However, there are circumstances where it is appropriate for a
defeater to be supported or refuted by a detailed subcase.  An example
is where different parties do not agree on the validity of a given
defeater: e.g., the evaluators of a case propose a defeater and are
not convinced by the developer's informal arguments against it, so
each party develops a subcase to advance its point of view.

Such investigation of the defeater should eventually resolve its claim
as \texttt{true} or \texttt{false}.  If a defeater is eventually
supported by a sound subcase, so that its claim is \texttt{true}, then
the defeater is said to be confirmed or \emph{sustained} and the main
argument, and possibly the system it is about, must be modified to
overcome the flaw that has been identified.  After these
modifications, the defeater and its subcase should no longer apply,
but we might like to retain them in the case as documentation to
assist future developers and evaluators.  Thus, a defeater can be
marked \emph{addressed} and it and its subcase will then be treated as
a comment.  Because the defeater no longer applies to the now modified
primary argument, a narrative description of the original problem and
its resolution should be added to the defeater node.  But this may be
difficult to understand because the context has changed from the
original to the modified argument, so another choice is to alter the
previously sustaining subcase for the defeater into a refuted subcase
(see below) for the modified primary argument.  An alternative
response to a sustained defeater, provided the identified flaw is
judged suitably insignificant (see Section \ref{residuals}), is to
explicitly accept it as a \emph{residual doubt}.

If we suspect that a defeater is a ``false alarm,'' or it is one that
has been overcome by modifications to the original case (as above),
then our task is to \emph{refute} it: that is, to develop a subcase
that shows it to be \texttt{false}.  One way to do this is with a
second-level defeater that targets the first defeater or some part of
its subcase; another is by use of \emph{counter-evidence}
\cite{Bloomfield-etal:defeaters24}.  If the assurance subcase for that
second-level defeater is sustained, then the first defeater is said to
be \emph{refuted} and it and its subcase play no part in the
interpretation of the primary case, but can be retained as commentary
to assist future developers and evaluators who may entertain doubts
similar to that which motivated the original defeater.  It can be
tedious to fully refute a defeater, so once we are satisfied that the
defeater is defeated, we may, with suitable narrative explanation,
leave the refutational subcase incomplete and manually mark the
defeater as \emph{addressed}.

The introduction of defeaters with subcases and refutational arguments
mean that our method for assessing completeness and logical validity
in assurance arguments needs to be revised.  In particular, we need to
consider how assessments of \texttt{false} propagate through the
argument.

It might seem that we could look to the methods of defeasible
reasoning or nonmonotonic logic for this purpose, but the goal of
these methods is to work out what can be concluded when there are
contradictory premises or when exceptions are added, whereas, in
Assurance 2.0, our goal is to determine which parts of a case
\emph{are} contested: that is, called into question by defeaters,
possibly at several levels.

Furthermore, we do not formally interpret the natural language of
claims when assessing validity in an assurance argument, and the same
applies to claims made by defeaters.  In particular, if a defeater
with claim $X$ pointing to a claim $A$ is sustained, we do not suppose
that some logical combination of $A$ and $X$ is thereby justified; we
accept that the claim $A$ is challenged and revise it and/or its
supporting subcase to overcome the source of doubt.  Of course, we
must make the human judgment that $X$ has some impact on the
credibility or relevance of $A$ but we do not reduce this to some
logical requirement such as $X \equiv \neg A$.  Having said that, in
Section \ref{elim} we will introduce a circumstance where we do
recognize the special case in which the defeater's claim is the
negation of that in the node that it points to; we call these
\emph{exact} defeaters (the general kind are then known as
\emph{exploratory} defeaters).

We now develop the propagation rules for validity in the presence of
defeaters.  Since a defeater can point to any kind of node, we define
the claim \emph{affected by} the defeater to be the node pointed to if
this is a claim or defeater, and otherwise the parent claim (which may
be a defeater) of the node pointed to.

We first consider propagation from a defeater to its affected claim.
When the claim in a defeater is assessed \texttt{false}
it means the defeater is refuted; hence, the main argument (or
subargument for lower-level defeaters) is exonerated and its claims
are assessed as if the defeater were absent.  When the claim in a
defeater is assessed \texttt{unsupported} (which also applies when the
defeater has no subcase\textbf{---}i.e., it is merely a doubt), then
so is the claim affected by the defeater.  And when the claim in the
defeater is assessed \texttt{true}, then the affected claim is also
assessed \texttt{unsupported}; it cannot be assessed \texttt{false}
because the defeater may not precisely refute the affected claim
(unless it is an exact defeater, which is considered later), but
merely call it into question.

These assessments override those due to any other nodes pointing to
the affected claim (which may affirm it as \texttt{true}): when a
claim is challenged by a \texttt{true} or \texttt{unsupported}
defeater, we have to accept that it is called into question.  However,
the appropriate response may require further diagnosis: if the
defeater is \texttt{unsupported} then the defeater's subcase needs more
work, while if it is \texttt{true} the main argument needs to be
revised (and possibly also the system concerned).

Finally, we consider propagation of assessments through reasoning
steps; recall, in NLD, individual reasoning steps are intended to be
deductively valid and are interpreted as material implications of the
form shown in formula (\ref{decomp-logic}).  The sideclaim and
subclaims constitute the \emph{antecedent} to this implication.  Thus,
when all claims in the antecedent are assessed \texttt{true} then, by
the rules of classical logic, so is the parent claim.  And, in our
extension to this interpretation, if any antecedent claims are
\texttt{unsupported}, then the parent claim is also.  But suppose some
claims in the antecedent are assessed \texttt{false}.  Since they are
conjoined, the whole antecedent becomes \texttt{false}; does this mean
we should assess the parent claim as \texttt{false} too?

It does not: in abstracted form, it would be attempting to derive
$\neg A \supset \neg B$
from $A \supset B$, and this is the logical fallacy of ``denying the
antecedent.''\excite{denying-antecedent-wiki}\footnote{An informal
illustration of denying the antecedent uses the subclaim/premise ``if
college admission is fair, then affirmative action is unnecessary'' to
fallaciously infer the claim/conclusion ``college admission is not
fair, so affirmative action is needed.''}  Moreover, there is a
further problem: if the antecedent is \texttt{false}, then it can
imply any parent claim: this is the ``false implies everything''
problem.\exfootnote{$A \supset B$ is equivalent to (or is defined as)
$\neg A \vee B$ so, if $A$ is \texttt{false}, $\neg A$ is
\texttt{true}, and $\neg A \vee B$ is \texttt{true} independently of
$B$.}  Thus, in general, we cannot propagate \texttt{false} upward
through reasoning steps;\footnote{There is a special case where the
(conjunction of) subclaims is \emph{equivalent} to the parent claim
(given the sideclaim) rather than merely entailing it: it is
legitimate to propagate \texttt{false} in this case but \clasce\ does
not do so (because it does not attempt to interpret the language of
claims).  Instead, the case should be modified to use an exact
defeater.} we must do something weaker and the appropriate response is
to assess the parent claim as \texttt{unsupported}.

\clasce\ has a ``validity plugin'' that can evaluate validity of an
argument in the presence of defeaters using the method presented
above, which is described in more detail in a separate report
\cite{Bloomfield-etal:defeaters23}.  It allows developers and
evaluators to see which defeaters have been refuted and which are
still active, and which parts of an argument are called into question
by the active defeaters.  This is intended to support dialectical
exploration of assurance arguments, so that confidence can be probed
and ultimately more firmly established.

Currently, defeaters are proposed by human developers and reviewers
but we are exploring systematic, and potentially automated, ways to
generate useful classes of defeaters.  One possible avenue of inquiry
is to partially interpret natural language claims within some ontology
so that in a claim of the form ``A refines B'' the relation
``refines'' will be recognized and defeaters proposed based on common
misuses of the term.

Defeaters introduce refutational reasoning to Assurance 2.0 and this
allows an alternative form of assurance argumentation, as we now
describe.

\subsubsection{Exact Defeaters and Counterarguments}

\label{exact}

The basic methodology of Assurance 2.0 supports development of
\emph{positive cases} where a constructive argument is developed in
support of some beneficial claim about a system.  Nonetheless, we
explicitly introduce defeaters and confirmation measures to help
address complacency and bias by inviting consideration of contrary
points of view.  Moreover, we recognize that it can sometimes be
useful to consider fully contrary claims, evidence and arguments, and
we introduce \emph{exact} defeaters for this purpose; effectively,
they allow us to introduce negation into an assurance argument.

An exact defeater is one that: a) points to a node that is either a
claim or another defeater that b) lacks a subcase, and c) whose own
claim is the negation of the one pointed to.  Because claims in
\clasce\ are written in natural language, it is not trivial to
determine if one claim is the negation of another.  Accordingly,
\clasce\ provides an explicit selection in its interface to indicate
that a defeater should be treated as the exact negation of the claim
or defeater that it points to.  Furthermore, the node pointed to may
have a subcase, but it will be ignored (and indicated so in the
graphical presentation) when the node becomes the target of an exact
defeater.  This is to support exploratory development of a case
without having to undo or redo previous work.

The propagation rules for exact defeaters are simple: if the exact
defeater is assessed \texttt{unsupported}, then so is the node that it
points to; otherwise the assessment of the node pointed to is the
logical negation of the assessment of the claim in the defeater.
Notice that whereas exploratory defeaters \emph{augment} the main
argument by providing an exploratory investigation or commentary,
exact defeaters are used as a reasoning step \emph{within} the main
argument.

Exact defeaters support the notion of a \emph{counterargument}, which
can be useful in circumstances quite apart from contested defeaters.
One such is where developers of an assurance case find it difficult to
build a subcase to justify some claim $A$: they may be able to develop
their understanding or to gain insight by attempting instead to
justify the \emph{counterclaim} $\neg A$, or by attempting to
\emph{refute} the claim $A$.  Another is where the developers of a
large case are unpersuaded by the subcase for a claim supplied by
others; again, it may be useful to develop and explore the case for a
counterclaim or refutation to the given claim.

Counterarguments can take two forms: positive or refutational.  A
\emph{positive counterargument} aims to establish a counterclaim $\neg A$
(introduced using an exact defeater) in the standard way, whereas a
\emph{refutational counterargument} aims to refute the original claim
$A$ (i.e., show it to be \texttt{false}).
Counterclaims and refutations are equivalent in classical logic: that
is to say, verifying $\neg A$ is the same as refuting $A$.  However,
exploiting this relationship may not be so straightforward in an assurance
argument.  One reason is philosophical: in an assurance case we
generally prefer a positive argument to a refutational one: that is,
we prefer to establish that the system is safe, rather than it is not
unsafe (eliminative argumentation would take the contrary position).
In logic, this preference for positive arguments corresponds to use of
\emph{intuitionistic} rather than classical reasoning: intuitionistic
logic eschews the axiom for ``excluded middle'' (i.e., $A \vee \neg
A$)\footnote{This also excludes elimination of double negation.} so
that all proofs must be of a positive, constructive nature.  It might
seem that this would be a good choice for assurance case arguments,
but there difficulties in doing so.  The interior argument steps of an
assurance argument are \emph{a priori} premises, and in an
intuitionistic setting we must be careful that these do not
accidentally introduce the excluded middle.  For example, most
assurance cases have essential steps that decompose over hazards and
since we cannot \emph{know} that all hazards have been identified
(although we try very hard to do so), these steps have a somewhat
non-intuitionistic character: instead of arguing that the system is
safe in all circumstances, we are saying that we know of no
circumstances where it is unsafe.

In general, it seems very onerous to insist and to check that all
argument steps are intuitionistic (it is hard enough to insist that
they are deductive) so we suggest it is best to conduct assurance
arguments within classical logic, but with an informal preference for
positive perspectives.

Refutational counterargument are those where we establish that some
claim (which may be in a defeater) is \texttt{false}.  However, there
are only two argument steps that can do this.  One is evidence
incorporation, when the evidence explicitly refutes its ``something
useful'' claim (e.g., when tests reveal failures), and the other is
when the claim is the parent (or ``target'') of an exact defeater, in
which case the subargument is transformed into a positive
counterargument.

\memo{Need to conclude this thought.}

\subsubsection{Support for Eliminative Argumentation}

\label{elim}

Exact defeaters allow \clasce\ to support Eliminative Argumentation,
which was introduced in Section \ref{elimarg}.  Recall that in
Eliminative Argumentation we attempt to refute a negative claim, such
as ``the system is unsafe'' rather than confirm its positive version
``the system is safe.''  To do this in \clasce, we retain a positive
top claim and then attach an exact defeater to it.  Owing to the
refutational context in negative cases, it can be useful to have
\emph{disjunctive} decomposition blocks and \clasce\ supports these
(see \cite{Bloomfield-etal:defeaters24} for details).  We can then
disjunctively decompose ``the system is unsafe'' claim into
hypothesized reasons why this might be so.  Eventually, we will arrive
at detailed negative claims, such as ``the software may generate a
runtime exception'' that can be refuted by counterevidence (e.g.,
static analysis guaranteeing the absence of runtime exceptions) or by
introducing another exact defeater at this lower level and generating
a positive subcase for the counterclaim ``the software generates no
runtime exceptions.''

\section{Residual Doubts and Risks}
\label{residuals}

A sound assurance case delivers confidence in its top claim.
Typically, this is either an ``internal'' probabilistic claim (recall
Section \ref{probs}) about some property related to critical failure
(e.g., a bound on failure rate or time to failure), or a logical claim
asserting that the system has no faults that could lead to a critical
failure.  In the latter case, the degree of confidence in the claim
can support a conclusion concerning rate of critical failure (see
Section \ref{cbi}).  Thus, in either case, we derive strong confidence
in the related claims that the system contains no critical faults and
that it will suffer few critical failures.

However, the assurance case may contain residual doubts: these are
potential defeaters that we are unable to eliminate or mitigate.  They
may be due to uncertainty in the environment: for example, the system
is designed to withstand two faults and historical experience
indicates this is sufficient, but it is always possible for it to
encounter more than that.  Or they may be due to limitations of human
review (e.g., human requirements tracing cannot be guaranteed to be
free of error), or of automated analysis (e.g., automated static
analysis may be unable to discharge some proof obligations, leading to
alarms that may be false and must be reviewed by humans).  If true,
these defeaters may expose a hazard and hence a fault.

In assessing soundness and probabilistic confidence in an assurance
case, we ignore residual doubts (recall Sections \ref{soundness},
\ref{propagation} and \ref{defeaters-in-cl}): thus we achieve
confidence in the absence of faults by ignoring the remaining doubts
and defeaters that could possibly expose their existence!  The
justification for doing this is that we assess the likelihood of such
faults, or more particularly the risk that they pose (i.e., the
likelihood of activating them multiplied by the potential cost of the
failure they may incur) to be insignificant.

Let us first consider the kinds and significance of residual doubts
that may be present.  Our concern is that these doubts may be
sufficient to undermine the indefeasible justification of a claim
$C$, so that we have to consider the possibility that $C$ is false and
$\neg C$ is true.  The probability of this event can be conditioned on
whether the argument supporting $C$ is deductive or not.  Thus,
\begin{eqnarray*}
P(\neg C) & = &
 P(\neg C\vbar \mbox{deductive}) \times P(\mbox{deductive})\\
&& \mbox{}+ P(\neg C\vbar \neg \mbox{deductive}) \times P(\neg \mbox{deductive}).
\end{eqnarray*}
It is conservative to set any term in this equation to 1.  Hence
\begin{eqnarray}
P(\neg C) & \leq &
P(\neg C\vbar \mbox{deductive}) \times 1
\mbox{}+ 1 \times P(\neg \mbox{deductive})\nonumber\\
& \leq & P(\neg C\vbar \mbox{deductive}) + P(\neg \mbox{deductive}).\label{ded}
\end{eqnarray}

If we have correctly identified all residual doubts to the argument,
then $P(\neg \mbox{deductive})$ will be the cumulative probability of
those doubts that concern deductiveness (e.g., the unsupported
``something missing here'' claims
of Figure \ref{decomp-doubts}), which we can write as
$P(\mbox{deductiveness doubts}).$

That takes care of the second term in the right hand side of
(\ref{ded}), so we now consider the first term.  We can condition this
on whether the deductive argument supporting $C$ is valid or not.
\begin{eqnarray*}
P(\neg C\vbar \mbox{deductive}) & = &
P(\neg C\vbar \mbox{deductive} \wedge \mbox{valid}) 
\times P(\mbox{deductive} \wedge \mbox{valid})\\
&&\mbox{} +
P(\neg C\vbar \mbox{deductive} \wedge \neg \mbox{valid}) 
\times P(\mbox{deductive} \wedge \neg \mbox{valid}).
\end{eqnarray*}
Again, It is conservative to set any term in the equation to 1.
Furthermore, validity of the argument can be assured mechanically (in
\clasce\ it is largely assured by construction, and checked by its
validity plugin) so we assume any invalid argument has already been
rejected; hence $P(\mbox{deductive} \wedge \neg \mbox{valid}) = 0.$
Thus
\begin{eqnarray}
P(\neg C\vbar \mbox{deductive}) & \leq &
P(\neg C\vbar \mbox{deductive} \wedge \mbox{valid}) 
\times 1 + 1 \times 0\nonumber\\
& \leq &
P(\neg C\vbar \mbox{deductive} \wedge \mbox{valid}). \label{valid}
\end{eqnarray}

\label{resinterior}

Finally, we consider (\ref{valid}) and condition
this on whether the deductively valid argument is sound or not.
\begin{eqnarray*}
\lefteqn{P(\neg C\vbar \mbox{deductive} \wedge \mbox{valid})=}\\
&\lefteqn{P(\neg C\vbar \mbox{deductive} \wedge \mbox{valid}
    \wedge\mbox {sound})
\times
P(\mbox{deductive} \wedge \mbox{valid} \wedge\mbox {sound})}\\
&&\mbox{}+
P(\neg C\vbar \mbox{deductive} \wedge \mbox{valid}
    \wedge \neg \mbox {sound})
\times
P(\mbox{deductive} \wedge \mbox{valid} \wedge \neg \mbox {sound}).
\end{eqnarray*}
If the argument is deductive and valid and sound, then $C$ is true and
$\neg C$ is false.  Hence, the first term on the right hand side is 0
and then, again setting some terms conservatively to 1, we have
\begin{eqnarray}
P(\neg C\vbar \mbox{deductive} \wedge \mbox{valid})& \leq &
0 \times 1 + 1
\times
P(\mbox{deductive} \wedge \mbox{valid} \wedge \neg \mbox {sound})\nonumber\\
& \leq & P(\mbox{deductive} \wedge \mbox{valid} \wedge \neg \mbox {sound}).\label{sound}
\end{eqnarray}
Now, $P(\mbox{deductive} \wedge \mbox{valid} \wedge \neg \mbox
{sound})$ will be the probability of all residual doubts concerning
soundness of the argument, which we can partition into those
concerning the evidence incorporation steps, which we write as
$P(\mbox{evidential doubts}),$ and those concerning the interior
reasoning steps of the argument, which we denote $P(\mbox{interior
doubts}).$ Thus combining (\ref{ded}) to (\ref{sound}),
we have
\begin{eqnarray}
\lefteqn{P(\neg C) \leq }\nonumber\\
& P(\mbox{deductiveness doubts})
+P(\mbox{evidential doubts})
+P(\mbox{interior doubts}).\label{doubts}
\end{eqnarray}
(It is possible this relationship can be
derived more readily by the ``sum of doubts'' methods described in
Section \ref{propagation}.)

We have already described deductiveness doubts; an example would be
the case that more than two faults afflict a system that is designed
to withstand only two faults.  Presumably the argument will contain a
decomposition block on the number and nature of faults, and this will
not be deductive unless the ``impossible'' case of more than two
faults is taken into account.  There are two ways to record this
concern: one is to add an assumption that there will be no more than
two faults and to attach some probabilistic doubt that will be
propagated by the methods of Section \ref{propagation}; the other is
to attach an unresolved defeater to the decomposition block that
asserts it is nondeductive.  The second of these will become a
residual risk and we prefer this treatment as it is a more explicit
recognition of the concern.

Evidential doubts are an interesting topic.  We advocate paying
considerable attention to the assessment of evidence (recall Sections
\ref{confirmation} and \ref{evincsec}) so any doubts are surely
already included in those assessments.  However, it may be that we
have systematic doubts about certain types of evidence: for example,
static analysis may generate ``false alarms'' that must be rejected by
fallible human review.  We could reduce our probabilistic assessment
$P(C \vbar E)$ for claims $C$ supported by such evidence $E$, but
presumably this reduction will be minor or we would do something about
it.  However, systematic concern about false alarms may then be lost
in the details of each evidential step of this kind and the cumulative
impact of these doubts may not be recognized.
Thus, we think there can be merit in recording small but systematic
concerns about evidence as residual evidential doubts.

Interior doubts are those where we accept that a reasoning step is
deductive and logically valid, but are unsure that it applies to its
context in the case (hence, it may be unsound).  These doubts can
arise for two reasons: one is that the justification for the step is
unconvincing; the other is that we may suspect the step could be
\emph{wrong}.  In the first case, it would be best to develop a
narrative justification that is convincing, but in some circumstances
it might be acceptable to attach a defeater marked as a residual
doubt.  In the second, we have doubts that the subclaims really do
entail the parent claim (and therefore we distrust the justification
also).  This kind of doubt could have serious consequences if it turns
out to be true, and therefore it cannot be considered merely ``residual''
and must be investigated and eliminated.

With these caveats, (\ref{doubts}) provides a classification for the
sources of residual doubts that should be investigated in validation
of an assurance case to ensure they do not impact its indefeasibility.

Ideally, assessment of residual doubts should consider the faults they
might precipitate and the failures those faults might cause, together
with the frequency of their occurrence and severity of their outcomes
(i.e., their \emph{risk}, the product of likelihood and consequences),
all in the worst case.  This is feasible for some doubts, such as
those concerning the maximum number of sensor failures that might
occur.  For others it seems less so.  Suppose for example, that we
have residual doubts about static analysis because it generates many
proof obligations that cannot be discharged automatically and require
human review.  Here, the best we can do might be to collect statistics
on human reliability for this task in an effort to constrain the
potential frequency of failure, since the potential consequences seem
very hard to ascertain.  If we assume the worst case, that all human
reviews are in error and these cases represent real bugs, then each
bug might be encountered very rarely, but collectively they could
arise unacceptably often.  This would be an example where each
instance of a residual risk is minor but their aggregate is not, and
action must therefore be taken to eliminate or further mitigate their
cumulative impact (e.g., by using a better theorem prover or a diverse
means of analysis).\footnote{Observe that autonomous systems, such as
self-driving cars, can exhibit risks like this, but for different
reasons.  For example, the vision system of the car may misinterpret
some scenes; on one hand, each reason for misinterpretation may apply
very rarely but somewhat persistently so that in total they cause the
vision system to fail quite often; alternatively, the
misinterpretation may apply only to isolated frames with preceding and
following frames interpreted correctly, so the consequences of the
misinterpretation is an inconsequential ``blip''
\cite{Jha-etal:Safecomp20}.}

It may not be necessary to strive for exactness in assessment of
residual doubts: all we need to know is that their associated risks
are well below some acceptable threshold.  At present, we suggest
these risks, and the threshold where they are considered significant,
should be assessed and documented by best-efforts expert review.
Later, it may be possible to develop ways of using historical
experience and conservative probabilistic modeling to assist this
process.

We suggest it is useful to categorize residual risks into three levels
(plus 1, see below).
\begin{description}

\item[Significant:] an individual residual doubt poses a risk that is
    above the threshold for concern.  In this case, the issue cannot
    be considered a merely ``residual'' doubt, but must treated as a
    defeater and later eliminated or mitigated.

\item[Minor:] an individual residual doubt poses a risk that is below
    the threshold for concern, but it is possible that many such might
    cumulatively exceed the threshold.  An example could be static
    analysis, where we use fallible human review to evaluate proof
    obligations that the automation cannot decide.  The number of
    such risks needs to be managed explicitly: 10 might be acceptable,
    but not 100.

\item[Negligible:] multiple residual doubts of a similar kind
    collectively pose a risk that is below the threshold for concern.
    This may arise when the source of doubt occurs many times but is
    adjudged to have trivial consequences.  An example (depending on
    local policy) might be ``style'' warnings from a static analyzer.

\end{description}

\yy \clasce\ allows active defeaters to be annotated with their
estimated severity, based on the scale above, and can report on the
totals in each category.  Those judged significant must be eliminated
or mitigated.  Those judged minor should be examined and their
cumulative potential impact assessed.  If the number and cumulative
severity of some category of minor risks can be kept below the
threshold of concern, then we consider that category to be
\textbf{Manageable}.  At final assessment, the only residual doubts
that remain should be those considered minor but manageable and those
considered negligible.

The considerations above provide assurance that defeaters labeled as
residual doubts really are ``residual.''  In addition, we may wish to
review how confidence in these assessments affects the overall case.
Section \ref{defeaters-in-cl} described how \clasce\ can propagate
probabilistic confidence and can color-code nodes according to
user-selected thresholds on these values (recall Figure
\ref{traffic}).  Furthermore, manual adjustment can be used to assign
confidence to assessment and propagation of residual doubts.  We
attach little significance to the actual values, but the visualization
can help comprehend and assess the potential impact of residual doubts
on the overall case.

\newpage
\section{Sentencing Statement}
\label{sentencing}

Assurance cases generally serve a singular purpose: to support the
decision whether a system may be deployed.  However, those who make
that decision typically have a different view on its assurance case
than those who develop it.  The task of the developers is to construct
a system that is safe and an assurance case that provides true and
compelling reasons for believing it is so.  The task of the assessors
is to be sure that the developers have accomplished this: they do not
repeat the work of the developers and reconstruct the assurance case,
they review it.  However, the reviews may involve individuals or teams
who occupy different roles within the processes leading up to the
deployment decision.  For example, an auditor might be focused on
process and regulatory compliance in the construction of the system
and its assurance case, whereas a technical evaluator will want to
gain a deep understanding of how the system works and how its safety
is ensured.  It is important that an assurance case and its supporting
tools provide information and means for comprehending it that support
these diverse perspectives.

Graydon contends \cite{Graydon17:visions} that cases developed to
support one perspective or ``vision'' may be misunderstood by those
with different perspectives.  We are not so pessimistic but we believe
that assurance cases and their tools must support communication, so
that diverse reviewers can develop confidence and consensus in their
understanding of the system and its case, and they must also support
reasoning, so that reviewers can test their understanding and can also
challenge the case.  We have described several mechanisms by which
Assurance 2.0 and \clasce\ hope to achieve these goals.  These include
a limited number of basic building blocks so that arguments are
readily interpreted, strict criteria on what constitutes a soundly
reasoned argument step and evidence of adequate weight, tools for
inspecting and navigating arguments and for evaluating them from both
logical and probabilistic perspectives, and methods and tools for
challenging them by means of defeaters.

The final assessment of an assurance case and the corresponding
decision on system deployment are serious matters.  The exact form of
the assessment and review varies according to the system and its
critical properties, and the industry concerned: assessment for
nuclear power generation is different than for civilian aircraft.  And
decisions on security are very different to those for safety.  Because
there are diverse concerns and practices across industries and diverse
roles and perspectives within individual assessment teams, we believe
that an assurance framework and its tools must provide means to
construct different viewpoints and their summaries on each assurance
case.  We call these \emph{metacases}: cases about the case,

We expect assessors, assisted by those in supporting roles, to avail
themselves of the tools and intellectual structures around an
assurance case to develop metacases that actively explore both the
positive and negative aspects of the case and that challenge their
understanding by proposing defeaters and exploring other questions.

The assessor's task should be informed by and conclude with a
``sentencing statement'' that indicates their diligent execution of
these tasks and declares their understanding of the system and its
context, the key points of its architecture and design, its hazards
and their mitigations, the soundness and probabilistic confidence of
the assurance argument with its supporting theories, models, and
evidence assembly, and the relationship of the top claim to acceptance
criteria for deployment.\\[0.9ex]
\noindent Possible headings for a sentencing statement could be the
    following.\\[0.5ex] ``\emph{On the basis of this assurance case and
    an examination of other relevant documentation, I judge the
    proposed system to be adequately safe/unsafe}\ldots'' (or
    ``\emph{the case is insufficient to make a judgment}'').\\[2ex]
``\emph{I believe my judgment of this case is sound and valid
because}\ldots
\emph{\begin{itemize}
\item I understand the context and criticality of the decision\ldots
\item I understand the system\ldots
\item I find a clear thread of reasoning from evidence to claim\ldots
\item The evidence provided is sufficient/insufficient to support evidence-based
decision making\ldots
\item I have actively explored doubts\ldots
\item I have also identified what evidence would be capable of disproving\ldots
\item I have considered and addressed biases and fallacies\ldots''
\end{itemize}}

\yy \cl\ is experimenting with the decision support needed to assist
evaluators make and substantiate these judgments, with explicit links
from Assurance 2.0 concepts and \clasce\ functions to the bullet points
above.

\newpage
\section{Summary and Conclusion}

We have explored and described methods for gaining and assessing confidence in
assurance cases based on Assurance 2.0 and its automated assistance
with \clasce.  Here, we summarize these methods and provide brief
conclusions.  We do not provide references here: they can be found in
the earlier sections specific to each topic.

Assurance 2.0 is more rigorous and demanding than earlier treatments
of assurance cases, but we argue that this simplifies the development
and assessment of cases because issues that were previously treated in
an \emph{ad hoc} manner and subject to contention and challenge are now made
explicit and treated systematically.

In particular, we are explicit that the goal of Assurance 2.0 is
indefeasible justification, meaning we must be confident there are no
overlooked or unresolved doubts that could change evaluation of the
case.  A consequence of this is a strong preference in Assurance 2.0
for argument steps to be deductive, and for steps that are merely
inductive to acknowledge this and to explicitly manage the doubts
thereby admitted.  Similarly, evidence in Assurance 2.0 is weighed
very deliberately using confirmation measures and we distinguish
carefully between facts established by the evidence (claims about
``something measured'') and inferences drawn from it (claims about
``something useful'').

These rigorous requirements and other supporting constraints enable
straightforward evaluation of the positive criterion for assurance
case arguments that we call soundness.  Note that we say the
evaluation is straightforward, meaning it is clear what must be done,
not that it is easy: it requires sophisticated technical judgment, but
this judgment can focus on technical issues without being distracted
by unmanaged doubts and contested interpretations.

Another way in which Assurance 2.0 simplifies the assessment of
assurance cases is by being clear about what is developed within the
assurance argument and what is referenced and integrated by it via
external theories and models.  The overall case will reference all
necessary theories, models, and evidence assemblies but the internal
detail of many of these items is absent from the assurance argument,
not because they are unimportant but because they each have
specialized form and content and are therefore well suited to
presentation and assessment by scientific and engineering methods
traditional to their fields and these are available in their theories
and other referenced documents.  The assurance case argument, on the
other hand, must integrate evidence and subcases supported
by these disparate items and its structured, logical form is tailored
for that function and dedicated to it.

Whereas traditional assurance case arguments seem almost arbitrary in
their structure, so that reviewers do not know what to expect,
arguments in Assurance 2.0 generally follow a systematic pattern where
general claims at the upper level are refined into more precise claims
using concretion steps, then substitution steps are used to elaborate
these claims about high level models into claims about low level
models and their implementations, and these lowest level claims are
discharged by evidence.  Substitution steps relate a claim about one
model to a possibly different claim about a possibly different model,
although either the claim or the model is typically held constant.
Along the way, the argument may divide into subcases using
decomposition or calculation steps that enumerate a claim over some
structure (e.g., over components, requirements, hazards, etc.) or
split the conjuncts of a compound claim.  This structure may recurse
within subcases.  For example, we may have a lower-level claim that
software development conforms to standards, and will use a concretion
step to refine this to a specific standard and then develop the rest
of the subcase in the manner just described.

Their systematic structure allows argument steps in Assurance 2.0 to
be limited to just the five basic forms mentioned above (concretion,
substitution, evidence incorporation, decomposition, and calculation).
These each have a precise and obvious purpose and it is generally
straightforward to decide which to use at each argument step (see the
``helping hand'' visual mnemonic of Figure \ref{helping-hand}).  Each
type of argument step has side-claims that ensure it is used
appropriately and with a sound (e.g., deductive) justification.

Soundness is the most fundamental valuation for an assurance case: it
tells us that the argument and its evidence truly do support the top
claim, but it does not tell us how strongly they do so.  We therefore
define a method for probabilistic valuation that does this and
\clasce\ can color argument nodes accordingly to support visual
comprehension of the weak and strong parts of an argument.  We apply
probabilistic valuation only to sound assurance case arguments, and
this eliminates (although we need to confirm this claim) the
vulnerabilities that have been found in other probabilistic forms of
assessment.  Our probabilistic methods are conservative and the
numerical valuations are of limited absolute value, but they do serve
to explore the risk posed by residual doubts and the relative
strengths of different cases for the same system.  This allows
rational tradeoffs of effort and cost versus confidence, which is
needed in developing graduated forms of assurance for different levels
of risk, as exemplified by the SILs (Software Integrity Levels) of IEC
61508, the ASILs (Automotive SILs) of SAE 26262, and the DALs (Design
Assurance Levels) of DO-178C.

While building a forceful positive case, the developers of an
assurance case must guard against confirmation bias.  This can be
assisted by vigorous and active exploration of challenges to, and
doubts about, the case during its construction.  In Assurance 2.0,
doubts are refined and recorded as defeaters, which are nodes in the
graphical representation of the assurance case argument that
explicitly challenge other nodes and that may have their own subcases
to validate or refute them.  Valid defeaters require revision to the
assurance case and possibly the system itself.  

\clasce\ can selectively
reveal or hide defeaters and their subcases and can display the
changes made in response to valid defeaters.
In addition to guarding against confirmation bias, the record of
doubts explored as defeaters assists assessors of the case.  Assessors
begin their work by gaining an understanding of the case, perhaps by
posing ``what-if'' questions, and then probing it more deeply for weak
spots and oversights.  When previously examined defeaters are recorded
as part of the case, assessors may find that their own questions and
doubts have been anticipated and answered, thereby streamlining their
task and also enabling a constructive, dialectical examination of the
case by ``eliminative argumentation.''

All identified defeaters must be examined and resolved.  However, a
conscious decision may be made to accept some as residual doubts.  For
example, a subcase that uses testing to justify absence of runtime
exceptions may have residual doubt due to incompleteness of testing.
The risks posed by such doubts (i.e., the likelihood that they may be
falsified, and the potential impact and cost if they are so) must be
assessed and only those considered tolerable (technically, those
considered minor but manageable, and those considered negligible, see
Section \ref{residuals}) can be allowed to remain as residual risks:
others must be eliminated or mitigated by revisions to the argument or
the system.  The probabilistic valuation of \clasce\ can be used to
help visualize the potential impact of residual doubts on the overall
argument.

In conclusion, Assurance 2.0 assesses confidence in an assurance case
by considering both positive and negative perspectives.  The positive
perspectives are logical soundness and (optionally) a probabilistic
valuation; the negative perspectives are vigorous exploration of
potential defeaters, and careful evaluation of all residual doubts.
During development and, optionally, during evaluation both positive
and negative aspects may be explored simultaneously, but at the
conclusion of both development and evaluation, all potential defeaters
should have been dismissed, or accepted and assessed as residual
risks, and the positive perspective should be one of indefeasible
soundness.  Assessors should not simply inspect and ``sign off'' on an
assurance case; we expect them to actively explore and question both
its positive and negative aspects and to conclude with a ``sentencing
statement'' that declares their understanding of the system and its
context, its hazards and their mitigations, the key points of its
architecture and design, the soundness and probabilistic confidence of
the assurance argument with its supporting theories, models, and
evidence assemblies, their defeaters and residual doubts, and the
relationship of the top claim to acceptance criteria.  \clasce\
provides assistance in these evaluations and together they provide a
comprehensive and rigorous assessment for assurance cases that should
be independent of the vagaries of individual assessors.

These ideas were developed and explored during construction of
\clasce\ using a variety of examples and represent a work in progress.
We plan to develop worked examples to will support more detailed
exploration and exposition that will be published in a companion
report.  In particular, we wish to explore what forms of guidance and
automated support are most useful for developers and assessors.

\paragraph{Acknowledgments.}

The work described here was developed in partnership with other
members of the \cl\ project, notably Varadarajan Srivatsan (and
previously Kevin Driscoll and Brendan Hall) of Honeywell, Gopal Gupta
of UT Dallas, and Kate Netkachova of Adelard.  Separately, N. Shankar
of SRI provided valuable criticism.

This material is based upon work supported by the Air Force Research
Laboratory (AFRL) and DARPA under Contract No. FA8750-20-C-0512.  Any
opinions, findings and conclusions or recommendations expressed in
this material are those of the author(s) and do not necessarily
reflect the views of the Air Force Research Laboratory (AFRL) and
DARPA.

\addcontentsline{toc}{section}{References}
\bibliographystyle{modplain}

\end{document}